%% file: 0_main.tex
\documentclass[mksc,nonblindrev]{informs3-for-posting} 

\OneAndAHalfSpacedXI
\input{src/packages}
\input{src/notations}

\TheoremsNumberedBySection  % (Theorem 1.1, Lema 1.1, Theorem 1.2)
\ECRepeatTheorems

\EquationsNumberedBySection % (1.1), (1.2), ...

\MANUSCRIPTNO{MKSC-2025-0009}

\begin{document}
%%%%%%%%%%%%%%%%

%\setcounter{page}{1}
%\DOI{10.1287/xxxx.0000.0000}%

\RUNAUTHOR{Wang, Zhang, Zhang}

\RUNTITLE{Large Language Models for Market Research}

\TITLE{Large Language Models for Market Research:\\ A Data-augmentation Approach}


\ARTICLEAUTHORS{%
\AUTHOR{Mengxin Wang$^1$, Dennis J. Zhang$^2$, Heng Zhang$^3$}
\AFF{
$^1$Naveen Jindal School of Management, The University of Texas at Dallas, Richardson, TX 75080\\
$^2$Olin School of Business, Washington University in St. Louis, St. Louis, MO 63130\\
$^3$W. P. Carey School of Business, Arizona State University, Phoenix, AZ 85069\\
\vspace{3pt}
\EMAIL{mengxin.wang@utdallas.edu, denniszhang@wustl.edu, hzhan388@asu.edu}}
}

\ABSTRACT{\edit{TBD}
}

\ABSTRACT{Large Language Models (LLMs) have transformed artificial intelligence by excelling in complex natural language processing tasks. Their ability to generate human-like text has opened new possibilities for market research, particularly in conjoint analysis, where understanding consumer preferences is essential but often resource-intensive. Traditional survey-based methods face limitations in scalability and cost, making LLM-generated data a promising alternative. However, while LLMs have the potential to simulate real consumer behavior, recent studies highlight a significant gap between LLM-generated and human data, with biases introduced when substituting between the two. In this paper, we address this gap by proposing a novel statistical data augmentation approach that efficiently integrates LLM-generated data with real data in conjoint analysis.  This results in statistically robust estimators with consistent and asymptotically normal properties, in contrast to na\"ive approaches that simply substitute human data with LLM-generated data, which can exacerbate bias.
\revise{We further present a finite-sample 
performance bound on the estimation error.}
We validate our framework through an empirical study on COVID-19 vaccine preferences, demonstrating its superior ability to reduce estimation error and save data and costs by 24.9\% to 79.8\%. In contrast, na\"ive approaches fail to save data due to the inherent biases in LLM-generated data compared to human data. Another empirical study on sports car choices validates the robustness of our results. Our findings suggest that while LLM-generated data is not a direct substitute for human responses, it can serve as a valuable complement when used within a robust statistical framework. }

%\HISTORY{Empty}

\maketitle
%%%%%%%%%%%%%%%%%%%%%%%%%%%%%%%%%%%%%%%%%%%%%%%%%%%%%%%%%%%%%%%%%%%%%%
 
% Samples of sectioning (and labeling) in MNSC
% NOTE: (1) \section and \subsection do NOT end with a period
%       (2) \subsubsection and lower need end punctuation
%       (3) capitalization is as shown (title style).
%
%\section{Introduction.}\label{intro} %%1.
%\subsection{Duality and the Classical EOQ Problem.}\label{class-EOQ} %% 1.1.
%\subsection{Outline.}\label{outline1} %% 1.2.
%\subsubsection{Cyclic Schedules for the General Deterministic SMDP.}
%  \label{cyclic-schedules} %% 1.2.1
%\section{Problem Description.}\label{problemdescription} %% 2.

\input{src/1_intro}
\input{src/2_reference}
\input{src/3_model}

\input{src/4_theory}
\input{src/5_empirical_1}

\input{src/6_empirical_2}

\input{src/7_conclusion}

{
\bibliographystyle{ormsv080} % outcomment this and next line in Case 1
\bibliography{references.bib} % if more than one, comma separated
}

% Appendix here
% Options are (1) APPENDIX (with or without general title) or
%             (2) APPENDICES (if it has more than one unrelated sections)
% Outcomment the appropriate case if necessary
%

\newpage

% \ECSwitch
% \ECDisclaimer
%%% Main head for the e-companion
\ECHead{
\begin{center}
$ $\\
% \vspace{-25pt}
\underline{Online Appendix}\\
\vspace{8pt} Harnessing Large Language Models for Market Research: A Data-augmentation Approach
\vspace{8pt} 
\end{center}
}

\begin{APPENDICES}
\input{src/app_1_example}
\input{src/app_2_regression}
\input{src/app_3_empirical}
\input{src/app_4_theory}

\end{APPENDICES}

%\ECSwitch
%
%\pagestyle{plain}
%\setlength{\footskip}{8ex} 
%%\renewcommand{\headrulewidth}{0pt} %
%\pagenumbering{arabic} 
%\setcounter{page}{1}
%\renewcommand{\thefigure}{R.L.\ \arabic{figure}}
%\setcounter{figure}{0}
%\input{8_responseLetter}

%%%%%%%%%%%%%%%%%
\end{document}

%% file: src/packages.tex
\usepackage[margin=1in,paper=letterpaper]{geometry}
\usepackage{mathrsfs}
\usepackage{booktabs}
\usepackage{dsfont}
\usepackage[font=scriptsize]{caption}
\usepackage{subcaption}
\usepackage{bm}
\usepackage{bbm}
\usepackage{comment}
\usepackage{todo, soul}
\usepackage[utf8]{inputenc}
\usepackage{pgfplots}
\DeclareUnicodeCharacter{2212}{−}
\usepgfplotslibrary{groupplots,dateplot}
\usetikzlibrary{patterns,shapes.arrows}
\pgfplotsset{compat=newest}
\usepackage{graphicx}
\graphicspath{{../Figures/}}
\usepackage{lineno}
\usepackage{multibib}
\usepackage{multirow}
\usepackage{makecell}
\usepackage[colorlinks=TRUE,citecolor=MyDarkBlue,urlcolor=MyDarkBlue,linkcolor=MyDarkBlue]{hyperref}
\definecolor{MyDarkBlue}{RGB}{158,0,0}

\usepackage{graphicx,booktabs,enumerate,paralist,mdwlist}
\usepackage{algorithm,algorithmic}
\usepackage{mdframed}

\usepackage[capitalize]{cleveref}
\crefname{assm}{Assumption}{Assumptions}
\crefname{defn}{Definition}{Definitions}
\crefname{lem}{Lemma}{Lemmas}
\crefname{app}{Appendix}{Appendices}
\crefname{prop}{Proposition}{Propositions}
\crefname{problem}{Problem}{Problems}
\creflabelformat{problem}{#2(#1)#3}
\crefname{part}{part}{parts}
\crefname{examp}{Example}{Examples}
\crefname{thm}{Theorem}{Theorems}

\usepackage{natbib}
 \bibpunct[, ]{(}{)}{,}{a}{}{,}%
 \def\bibfont{\small}%

%% file: src/notations.tex
\newcites{appendix}{Online Appendix References}
\newtheorem{thm}{Theorem}[section]

\newtheorem{lem}[thm]{Lemma}
\newtheorem{prop}[thm]{Proposition}
\newtheorem{assm}[thm]{Assumption}

\newtheorem{examp}[thm]{Example}

\makeatletter
\def\moverlay{\mathpalette\mov@rlay}
\def\mov@rlay#1#2{\leavevmode\vtop{%
   \baselineskip\z@skip \lineskiplimit-\maxdimen
   \ialign{\hfil$\m@th#1##$\hfil\cr#2\crcr}}}
\newcommand{\charfusion}[3][\mathord]{
    #1{\ifx#1\mathop\vphantom{#2}\fi
        \mathpalette\mov@rlay{#2\cr#3}
      }
    \ifx#1\mathop\expandafter\displaylimits\fi}
\makeatother

\newcommand{\ind}{\mathbbm{1}}
\newcommand{\Real}{\mathbb{R}}
\newcommand{\x}{\mathbf{x}}
\newcommand{\y}{\mathbf{y}}
\newcommand{\thetavec}
{\bm{\theta}}
\newcommand{\Sigmavec}
{\bm{\Sigma}}
\newcommand{\betavec}{\bm{\beta}}
\newcommand{\Prob}{\mathbb{P}}
\newcommand{\Dscr}{\mathcal{D}}
\newcommand{\Lcal}{\mathcal{L}}
\newcommand{\Kscr}{\mathcal{K}}
\newcommand{\Fscr}{\mathscr{F}}
\newcommand{\fvec}{\mathbf{f}} 
\newcommand{\bvec}{\mathbf{b}}
\newcommand{\avec}{\mathbf{a}}
\newcommand{\hvec}{\mathbf{h}} 
\newcommand{\Nscr}{\mathcal{N}}
\newcommand{\Xscr}{\mathcal{X}}
\newcommand{\E}{\mathbb{E}}
\newcommand{\event}
{\mathcal{E}}
\newcommand{\pconverge}{\overset{\text{\upshape{P}}}{\rightarrow}}
\newcommand{\Indicator}{
 \mathds{1}}

\newcommand{\evec}
{\mathbf{u}}
\newcommand{\wvec}
{\mathbf{w}}
\newcommand{\qvec}{\mathbf{q}}
\newcommand{\alphavec}{\bm{\alpha}}
\newcommand{\rvec}{\mathbf{r}}
\newcommand{\uvec}{\mathbf{u}}
\newcommand{\Avec}{\mathbf{A}}
\newcommand{\Bvec}{\mathbf{B}}
\newcommand{\Jvec}{\mathbf{J}}
\newcommand{\Ivec}{\mathbf{I}}
\newcommand{\Bscr}{\mathcal{B}}
\newcommand{\gvec}{\mathbf{g}}
\newcommand{\Omegavec}{\bm{\Omega}}
\newcommand{\Lambdavec}{\bm{\Lambda}}
\newcommand{\epsvec}{\bm{\epsilon}} 
\newcommand{\psivec}{\bm{\psi}}
\newcommand{\Gammavec}{\bm{\Gamma}}

\newcommand{\Ovec}{\bm{0}}

\newcommand{\revise}[1]{{\color{black}#1}}

%% file: src/1_intro.tex
\section{Introduction}
\label{sec:intro}

Large Language Models (LLMs) have revolutionized artificial intelligence (AI) by delivering unprecedented capabilities in natural language processing. These models are built on advanced deep learning architectures known as transformer networks, which excel at handling sequential data and understanding context \citep{vaswani2017attention,radford2018improving}. Trained on extensive and diverse datasets—including vast amounts of text from books, articles, and websites—LLMs are capable of generating human-like responses and performing complex language tasks. For example, Llama 3 was trained on 15 trillion tokens of data \citep{llama}. This training enables LLMs to comprehend and produce coherent, contextually relevant text, making them invaluable tools in a wide range of applications \citep{brown2020language}.

The rapid development of LLMs also signifies a transformative moment in the social sciences. In market research—a critical area in marketing science and operations management—understanding market demand for products or services is paramount. Traditional market research methods, such as surveys, focus groups, and conjoint analysis, often require significant resources and time. Gathering data from real subjects involves designing intricate questionnaires, recruiting participants, and analyzing responses, all of which can be both costly and logistically challenging \citep{hair2019essentials}. LLMs present a novel solution by enabling the generation of synthetic data that simulates real consumer behavior without the need for extensive fieldwork. This potential has sparked growing interest among researchers. Thus, a key advantage of using LLMs in market research is their ability to scale data generation: whereas traditional methods often limit the volume of data that can be feasibly collected due to constraints on time and resources, LLMs can rapidly generate large volumes of data, facilitating more comprehensive and granular analysis.
 
LLMs are trained on vast amounts of data from diverse sources across the internet, including reviews, purchase decisions, and even real survey data on products and services from various categories. The underlying assumption, expressed with some exaggeration, is that LLMs have been exposed to everything there is to see. Researchers hypothesize that the responses LLMs generate for market research surveys could, to some extent, reflect the types of responses that real consumers---represented in the training data---would have given to the same questions. These factors suggest that LLMs may become invaluable tools for gaining insights into consumer preferences, thanks to their ability to mimic or replicate human responses \citep{brand2023using}.
Recent studies provided encouraging evidence supporting this potential \citep{argyle2023out,chen2023emergence,horton2023large}. For instance, research showed that by fine-tuning a model \citep{gururangan2020don} or employing prompting techniques such as chain-of-thought (CoT) reasoning \citep{wei2022chain}, it is possible to generate simulated responses in market surveys that resemble those of real human subjects \citep{brand2023using,goli2024frontiers}. These explorations suggest the enormous potential of LLMs for advancing market research.

Nonetheless, recent literature also highlights several caveats. Despite some encouraging findings by \cite{goli2024frontiers}, the authors acknowledged that even the most up-to-date prompt engineering methods and advanced LLMs, such as OpenAI's Generative Pre-trained Transformer (GPT), cannot completely eliminate discrepancies between human and LLM responses in market surveys. For example, they observe that GPT-4 exhibits a pronounced level of impatience compared to human participants. This led the authors to conclude that \textit{``directly eliciting preferences using LLMs can yield misleading results,''} suggesting that LLMs should be used more as tools to facilitate hypothesis exploration rather than as direct substitutes for human responses. Similarly, \cite{gui2023challenge} pointed out that LLMs tend to make unintended assumptions about unspecified details of the choice environment, further complicating their use in market research.

We acknowledge the significant potential of LLM-generated labels for market research, but we also believe that the aforementioned gap is likely to persist, even with the most advanced future developments in AI. Consider conjoint analysis, which is arguably the most fundamental methodology for understanding consumer preferences and decision-making processes in market research \citep{green1990conjoint}. Typically, the researcher seeks to understand how consumers value different features or attributes of a product or service (for details on the setup, please refer to Section \ref{sec:model}).
At its core, customers aim to maximize their benefits by selecting the optimal products based on their past experiences. However, we emphasize that LLMs do not possess real-life experiences. Additionally, consumer preferences evolve over time due to trends, technological advancements, and economic and cultural shifts \citep{solomon2020consumer}, changes that LLMs may not accurately capture.

\begin{figure}[!t]
    \centering
    \includegraphics[width=0.85\textwidth]{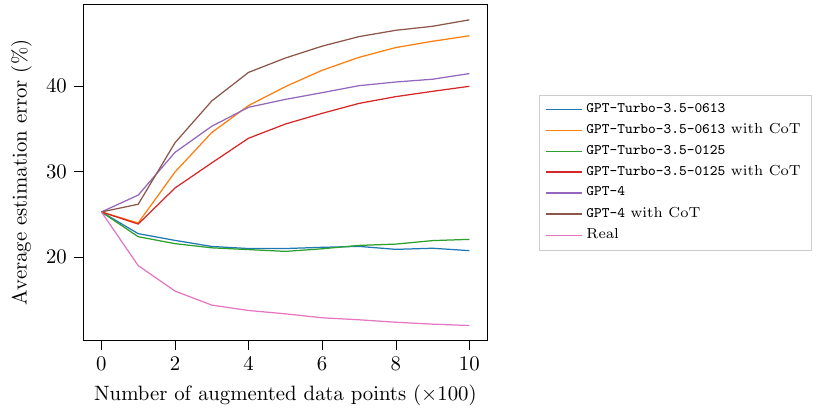}
    \caption{LLM-generated Data $\neq$ Human Data}
    \label{fig:llm-human}
    \begin{minipage}{\textwidth}
    \footnotesize
    \begin{tabular}{p{1cm} p{15cm}}
         \hspace{-1mm}\textbf{Notes}: & \setlength{\baselineskip}{1.2\baselineskip} The plot illustrates the percentage error of estimation, calculated as \(\big\lVert (\hat{\betavec} - \betavec^*)/{\betavec^*} \big\rVert_1\), where the ratio is computed component-wise. This error is based on pooling 50 real customer response data points with LLM-generated data of varying sizes. Each curve represents estimators obtained using different combinations of AI-generated or real labels. For further details, please refer to Section \ref{sec:empirical-1}. The horizontal axis indicates the quantity of data pooled into the model.
    \end{tabular}

    % \begin{itemize}[leftmargin=1cm]
    %     \item [\textbf{Notes}:] The plot shows the percentage error of estimation based on pooling the real customer response data and LLM-generated data, $\big\lVert {\big(\hat{\betavec} - \betavec^*\big)}/{\betavec^*}\big\rVert_1$, where the ratio is taken component-wise. Different curve corresponds to estimators obtained using different AI-generated or real labels. For more details, please refer
    %   to Section \ref{sec:empirical-1}. The horizontal axis gives the quantity of data pooled into the model.
    % \end{itemize}
  \end{minipage}
\end{figure}

The gap between an LLM's predictions and actual customer preferences can lead to misleading results if a researcher simply replaces real data with AI-generated data or na\"ively combines the two—a point underscored in \cite{goli2024frontiers} through careful empirical studies. This observation aligns with our findings. We conducted an empirical study on the conjoint analysis of COVID-19 vaccine preferences, pooling real data with LLM-generated responses based on the same set of product features and performing the conjoint analysis in a manner similar to \cite{brand2023using} (see Section \ref{sec:empirical-1} for details of the study).
The results are presented in Figure \ref{fig:llm-human}. 
Regardless of the version of the LLM used or whether state-of-the-art prompting techniques are applied, there is a significant gap between the ground-truth model coefficients and those estimated using pooled data. This gap does not diminish with the inclusion of more LLM-generated data; in most models, it actually worsens as more LLM-generated data is added. In contrast, as one would expect, the gap decreases when more real data is incorporated, eventually vanishing to zero, highlighting a stark contrast with the LLM-generated data.

Recognizing that AI-generated data, while informative, is inherently imperfect and likely to be misleading if used na\"ively, our goal is to develop a scheme for its efficient utilization. Although engineering techniques such as fine-tuning, CoT prompting, and in-context learning can potentially improve the quality of data from LLMs, our objective is to introduce a statistical approach that can be broadly applied independently of these methods and have theoretical guarantees. Our research question is then: \emph{Can we improve conjoint analysis by effectively integrating LLM-generated data with real data to produce more accurate estimators of customer preferences, compared to using human data, LLM-generated data, or a na\"ive combination of both?} Our method is inspired by the widely recognized transfer learning and knowledge distillation approach in the deep learning literature \citep{hinton2015distilling}. By formalizing and adapting their core principles to our context, we present a data-augmentation statistical approach for extracting value from LLM- and AI-generated conjoint data. Specifically, we make two key contributions.

\vspace{1mm}
\noindent
\revise{\underline{\textit{Novel Theoretical Framework for Data Augmentation.}} 
We develop a general framework for statistical data augmentation using generative models, such as LLMs, to capture human preferences in conjoint analysis, and propose the AI-Augmented Estimator
(AAE). The framework is broadly applicable to hybrid datasets where true human labels (ground truth) are available for only a subset of data points, while (AI-)generated labels can be created for all observations. These AI-generated labels may be produced by a traditional machine learning model or a generative model such as an LLM. Although our framework is compatible with any source of surrogate labels, results presented in Section \ref{sec:empirical-1} and Appendix \ref{app:comparison} show that the estimator performs best when the AI-generated data are more informative. This finding suggests that, at the time of writing, the most effective application of our framework is to augment data generated by LLMs.}

\revise{
Conceptually, our approach leverages the joint distribution among features, surrogate labels, and true labels within the subset of data where both types of labels are observed. This joint structure allows us to learn the distributional discrepancy between surrogate and true labels and subsequently regularize the surrogate labels so that they more closely approximate the true ones. By doing so, we effectively expand the labeled dataset, enabling more accurate estimation of human preferences. These principles give rise to theoretically grounded estimators: our estimator is consistent and asymptotically normal (Theorem \ref{thm:consistency}) and satisfies a finite-sample, high-probability performance bound (Theorem \ref{thm:aae-bound}). Furthermore, we demonstrate that our method outperforms approaches that rely solely on human data, solely on LLM-generated data, or on naïve combinations of the two.}

\vspace{1mm}
\noindent
\revise{
\underline{\textit{Large-scale 
 empirical validation.}} We conduct extensive empirical analyses across multiple conjoint settings to validate the effectiveness and robustness of our proposed AAE. In the first empirical study, based on a real-world COVID-19 vaccination dataset, AAE consistently reduces estimation error relative to naïve augmentation and AI-only baselines, even when GPT-generated labels are imperfect. Its regularization capability remains robust across different LLM versions and prompting strategies—basic, CoT, few-shot, and fine-tuned—with GPT-4o and CoT prompting achieving the greatest gains. The results confirm that combining LLM-generated and human data through a statistically grounded augmentation framework yields reliable, unbiased preference estimates while reducing survey data requirements by 24.9\% to 79.8\%. In Section 6, we extend these findings to a separate sports-car choice dataset and additional robustness checks. AAE again outperforms alternative estimators across all models and prompts, outperforming even fine-tuned GPT-4o baselines. Further comparisons show that AAE matches  prediction-powered inference (PPI), an alternative data augmentation strategy
 \cite[see the literature review and ][]{angelopoulos2023prediction, angelopoulos2023ppi++} in low-dimensional cases but surpasses it in high-dimensional conjoint features due to its more granular correction of AI-human discrepancies.
 % Finally, under model misspecification and nonparametric estimation settings, AAE continues to exhibit strong stability and accuracy. 
 Collectively, these empirical results demonstrate that AAE offers a scalable, bias-controlled, and cost-efficient solution for integrating LLM-generated data into real-world market research.}

We briefly discuss the organization of our paper. In Section \ref{sec:literature}, we present a literature review.
In Section \ref{sec:model}, we set up our model and formally state our data-augmentation approach. We discuss the theoretical properties of our proposed approach in Section \ref{sec:consistency-normality} and two empirical studies in Sections \ref{sec:empirical-1} and \ref{sec:empirical-2}. In Section \ref{sec:conclusion}, we conclude. 

%% file: src/2_reference.tex
\section{Literature}
\label{sec:literature}

In this section, we review several key streams of literature relevant to our research. 

\vspace{1mm}
\noindent
\underline{\textit{Technical development of natural language processing (NLP) and LLM}.}
The development of LLMs represents a culmination of decades of advances in NLP and deep learning techniques. The foundational work of \cite{chomsky1956three} introduced formal models for language description, paving the way for computational approaches to linguistic analysis. Modern LLMs owe much to significant technical breakthroughs in recent years, such as the introduction of sequence-to-sequence models \citep{sutskever2014sequence} and the transformer architecture \citep{vaswani2017attention}, which enables efficient handling of sequential data and contextual understanding.

Key milestones in this area include the development of BERT \citep{devlin2018bert}, which introduced bidirectional context modeling, and GPT \citep{radford2018improving}, which leveraged generative pre-training to advance NLP capabilities. These innovations culminated in LLMs like GPT-4 \citep{openai2023gpt}, characterized by their ability to perform complex language tasks after training on massive datasets.
Importantly, further improvements, such as domain-specific adaptations as effective strategies for fine-tuning \citep[e.g., see][]{beltagy2019scibert, gururangan2020don, parthasarathy2024ultimate} and few-shot learning \citep{brown2020language} enable these models to achieve remarkable performance in specific tasks.
We refer the readers to \cite{naveed2023comprehensive}
for a review of the broader trajectory of LLM research. These technical advancements underpin the transformative potential of LLMs in applications such as social science and consumer behavior modeling.

\vspace{1mm}
\noindent
\noindent
\underline{\textit{Using LLMs for social sciences and market research}.}
AI, especially LLMs, is making a pronounced impact on social sciences. Researchers have explored how these models can simulate human behavior and provide insights into societal phenomena. For instance, \cite{ziems2024can} discussed the potential of LLMs to transform computational social science by generating synthetic data and modeling complex social dynamics. However, questions remain about the reliability of these models in replicating real-world behavior, as highlighted by \cite{huang2024social} and \cite{yang2024large}, who cautioned against over-reliance on LLMs for social predictions due to their lack of lived experiences and contextual nuances. Similarly, \cite{gui2023challenge} pointed out challenges in using LLMs to simulate human behavior, especially from a causal inference perspective, where the models may fail to account for underlying causality in decision-making processes.

In the domain of market research, LLMs have emerged as powerful tools for generating synthetic data that mimics human responses. Studies such as \cite{brand2023using} and \cite{argyle2023out} demonstrated the feasibility of using LLMs for simulating consumer preferences, while \cite{chen2023emergence} and \cite{horton2023large} highlighted the potential of these models as simulated economic agents. Still, the extent to which LLMs can faithfully replicate human preferences remains a point of debate, as explored by \cite{goli2024frontiers}, who emphasized the challenges in achieving alignment between AI-generated and human data.
Our work provides an alternative perspective on this question: while LLMs may not fully replicate human behavior or replace humans in social sciences or market research, they can nonetheless offer valuable insights to improve existing methods. With the implementation of robust statistical procedures for correcting bias, LLMs can serve as effective tools for enhancing the accuracy and efficiency of current methodologies.

\vspace{1mm}
\noindent
\noindent
\underline{\textit{Conjoint analysis}.}
Conjoint analysis has long been a cornerstone of market research, providing insights into consumer preferences by evaluating trade-offs between product attributes \citep{green1978conjoint, green1990conjoint}. Over time, this methodology has seen significant methodological advancements. On the decision-making side, early work by \cite{kohli1990heuristics} and \cite{wang2009branch} explored heuristics and optimization approaches for conjoint-based product line design, while \cite{kessels2008optimal} focused on optimizing experimental design for conjoint studies. Recent contributions, such as \cite{dzyabura2018offline}, extend the conjoint analysis to multi-channel settings, incorporating both online and offline consumer behaviors.

Our research builds on the rich tradition of integrating methods from statistics or ML into conjoint analysis. While prior studies have leveraged advanced statistical models \citep{allenby2006hierarchical} and Bayesian frameworks \citep{eggers2021choice}, we introduce a novel data augmentation approach that bridges the gap between AI-generated and real-world data. Furthermore, as mentioned, \cite{goli2024frontiers} emphasized the limitations of using LLM-generated data alone for preference estimation, underscoring the need for innovative solutions like our proposed methodology. This methodological contribution represents a significant step forward in enhancing the robustness and scalability of conjoint analysis.

\vspace{1mm}
\noindent
\revise{
\underline{Prediction-Powered Inference (PPI).}
Notably, our work also connects to PPI literature. PPI provides a framework for combining a small labeled dataset with a large pool of ML predictions to construct unbiased, valid confidence intervals that shrink as prediction quality improves \citep{angelopoulos2023prediction}. PPI++ \cite{angelopoulos2023ppi++} builds on this by replacing computationally intensive test-based constructions with simple Wald-type intervals, introducing a power-tuning mechanism that automatically adapts to prediction quality, and ensuring efficiency gains without risk of degradation when predictions are weak. Most recently, \cite{ji2025surrogates} reinterpret PPI through the surrogate-outcome literature, proposing RePPI, which recalibrates the imputed loss with flexible models to achieve efficiency-optimal inference under convex settings, dominating PPI and PPI++ when recalibration is consistent and still never worse than using labels alone. Together, the three works establish a coherent progression: PPI for validity, PPI++ for computational and adaptive efficiency, and RePPI for optimal use of predictions as modern surrogates.}

\revise{
At a high level, AAE and PPI share the same guiding principle: use a small dataset containing both real and AI labels to learn the discrepancy between them, and then apply this correction to a larger dataset containing only AI labels. 
However, we note two key differences. 
The key difference lies in how the two methods correct the discrepancy. PPI operates through a rectified loss function, subtracting the empirical log-likelihood difference from the primary set from the empirical log-likelihood of the auxiliary set. However, log-likelihood functions tend to flatten in high-dimensional feature spaces, which can lead to information loss when adjusting the aggregated loss.
AAE, on the other hand, directly models the discrepancy between AI and real data as a function of the feature vector $x$ and AI label $z$. It corrects AI labels directly by replacing them with predictions of the real labels given the AI label and product features, enabling a more granular adjustment. As shown in our empirical findings in Section 6.3: AAE performs similarly to PPI when $x$ is low-dimensional, but outperforms PPI in higher-dimensional settings.}

\vspace{1mm}
\noindent
\underline{\textit{Technology, Innovation, and Entrepreneurship}.}
Our work also speaks to the literature of technology, innovation, and entrepreneurship, the intersection of which has been a focal point of operations management research. Foundational works such as \cite{shane200450th} provide a comprehensive overview of how technological advancements drive product development and entrepreneurial activities. Recent studies emphasize the role of AI and ML in reshaping operational tasks. For instance, \cite{terwiesch2019om} and \cite{bastani2022applied} highlight the potential of AI tools to optimize decision-making processes, while \cite{yoo2024next} identify new frontiers in digital innovation.

Our work contributes to this stream by demonstrating how LLMs represent a disruptive innovation in market research. By democratizing access to market insights, LLMs enable startups and smaller firms to conduct sophisticated analyses previously reserved for resource-intensive operations \citep{olsen2020industry, choi2022disruptive}. Furthermore, our findings align with \cite{girotra2023ideas}, which discusses the transformative role of LLMs in idea generation and business innovation.
Along a similar line, \cite{connell2024estimating} discusses how to correct the misclassification error in using LLM in place of human for classifying textual data, and \cite{ludwig2024large} develops an econometric framework to assess the validity of using LLMs in economics research, emphasizing the need to avoid training data leakage for prediction tasks and to ensure gold-standard measurement equivalence for estimation tasks, while providing guidance on mitigating potential biases and limitations. 
Collectively, these insights underscore the potential of new technologies to revolutionize traditional operational practices, particularly for data collection and consumer analysis.

%% file: src/3_model.tex
\section{Model and the Data-augmentation Approach}
\label{sec:model}

In this section, we establish the theoretical framework. We focus on conjoint analysis as the primary use case to present the theory and conduct the empirical analysis detailed in Section \ref{sec:empirical-1}.

\subsection{Setup}
\label{sec:setup}

We first provide a detailed explanation of the setup, starting with an in-depth discussion of the theoretical framework underlying the data generation process. We will outline the assumptions, structure, and specific parameters that guide this process, building a foundation for understanding how the data is conceptualized and modeled within this framework.

\subsubsection{Data Generation Process and the Best-in-class Estimation.}
\begin{figure}[h!]
    \centering
    \includegraphics[width=165mm]{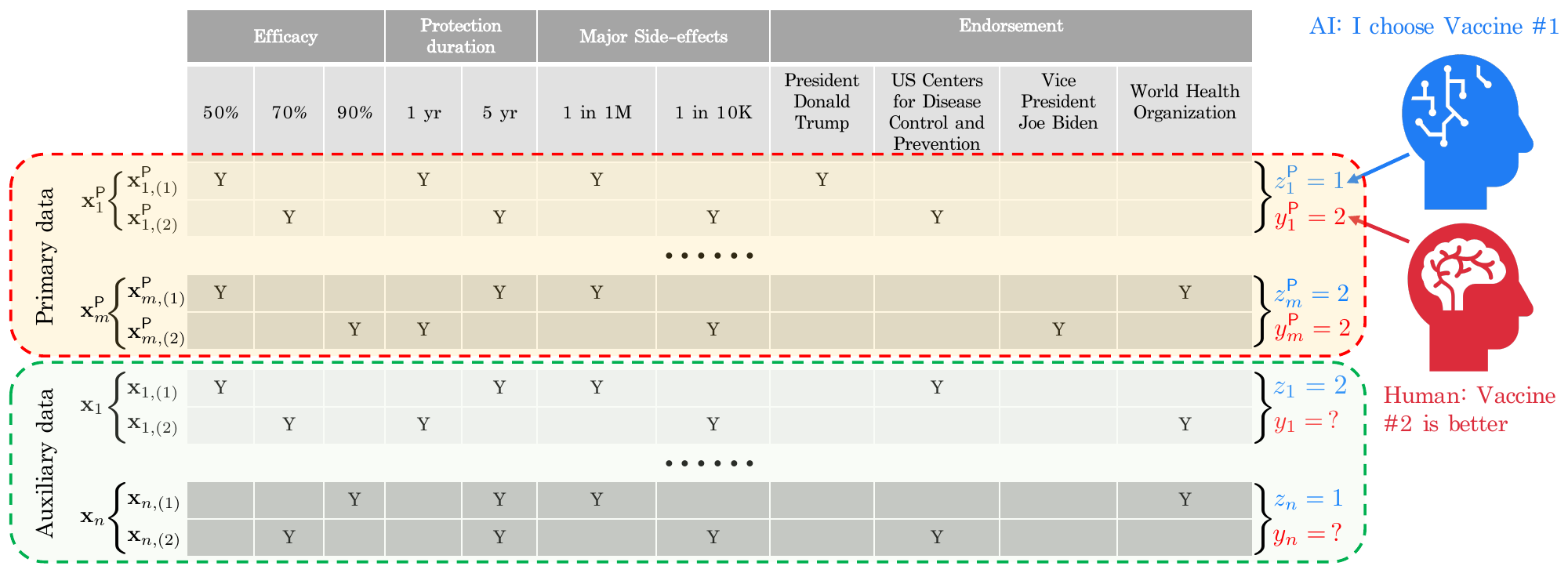}
    \caption{Illustration of the dataset}
    \label{fig:illustration-data}
    \begin{minipage}{\textwidth}
    \footnotesize
    \begin{tabular}{p{1cm} p{15cm}}
         \hspace{-1mm}\textbf{Notes}: & \setlength{\baselineskip}{1.2\baselineskip} In this figure, we illustrate the two datasets, using an example simplified from the empirical setting in Section \ref{sec:empirical-1}. 
         For each data point in the primary data of size $m$, two vaccine options are presented to both the human respondent and the AI. 
         Corresponding to the first data point, the first vaccine has the following features: 50\% efficacy, a one-year protection duration, a major side-effect rate of one in a million, a minor side-effect rate of one in ten, and endorsement by President Donald Trump. These features can be converted into a numerical feature vector, $\x^{\sf P}_{1, (1)}$, suitable for conjoint analysis, for example, using one-hot encoding. 
         Note that the options chosen by the human subject and AI may agree with each other (e.g., in the last data point) or differ from each other (e.g., in the first data point). 
         The auxiliary data, which is of size $n$, is different from the primary data in that we only have the label solicited from the AI. 
    \end{tabular}
    
    % \begin{itemize}[leftmargin=0pt]
    %     \item [\textbf{Notes}:] In this figure, we illustrate the primary dataset using an example simplified from the empirical setting in Section \ref{sec:empirical-1}. For instance, corresponding to the first data point, two vaccine options are presented to the human respondent. The first vaccine has the following features: 50\% efficacy, a one-year protection duration, a major side-effect rate of one in a million, a minor side-effect rate of one in ten, and endorsement by President Donald Trump. These features can be converted into a numerical feature vector, $\x^{\sf P}_{1, (1)}$, suitable for conjoint analysis, for example, using one-hot encoding. Based on these features, the AI recommends selecting the first vaccine (i.e., $z^{\sf P}_{1} = 1$), which aligns with the human respondent’s choice (i.e., $y^{\sf P}_{1} = 1$). Note that the second human respondent opts to select no vaccine from the two offered. The auxiliary dataset is similar to the primary dataset, except that it lacks the human-generated labels for these data points.
    % \end{itemize}
  \end{minipage}
\end{figure}

Consider a setting where we observe two datasets. The first, referred to as the \textit{primary data}, consists of $m$ data points, 
$\left\{\big(\x^{\sf P}_{i}, y^{\sf P}_{i}, z^{\sf P}_{i}\big)\right\}_{i = 1}^m$.
For a fixed positive integer $k$, we denote the set $\{1,\dots, k\}$ by $\Kscr$ and define $\Kscr^+ = \Kscr \cup \{0\}$. 
\revise{These data points are independently and identically distributed (i.i.d.) according to the same distribution as the random vector $(\x, y, z)$, where $y, z \in \Kscr^+$ and $\x \in \Real^{dk}$.
To clarify, in the context of conjoint analysis with AI-generated labels, $\x$ represents the features or context of a choice setting, and $y$ denotes the respondent's choice. Here, $\Kscr^+$ corresponds to the $k+1$ options available to the respondent, where $0$ represents the outside option. Additionally, $\x = (\x^{\top}_{(1)}, \x^{\top}_{(2)}, \dots, \x^{\top}_{(k)})^{\top}$, where $\x^{\top}_{(j)} \in \Real^d$ corresponds to the features of option $j$ for each $j \in \Kscr$.} Throughout this paper, we refer to the random variable $y$ as the \textit{real label}. 
The variable $z$ represents a predicted label generated by an AI model based on $\x$, and we refer to it as the \textit{AI-generated label}. 
The second dataset, called the \textit{auxiliary data}, consists of $\left\{\big(\x_{i}, z_{i}\big)\right\}_{i = 1}^n$, which are independent of the primary data but follow the same distribution. However, in the auxiliary data, the true label $y_i$ is missing for each data point. 
In Figure \ref{fig:illustration-data}, we illustrate this theoretical setup using an example simplified from the empirical setting in Section \ref{sec:empirical-1}. 
\revise{In contrast, in typical conjoint analysis settings, the AI-generated labels $z$ is not available, and 
our goal is to fit the vector $\betavec  \in \Real^{d}$ to capture the choice probabilities as accurate as possible using  Multinomial Logit (MNL) models, i.e., 
$\betavec$ such that 
\begin{align}
\label{eqn:mnl-conjoint}
    \sigma_j(
\x;\betavec 
)~=~ \frac{e^{\x^{\top}_{(j)}\betavec}}{1+\sum_{\ell \in \Kscr} e^{\x^{\top}_{(\ell)}\betavec}},~~\forall\, j \in \Kscr, 
\end{align}
best approximates $\Prob(y\,|\,x)$.
}

How is this data generation process implemented in practice? A manager facing a conjoint study problem can begin by collecting a small sample of human response data. Using the same questions presented to human respondents, the manager can query an LLM to generate the primary dataset containing both labels $y$ and $z$. Subsequently, the manager can generate additional responses by querying the LLM with a different set of choices, which forms the auxiliary dataset where the true label $y$ is missing. Since the cost of querying an LLM is negligible compared to recruiting human subjects in many cases, the auxiliary data is typically much larger than the primary data in size.

We assume that the distribution $\Prob(y\,|\,\x, z)$ can be parameterized as $\Prob(y = j\,|\,\x, z)\,=\,g_j(\x, z; \thetavec^*)$ for each $j \in \Kscr^+$, where $\thetavec^* \in \Real^q$ is not known to us. 
This assumption is motivated by the fact that modern artificial AI models, such as an LLM, are trained with vast amounts of text data, including information from numerous sources on the internet, that may contain product reviews, messaging boards, and other online forums with contributions from a wide range of consumers discussing the products they shop for and purchase \citep{brand2023using}.
Therefore, AI can ``mimic" how respondents make choices, 
leading to highly informative AI-generated labels. Therefore, the model $\big(g_j(\x, z; \thetavec^*):j\in \Kscr^+\big)$, which can be chosen as the MNL model or a neural network, 
is specified to capture the conditional probabilities $\Prob(y\,|\,\x, z)$ in practical scenarios.
We also assume throughout that $|g_j(\cdot)| \leq 1$ for all 
$j\in \Kscr^+$, which is reasonable for a wide range of models. 

\revise{
In this work, we frame the main goal of our conjoint analysis as identifying the \textit{best-in-class} estimator of \eqref{eqn:best-in-class}—the one that most closely approximates the true choice probabilities within the set of all MNL models.}\footnote{Note that we do not assume that the data is generated by a MNL model. We recognize that choice probabilities parameterized by the MNL model may be misspecified. In practice, customer choice probabilities can exhibit considerable complexity, yet researchers and practitioners often rely on the simpler MNL model as an approximation in conjoint analyses.} The quality of this approximation is measured by the Kullback–Leibler (KL) divergence. Formally, the best-in-class estimator is defined as follows:
% For each fixed $\betavec \in \Real^{d}$, we use the Kullback–Leibler divergence (KL-divergence) to measure the quality of the MNL model parameterized by $\betavec$. 
% We define 
\begin{align}
\label{eqn:best-in-class}
    \betavec^* ~\in~
    \argmin_{\betavec \in \Real^{d}}
    \left\{
    \E_{\x}\Big[
    {\sf KL}\big( \Prob(y| \x)\,\big\vert\,\sigma_{y}(\x, \betavec) \big)
    \Big]~=~
    \E_{\x}\left[
    \sum_{j\in \Kscr^+} \Prob\big(y = j| \x\big)
    \log\left(
    \frac{\Prob\big(y = j| \x\big)}{\sigma_{j}(\x; \betavec)}
    \right)
    \right]\right\}.
\end{align}
That is, measuring the loss of approximation using the KL-divergence,  
$\big(\sigma_j(\x; \betavec^*):j\in \Kscr^+\big)$ gives us the approximation to the true choice probabilities that is best in the class of MNL models.  
Under reasonable assumptions, this minimizer exists and is uniquely defined (see Theorem \ref{thm:consistency} below). 

We further clarify the distinction between our problem and the measurement error issues discussed in the econometrics literature \citep[e.g., see][]{bound2001measurement} and in recent work by \cite{connell2024estimating}. Specifically, in our context, conditional on $\x$, $z$ should not be interpreted as a noisy realization of $y$, as is commonly assumed in the measurement error framework. In our setting, $y \perp z \,|\, \x$ because the AI cannot access the internal thought and decision-making processes of a human subject, and any relationship between $y$ and $z$ is mediated entirely through $\x$. Nonetheless, we emphasize that our estimation procedure and theoretical results impose no assumptions on the joint distribution of $y$ and $z$ given $\x$. Therefore, even if $z$ indeed represents $y$ measured with error, our approach remains valid.

\subsection{Primary, Auxiliary and Na\"ive Estimators} 
\label{sec:3-estimators}
With no access to auxiliary data such as AI-generated labels, the default approach is to fit the parameter $\betavec$ using only the primary data with its real labels $\left\{\big(\x^{\sf P}_{i}, y^{\sf P}_{i}\big)\right\}_{i = 1}^m$ using the standard Maximum Likelihood Estimation (MLE). We use $\hat{\betavec}^{\sf P}$ to denote the estimator obtained using only the primary data. When the primary data size is sufficiently large, $\hat{\betavec}^{\sf P}$ can be close to $\betavec^*$. However, the size of the primary data is often restricted due to the costs of recruiting real subjects. A small primary set yields a \textit{inaccurate} estimator that is far away from $\betavec^*$. 
On the flipping side, we may directly perform MLE on the auxiliary data $\left\{\big(\x_{i}, z_{i}\big)\right\}_{i = 1}^n$, using $z_{i}$ as labels. 
Let us define this estimator as $\hat{\betavec}^{\sf A}$. Given the distribution of LLM-generated data and human data are different (i.e., the distributions of $ y $ and $ z $, conditioned on $ \x $, are different), it is easy to recognize that $\hat{\betavec}^{\sf A}$ could be severely biased regardless of the size of auxiliary dataset compared to $\betavec^*$.

Due to this limitation, people intend to utilize auxiliary data and AI-generated labels to facilitate the estimation of $\betavec^*$. A natural idea for utilizing the AI labels is to simply pool the primary and auxiliary sets. In particular, we define
$$
(\tilde{\x}_{i}, \tilde{y}_i)~=~
\begin{cases}
    ({\x}_{i}^{\sf P}, y_{i}^{\sf P})~&\text{if } 1 \leq i \leq m \\
    ({\x}_{i-m}, z_{i-m})~&\text{if } m+1 \leq i \leq m+n
\end{cases},
$$
and perform the standard MLE on $(\tilde{\x}_{i}, \tilde{y}_i)_{i=1}^{m+n}$.
We call this approach the \textit{na\"ive augmentation} and write the estimator as $\hat{\betavec}^{\sf Naive}$. The auxiliary data largely enlarges the data size. However, since $z$ and $y$ in general come from different distributions, $\hat{\betavec}^{\sf Naive}$ again can be biased, even with large data size. Note that both $\hat{\betavec}^{\sf P}$ and $\hat{\betavec}^{\sf A}$ are both special cases of $\hat{\betavec}^{\sf Naive}$ by setting $n = 0$ or $m = 0$. In the following, we present a simple example to demonstrate the limitations of these estimators discussed above and summarize the observation in Proposition~\ref{prop:bias}.

\begin{examp}[AI-Generated Data Cannot Replace Human Data]
\label{examp:ai-bias}
{\rm 
Consider a simple setting with 
only one product and no features.
Conditional on $z = j$, where $j \in \Kscr^+ = \{0,1\}$, $y = j$ with probability $p$.
Suppose that $\Prob(z = 1) = \alpha$. 
For clarity, we display the joint distribution in Table \ref{tab:joint-y-z}.

\begin{table}
\renewcommand{\arraystretch}{1.5}
\centering
{\footnotesize
\caption{Joint distribution of $y$ and $z$ in Example \ref{examp:ai-bias}}
\label{tab:joint-y-z}
\begin{tabular}{  c |c  c }
\hline
\hline
  & $y=1$ & $y=0$ \\ \hline
 $z=1~$ & $~\alpha p~$ & $~\alpha (1-p)~$ \\  \hline
 $z=0~$ & $~(1-\alpha) p~$ & $~(1-\alpha) (1-p)~$
 \\\hline\hline
\end{tabular}
}
\end{table}

Therefore, the probability that $y = 1$ is given by $c_1(\alpha, p):=\alpha p +(1 - \alpha)(1 - p)$ and clearly the best-in-class estimator in \eqref{eqn:mnl-conjoint} is given by $\beta^* = \log\left(\frac{c_1(\alpha, p)}{1- c_1(\alpha, p)}\right)$, which is computed through
$
\frac{e^{\beta^*}}{1 + e^{\beta^*}} = \Prob(y = 1) = c_1(\alpha, p)
$.
Suppose that $n, m \rightarrow \infty$ and $n/m \rightarrow \rho \neq 0$. 
Asymptotically, the na\"ive estimator satisfies 
$$
c_2(\alpha, p):=\frac{\alpha \rho}{1 + \rho}+ 
\frac{\alpha p +(1 - \alpha)(1 - p)}{1 + \rho}~\approx~\frac{e^{\hat{\beta}^{\sf Naive}}}{1 + e^{\hat{\beta}^{\sf Naive}}},
$$
or equivalently 
$
\hat{\beta}^{\sf Naive}~\rightarrow~ \log\left(
\frac{c_2(\alpha, p)}{1-c_2(\alpha, p)}
\right).
$
One can verify that unless $\alpha = 1/2$ or $p = 1$,  this limit is not equal to $\beta^*$. 
This is always the case even when $n,m\rightarrow \infty$, as long as $\rho \neq 0$. 
Also note that even when $p$ is close to, but different from, one, $z$ is very informative of $y$ but the na\"ive methods still cannot lead to desired estimates. }
\end{examp}

\begin{prop}[Bias of $\hat{\betavec}^{\sf A}$ and $\hat{\betavec}^{\sf Naive}$.]
\label{prop:bias}
When $m, n \rightarrow \infty$, in general, 
    $\hat{\betavec}^{\sf A}$ and $\hat{\betavec}^{\sf Naive}$ are not consistent estimators of ${\betavec}^*$.
\end{prop}

\subsection{Estimation with AI-augmented Data}
\label{sec:estimation}
These challenges lead to our research question: \textit{How do we extract value from the AI-augmented data to facilitate the estimation of $\betavec^*$?} To answer this question, we propose the following data augmentation approach that allows us to use the AI-generated data to fit the model.\footnote{For simplicity, here we assume that an optimizer $\hat{\betavec}^{\sf AAE}$ is obtained. 
Otherwise, we may optimize approximately and alternatively choose 
$\hat{\betavec}^{\sf AAE}$ such that $
\widehat{Q}(\hat{\thetavec};\hat{\betavec}^{\sf AAE}) = \sup_{\betavec \in \Real^{d}} \widehat{Q}(\hat{\thetavec};\betavec) - o_{\text{P}}(1)
$. 
In this case, our key theoretical result, Theorem \ref{thm:consistency}, is still valid with a slight adaption of the analysis. Interested readers may refer to the proof of Theorem \ref{thm:consistency} in the appendix.}

\vspace{10pt}

\begin{mdframed} \footnotesize \underline{{\sc Estimation with AI-Augmented Data}} 

\noindent \textbf{Step 1.} 
Obtain an estimator $\hat{\thetavec}$ to $\thetavec^*$, where $\Prob(y = j\,|\,\x, z)\,=\,g_j(\x, z; \thetavec^*)$, using the primary data.

\noindent \textbf{Step 2.} With the auxiliary data, we construct the estimator $\hat{\betavec}^{\sf AAE}$ as 
\begin{align}
\label{eqn:AAE-def}
    \hat{\betavec}^{\sf AAE} ~\in~\argmax_{\betavec \in \Real^{d}}\,\left\{ \widehat{Q}(\hat{\thetavec};\betavec)\,=\, \frac{1}{n}\sum_{i=1}^n\sum_{j\in \Kscr^+}g_j(\x_i, z_i; \hat{\thetavec})\log \sigma_j(\x; \betavec)\right\}.
\end{align}
\end{mdframed}

\vspace{10pt}

We refer to $\hat{\betavec}^{\sf AAE}$ as the \textit{AI-augmented estimator} (AAE). 
In the next section, we demonstrate that our estimator $ \hat{\betavec}^{\sf AAE} $ successfully recovers $ \betavec^* $ and exhibits the desired asymptotic properties. The key idea in our approach is to use the primary data---which contains both LLM-generated and human-produced data---to learn an efficient mapping between human and LLM-generated data. We then leverage this mapping to construct the estimator. This strategy closely resembles a transfer learning approach, where knowledge from one domain (in our case, the LLM-generated data) is transferred to another domain (the human-produced data). The fundamental assumption underpinning such transfer learning techniques is that learning the mapping function between two domains, denoted by $g(\cdot)$ in our context, is inherently simpler and requires fewer data points than carrying out the individual learning tasks within either domain on its own. In other words, in our context, the AI model undertakes the ``heavy lifting'' of capturing human choice behavior in a fuzzing way, and the first stage in our estimation is trying to capture how fuzzy this AI model is compared to the human model. Moreover, the mapping will become easier to learn with lower LLM costs and higher LLM quality. This is because, as the accuracy and the size of AI-generated labels improves, we anticipate a high-quality estimation of $ \thetavec$ due to a decrease in the asymptotic variance of $ \thetavec$ estimation; see Section \ref{sec:AAE} and Appendix \ref{app:comparison} for further discussions.

\revise{
We highlight the intuition of our method from two alternative angles. 
First, our approach is motivated by principles from transfer learning and knowledge distillation, though adapted to the context of conjoint analysis. In transfer learning, knowledge from one domain is leveraged to reduce data needs in another \citep{pan2009survey,zhuang2020comprehensive}; similarly, our method transfers information contained in the joint distribution of product features and LLM-generated labels ($\gvec(\x, z; \thetavec^*)$) to improve inference on human preferences. Likewise, while classical knowledge distillation involves a student model mimicking a teacher \citep{hinton2015distilling}, our framework echoes this idea by learning the conditional probability of real labels given LLM labels and product features. Although our method does not involve parameter reuse across tasks or a teacher–student architecture in the narrow sense, the underlying principle is the same: using a large, complex source of information (the LLM) to guide a simpler, statistically grounded estimator.
Such a principle can be used to integrate AI-generated labels in other settings, e.g., a continuous target in a regression problem (see discussions in Section \ref{sec:additional}
and Appendix \ref{app:regression}).
}

\revise{Second, while LLMs encode vast information, their outputs are systematically different from human responses. Importantly, this discrepancy is often easier to learn from small samples than estimating preferences directly. Our method builds on this intuition: we use a small human dataset to capture the gap between human and LLM labels, and then apply this correction to the much larger LLM-generated dataset. In the AAE algorithm, the first step models the discrepancy between AI and human labels conditional on product features $\x$. The second step then adjusts estimation using the predicted human label based on the AI label and features. When the first-stage model is sufficiently accurate, the second-stage estimator reduces variance by leveraging the enlarged AI-labeled dataset (Proposition \ref{prop:weak-dom}). Thus, the more precise the first-stage predictions, the greater the bias reduction in the second stage, as formalized in Theorem \ref{thm:aae-bound}.}

%% file: src/4_theory.tex
\section{Theoretical Properties of the AAE Estimator}
\label{sec:consistency-normality}

This section establishes the large-sample and finite-sample properties of our estimator, $\hat{\betavec}^{\sf AAE}$.
\revise{We first show that it is consistent and asymptotically normal under mild conditions (Theorem \ref{thm:consistency}), extending classical extremum-estimator arguments to the two-stage setting with AI-generated data.
We then compare its efficiency with existing estimators. Unlike naïve approaches that are generally biased, AAE is asymptotically unbiased and achieves variance reduction. In particular, Proposition \ref{prop:weak-dom} shows that the variance of AAE weakly—and often strictly—dominates that of the plug-in estimator $\hat{\betavec}^{\sf P}$, especially with informative auxiliary data.
Finally, we provide a finite-sample high-probability bound (Theorem \ref{thm:aae-bound}), which highlights the dependence on first-stage estimation error. In the logistic regression case, we derive a new KL-based risk bound (Proposition \ref{prop:kl}). Overall, AAE enjoys both classical asymptotic guarantees and strong finite-sample performance.}

\subsection{Asymptotic Analysis}
\label{sec:main-results}

To deliver the analysis, a set of common regularity assumptions are needed. The first set of assumptions is used in the analysis of the consistency of $\hat{\betavec}^{\sf AAE}$.

\begin{assm}[Regularity Conditions for Consistency] 
\label{assumption:regularity-consistency}
The following assumptions hold. 
\begin{itemize}
\setlength\itemsep{-0.3em}
    \item [$(i)$]
    We assume that
    $\x \in \Xscr$ with probability one, where $\Xscr$ is bounded. Also, 
    $\E_{\x}[\x \x^{\top}] \succ 0$ and $\Prob(y\,|\,\x) \geq c$ for all $\x \in \Xscr$, $y \in \Kscr$ and some constant $c > 0$.
    \item [$(ii)$] $\hat{\thetavec} \pconverge \thetavec^*$. 
    \item [$(iii)$] $\big(g_j(\x, z; \thetavec):j\in \Kscr^+\big)$ is continuous in $\thetavec$ for all $\x\in \Xscr$ and $z \in \Kscr^+$.
\end{itemize}    
\end{assm}

Comparing to those assumptions required in classical consistency results, such as Theorem 5.1 in \cite{van2000asymptotic} or Theorem 2.1 in \cite{newey1994large}, Assumption \ref{assumption:regularity-consistency} presents similar or even weaker regularity conditions that guarantee consistency in a setting that significantly extends that of classical analysis of MLE under the MNL model, by exploiting the concavity in the problem structure. Part $(i)$ in Assumption \ref{assumption:regularity-consistency} essentially serves as an identification condition, which should be satisfied in practical scenarios. Part $(ii)$ states that the first-step estimator needs to be consistent, which usually is satisfied as long as the models $\gvec(\cdot) := \big(g_j(\x, z; \thetavec^*):j\in \Kscr^+\big)$ are appropriately specified and reasonable estimation strategy such as the MLE is adopted. Part $(iii)$ is a regularity condition on the functional smoothness. This assumption holds for many widely used models, for example, the MNL model or the neural networks.

Next, the second set of assumptions helps establish asymptotic normality. 

\begin{assm}[Regularity Conditions for Asymptotical Normality] 
\label{assumption:regularity-normality}
We assume that the following conditions hold. 
\begin{itemize}
\setlength\itemsep{-0.3em}
    \item [$(i)$] $\sqrt{m}(\hat{\thetavec} - \thetavec^*)\rightsquigarrow N(\Ovec, \Lambdavec)$.
    \item [$(ii)$] $\frac{n}{m}\rightarrow \rho$, where $0 < \rho < \infty$. 
    \item [$(iii)$] For each $j\in \Kscr$, $g_j(\x, z; \thetavec)$ is differentiable for all $\theta$, $\x \in \Xscr$ and $z\in \Kscr$ and there exists an open neighborhood $\Nscr$ of $\thetavec^*$ such that $\,\E_{\x,z}\big[\sup_{\thetavec \subset \Nscr}\lVert \nabla_{\thetavec} g_j(\x, z; \thetavec) \rVert_2\big] \,<\, \infty$.  
\end{itemize}    
\end{assm}

\revise{Assumption \ref{assumption:regularity-normality} is again mild. The first item states that $\hat{\thetavec}$ is asymptotically normal, which is expected.} The second item maintains that the ratio between the sample sizes of two data sets needs to be reasonable. The last item is a smoothness condition. Furthermore, to simplify the presentation, some definitions are necessary. We first define 
$\Avec(\x; \betavec) \in \Real^{k \times k}$ with 
\begin{align}
\label{eqn:A-def}
    \Avec(\x; \betavec)~:=~ \sum_{j \in \Kscr}\sigma_{j}(\x; \betavec)\left(
\x_{(j)} \x_{(j)}^{\top} - 
\sum_{j' \in \Kscr} \sigma_{j'}(\x; \betavec)\x_{(j)} \x_{(j')}^{\top}
\right)
\end{align}
and $\Omegavec \in  \Real^{k \times k}$ and $\Gammavec \in \Real^{k \times q}$ with 
\begin{align}
\label{eqn:Omega-Gamma}
    \Omegavec := \E_{\x}\left[\Avec(\x;\betavec^*) \right] ~\text{ and }~ \Gammavec := \E_{\x,z}\left[\sum_{j \in \Kscr^+} \left(\ind_{\{j\neq 0\}}\x_{(j)} - \sum_{j' \in \Kscr}\sigma_{j'}(\x, \betavec^*)\x_{(j')}\right)\nabla_{\thetavec} g_j(\x, z, \thetavec^*)\right].
\end{align}
Further, we let 
$$
\Jvec ~:=~ \E_{\x,z}\left[
\left(\sum_{j\in \Kscr}\big(g_j(\x, z; {\thetavec}^*) - \sigma_{j}(\x; \betavec^*)\big)\x_{(j)}\right)
\left(\sum_{j\in \Kscr}\big(g_j(\x, z; {\thetavec}^*) - \sigma_{j}(\x; \betavec^*)\big)\x_{(j)}\right)^{\top}
\right] \in \Real^{k \times k}.
$$
Equipped with these assumptions and definitions, we show the following key result. 

\begin{thm}[Consistency and Asymptotic Normality of AI-augmented Estimator]
\label{thm:consistency}
~
\begin{itemize}
    \item [$(i)$] Under Assumption \ref{assumption:regularity-consistency}, the optimizer  
    $\betavec^*$ defined in \eqref{eqn:best-in-class} is unique and the AAE satisfies 
    $\hat{\betavec}^{\sf AAE} \pconverge \betavec^*$, when $m,n\rightarrow \infty$.

    \item [$(ii)$] Under Assumptions \ref{assumption:regularity-consistency} and \ref{assumption:regularity-normality}, it holds that 
    \begin{align*}
            \sqrt{n}(\hat{\betavec}^{\sf AAE} - \betavec^*)&= \Omegavec^{-1} \left(\frac{1}{n}\sum_{i=1}^n  \sum_{j\in \Kscr}\big(g_j(\x, z; {\thetavec}^*) - \sigma_{j}(\x; \betavec^*)\big)\x_{(j)} + \sqrt{\frac{n}{m}} \Gammavec\times \sqrt{m}(\hat{\thetavec} - \thetavec^*)\right) + o_{\text{\upshape{P}}}(1)\\
    &\rightsquigarrow N\Big(\Ovec,\, \Omegavec^{-1}\big(\Jvec + \rho\times\Gammavec\Lambdavec\Gammavec^{\top}\big)\Omegavec^{-1}\Big).
    \end{align*}
\end{itemize}
\end{thm}

While some aspects of our proof of Theorem \ref{thm:consistency} draw on classical analysis of extremum estimators, the consistency argument we present for a misspecified model with two-stage estimators appears to be novel.
First, the identification of the model parameter and its justification (Lemma \ref{lem:unique-optimizer}) is, to the best of our knowledge, new. 
Second, with the presence of auxiliary data $z$, we reformulate the loss function as shown in \eqref{eqn:best-in-class}. This leads to a novel approach that converts the consistency of $\hat{\thetavec}$ into the uniform convergence of $\widehat{Q}(\hat{\thetavec}; \betavec)$ (Lemma \ref{lem:uniform-convergence}). 
Third, our concavity-based consistency analysis also seems fresh.

\subsection{Value of AI-Augmented Estimation}
\label{sec:AAE}
We compare our proposed estimator, $\hat{\betavec}^{\sf AAE}$, with the three benchmark approaches $\hat{\betavec}^{\sf Naive}$, $\hat{\betavec}^{\sf A}$ and $\hat{\betavec}^{\sf P}$ that were discussed in Section \ref{sec:3-estimators}. 
With Theorem \ref{thm:consistency} and Proposition \ref{prop:bias}, the merit of $\hat{\betavec}^{\sf AAE}$ over $\hat{\betavec}^{\sf Naive}$ and $\hat{\betavec}^{\sf A}$ is clear: while $\hat{\betavec}^{\sf AAE}$ delivers asymptotically unbiased estimation, both $\hat{\betavec}^{\sf Naive}$ and $\hat{\betavec}^{\sf A}$ are in general biased. 
Thus, the use of $\hat{\betavec}^{\sf AAE}$  enables valid estimation, inference, or
subsequent downstream optimization, for example, for product design, while the other two estimators lead to inferior decisions. 
Thus, let us focus on the comparison of $\hat{\betavec}^{\sf AAE}$ and $\hat{\betavec}^{\sf P}$ to demonstrate that the convergence speed of $\hat{\betavec}^{\sf AAE}$ being faster than that of $\hat{\betavec}^{\sf P}$. 

 In this case, standard analysis shows that, under appropriate regularity conditions, the resulting estimator $\hat{\betavec}^{\sf P}$ satisfies:
$
\sqrt{m}\big(\hat{\betavec}^{\sf P} - \betavec^*\big) \rightsquigarrow N\Big(\Ovec,\, \Omegavec^{-1}\check{\Jvec}\Omegavec^{-1}\Big),
$
where $\Omegavec$ is defined in \eqref{eqn:Omega-Gamma}, and
$$
\check{\Jvec}~:=~\E_{\x,y,z}\left[
    \left(\sum_{j\in \Kscr}\big(\ind_{\{y = j\}} - \sigma_{j}(\x; \betavec^*)\big)\x_{(j)}\right)
    \left(\sum_{j\in \Kscr}\big(\ind_{\{y = j\}} - \sigma_{j}(\x; \betavec^*)\big)\x_{(j)}\right)^{\top}
\right] \in \Real^{k \times k}.
$$
By Theorem \ref{thm:consistency}, the variance of the AAE, $\hat{\betavec}^{\sf AAE}$, can be expressed as
$
\Omegavec^{-1}\Jvec\Omegavec^{-1}/n + \Omegavec^{-1} \Gammavec\Lambdavec\Gammavec^{\top}\Omegavec^{-1}/m.
$
Thus, our AAE performs better if and only if:
\begin{align}
\label{eqn:variance-comparison}
    {\sf Var}^{\sf P}~:=~ \frac{1}{m}\Omegavec^{-1}\check{\Jvec}\Omegavec^{-1} ~\succeq~ {\sf Var}^{\sf AAE}~:=~\frac{1}{n}\Omegavec^{-1}\Jvec\Omegavec^{-1} + \frac{1}{m}\Gammavec\Lambdavec\Gammavec^{\top}\Omegavec^{-1},
\end{align}
or equivalently $\check{\Jvec}/m \succeq {\Jvec}/n + \Omegavec^{-1}\Gammavec\Lambdavec\Gammavec^{\top}/m$.
The next result follows. 

\begin{prop}[Dominance of ${\sf Var}^{\sf AAE}$.] \label{prop:weak-dom}
Assume Assumptions \ref{assumption:regularity-consistency} and \ref{assumption:regularity-normality} hold and $$\Lambdavec ~=~\E_{\x,y,z}\big[
    \nabla_{\theta} \log g_y(\x, z, \thetavec^*)
    \nabla_{\theta} \log g_y(\x, z, \thetavec^*)^{\top}
    \big]^{-1}.$$ 
    \begin{itemize}
        \item [$(i)$] It holds that $\check{\Jvec} \succeq \Gammavec\Lambdavec\Gammavec^{\top}$. Therefore, for any $\delta > 0$ and any $m$, ${\sf Var}^{\sf AAE} \prec {\sf Var}^{\sf P} + \delta \mathbf{I}$ for all $n$ sufficiently large. 
    \item [$(ii)$] If $\check{\Jvec} \succ \Gammavec\Lambdavec\Gammavec^{\top}$, for any $m$, ${\sf Var}^{\sf AAE} \prec {\sf Var}^{\sf P}$ for all $n$ sufficiently large. 
    \end{itemize}
\end{prop}

The assumption we introduce regarding the form of $ \Lambdavec $ is quite mild—it holds as long as $ \hat{\thetavec} $ is estimated via MLE and the standard regularity conditions are met \citep{van2000asymptotic}. 
The first part of the proposition asserts that the variance of AAE weakly dominates the variance of $ \hat{\betavec}^{\sf P} $: as long as the auxiliary dataset is sufficiently large, the variance of $ \hat{\betavec}^{\sf AAE} $ does not exceed that of $ \hat{\betavec}^{\sf P} $, up to an arbitrarily small constant. This conclusion follows directly from $ \check{\Jvec} \succeq \Gammavec\Lambdavec\Gammavec^{\top} $ because we can let $ n $ tend toward infinity in \eqref{eqn:variance-comparison}. The argument hinges on representing $ \check{\Jvec} - \Gammavec\Lambdavec\Gammavec^{\top} $ as the variance of the residual from projecting $ \sum_{j\in \Kscr}\big(\ind_{\{y = j\}} - \sigma_{j}(\x; \betavec^*)\big)\x_{(j)} $ onto the linear space spanned by $ \nabla_{\theta} \log g_y(\x, z, \thetavec^*) $.
For the second part, if $ \check{\Jvec} \succ \Gammavec\Lambdavec\Gammavec^{\top} $, the strong dominance $ {\sf Var}^{\sf AAE} \prec {\sf Var}^{\sf P} $ for all sufficiently large $ n $ follows by a similar reasoning.

Is $ \check{\Jvec} \succ \Gammavec\Lambdavec\Gammavec^{\top} $ generally true? By interpreting $ \check{\Jvec} - \Gammavec\Lambdavec\Gammavec^{\top} $ as the variance of the residual from projecting $ \sum_{j\in \Kscr}\big(\ind_{\{y = j\}} - \sigma_{j}(\x; \betavec^*)\big)\x_{(j)} $ onto the space spanned by $ \nabla_{\theta} \log g_y(\x, z, \thetavec^*) $, we observe that this inequality should hold in many cases, although providing a general, primitive sufficient condition seems challenging without a specific form of $g(\cdot)$. To illustrate this further and gain more insights, we refer the readers to Appendix \ref{app:comparison}, 
where we show that under specific parametric forms of $g(\cdot)$, item (ii) in the proposition indeed holds in general, and more informative AI-generated labels leads to more variance reduction and therefore higher estimation accuracy.

\revise{Finally, we remark that while our original asymptotic theory assumed $m \to \infty$, which is unrealistic, it is well known (e.g., via Berry-Esseen’s Central Limit Theorem, \cite{feller1971introduction}) that asymptotic variance formulas often provide accurate approximations even at moderate sample sizes. Thus, we believe our asymptotic results still offer meaningful insights into how the AAE estimator reduces variance and bias relative to standard baselines. Moreover, the asymptotic analysis shows that although the machine data are biased, the correction yields unbiased estimates, an essential property for any debiasing method. In contrast, na\"ive approaches, such as simply concatenating two datasets, produce biased estimates even with large sample sizes. 
Motivated by this, we next present a finite sample analysis that is more realistic.} 

\subsection{Finite Sample Analysis}
\label{sec:finite}

\revise{
Given that we typically aim to augment a limited amount of human-labeled data with AI-generated data, we extend our theoretical analysis in this section to derive a finite-sample, high-probability performance bound when human data is finite. Throughout, we assume that $\x = (\x^{\top}_{(1)}, \x^{\top}_{(2)}, \dots, \x^{\top}_{(k)})^{\top}$ is a random vector such that each subvector $\x_{(j)}$ has sub-Gaussian norm $\sigma_{\x_{(j)}}$ and is on a bounded support. Without loss of generality, we assume with probability one that $\lVert \x_{(j)} \rVert_{\infty} \leq 1$ for all $j \in \Kscr$. Let $\Sigmavec = \E\left[\sum_{j \in \Kscr} \x_{(j)} \x_{(j)}^{\top}\right]$ denote the population covariance, which we assume is positive definite. We further define
$
\kappa = \E\left[\max_{j \in \Kscr} \lVert \x_{(j)} \rVert_2^2\right] \leq \operatorname{trace}(\Sigmavec).
$
Additionally, we assume prior knowledge of an $\ell_1$-norm bound on the true parameter vector, i.e., $\lVert \betavec^* \rVert_1 \leq R$ for some known constant $R$, and we solve \eqref{eqn:AAE-def} under the constraint $\lVert \betavec \rVert_1 \leq R$.
These assumptions are standard in finite-sample analyses of logistic-type models \citep{chardon2024finite, chen2022statistical, wainwright1945high}. Since the specification of $\gvec(\cdot)$ can vary depending on the application, the performance of the first-stage estimation may differ accordingly. To address this, we introduce the following critical assumption. Let $\Tilde{d}$ denote the dimensionality of the parameter space in $\gvec(\cdot)$.

\begin{assm}[First Stage Performance]
With probability at least $1 - \delta$, it holds that 
$$
\E_{\x, z}\Big[
    {\sf KL}\big( \mathbf{g}(\x, z, \thetavec^*)\,\big\vert\,\mathbf{g}(\x, z, \hat{\thetavec})\big)
    \Big] \preceq \epsilon_1\big(m,\Tilde{d}, \delta\big).
$$    
\end{assm}

Under this assumption, we state our key result as follows. 

\begin{thm}[AAE Finite Sample Bound]
\label{thm:aae-bound}
    It holds that
    with probability at least $1-3\delta$,  $$
    \big\lVert \hat{\betavec}^{\sf AAE} - {\betavec}^* \big\rVert_2~\leq~\frac{
\left(2\sqrt{2\kappa} + 8  
       \right)e^R(1 + ke^R)^2\sqrt{\epsilon_1\big(m,\Tilde{d},\delta\big)}
}{\lambda_{\min}(\Sigmavec)}$$
    provided that ($c_1$ and $c_2$ are absolute constants) $$
     n \geq 
\max\left\{
\frac{d
\left(
\frac{1}{c_2}\log\left(\frac{k}{\delta}\right) \vee 1\right)
}{\frac{\lambda_{\min}(\Sigmavec)}{2K}\wedge 1},~ \frac{2d\left(\sum_{j \in \Kscr}\sigma_{\x_{(j)}}\right)^2}{c_1\epsilon_1\big(m,\Tilde{d},\delta\big)}\log\left(\frac{4}{\delta}\right)
    \right\}
    $$
\end{thm}

A key insight of the theorem is that the error is critically dependent on the error of estimating the conditional probability, $\Prob(y | \x, z)$, in the first stage. While the specific error bounds vary depending on the model specification, an order of $\sqrt{\tilde{d}/m}$ appears typical.
We illustrate this insight with the following example, which applies to a range of potential applications.

\begin{examp}[Logistic Regression for First Stage]
\label{exam:logistic}
{\upshape
    Let us assume that 
    $$
    g_{j}(\x, z; \thetavec) = \frac{e^{\theta^{\top}\phi_j(\x, z)}}{1+\sum_{j' \in \Kscr}e^{\theta^{\top}\phi_{j'}(\x, z)}}~~\forall j \in \Kscr.
    $$
    where for every $j$, $\phi_{j}(\x, z) : \Real^{d+1} \mapsto \Tilde{d}$ is a feature mapping. For a simplistic instance, we may have $\phi_{(j)}(\x, z) = (\Indicator\{z = j\}, \x^{\top})^{\top}$ while there are many other possibilities. 
    For notation simplicity, let us write $\tilde{\x} = (\tilde{\x}^{\top}_{(1)}, \tilde{\x}^{\top}_{(2)}, \dots, \tilde{\x}^{\top}_{(k)})^{\top}$, where $\tilde{\x}_{(j)} = \phi_{j}(\x, z)$ for all $j\in \Kscr$. 
    we assume it has a distribution such that each subvector $\tilde{\x}^{\top}_{(j)}$ has a sub-Gaussian norm $\sigma_{\tilde{\x}_{(j)}}$ and $\x^{\top}_{(j)}$'s infinity norm is bounded by one.
    Also, we assume that it is known that $\lVert \thetavec^* \rVert_{1} \leq \tilde{R}$ for some constant $\tilde{R}$ so estimate $\thetavec^*$ by solving the 
    maximum likelihood problem    with the constraint that $\lVert \thetavec^* \rVert_{1} \leq \tilde{R}$.
    Under these assumptions, we derive the following result on the information-theoretic excess-risk inequality, which generalizes that of classical regression or logistic regression analysis based on a localization argument. 
    To our knowledge, it is new.
    The hidden constants in the result depends on $R$,  $\Tilde{R}$, $k$, $\sigma_{\tilde{\x}_{(j)}}$, $\sigma_{{\x}_{(j)}}$, $\Sigmavec$ and $\kappa$.
    Since in typical applications of our approach, the number of products offered to the customer, $k$, is small, so we treat it as a constant and focus on the dependency of our finite sample bound on $m$, the sample size of primary data, $d$ and $\tilde{d}$, the number of parameters. 
    }

    \begin{prop}[Finite Sample KL-excess-risk Bound]
    \label{prop:kl}
    Under the setup in Example \ref{exam:logistic}, with probability at least $1 - \delta$, 
    $$
    \E_{\x, z}\Big[
    {\sf KL}\big( \mathbf{g}(\x, z, \thetavec^*)\,\big\vert\,\mathbf{g}(\x, z, \hat{\thetavec})\big)
    \Big] ~\preceq~ \frac{\tilde{d}\log (m/\delta) }{m}.
    $$
    Thus, if $n$ is sufficiently large, with probability $1-3\delta$, 
    $$
    \big\lVert \hat{\betavec}^{\sf AAE} - {\betavec}^* \big\rVert_2~\preceq~
    \sqrt{\frac{\tilde{d}\log (m/\delta)}{m}}.$$
    \end{prop}

    \upshape{
    When $\tilde{d} \lesssim d$, our estimator enjoys the finite-sample error bound $
\bigl\lVert \hat{\betavec}^{\sf AAE} -\betavec^{*}\bigr\rVert_{2}\;\lesssim\;\sqrt{\frac{d\log(m/\delta)}{m}}$.
One can compare existing finite sample analysis of MNL models \citep[e.g.,see][]{chen2022statistical} in the classical setting with ours. 
We observe that our result do not require any assumptions on the magnitude of $m$, which is typically required to ensure that approximate identification conditions
of the loss function is met to approximately recover $\betavec$. 
Instead, equipped with this strong KL bound from Step 1, Step 2 can simply draw a larger auxiliary sample (take $n$ large) to offset any conditioning issues.
Therefore, viewed this way, our two-step procedure offers a practical route to high-quality prediction without the stringent sample-size and conditioning requirements usually imposed for parameter estimation.
In essence, the AAE simply propogates the the first-stage risk, measured by the KL-divergence, into the second stage. 
}
\end{examp}}
\vspace{-5mm}
\subsection{Additional Discussions}
\label{sec:additional}

\revise{
Before we embark on the empirical discussions, we highlight two comments.
First, we remark that our framework of using AI-augmented data to facilitate the estimation and boost the accuracy extends beyond the setting for conjoint analysis. We believe that it is generally applicable for many statistical problems.
While leaving the study of more general problems to future work, in Appendix \ref{app:regression}, we extend our proposed data-augmentation approach to the regression setting with continuous target variables, providing an AI-Augmented Regression Estimator (AARE). 
Specifically, the setup for AARE maintains the core focus on two datasets: a small primary dataset containing real labels $y$ and AI-generated labels $z$, and a large auxiliary dataset with only features $\x$ and AI-generated labels $z$. The objective is to estimate the best linear predictor parameter $\betavec^{*}$ that minimizes the expected squared error, $\E[(y-\x^{\top}\betavec)^2]$. The AARE estimation proceeds in two steps that are analogous to AAE. In the discussion in Appendix \ref{app:regression}, we also provide a theoretical analysis of AARE, demonstrating that it possesses the desired asymptotic properties mirroring the key properties of AAE that have been discussed in this section.}

\revise{
Second, following our overview in the literature review comparing PPI \citep[e.g.,][]{angelopoulos2023prediction,angelopoulos2023ppi++} with our approach, we now provide a more detailed discussion. Our method directly learns the transformation of outcomes conditional on features (including the AI-generated outcome), whereas PPI focuses on subtracting the loss function of one outcome from another. The key distinction lies in the amount of information retained: by relying only on the loss function rather than the full mapping between features and outcomes, PPI risks discarding important information. When no such information is lost, the two approaches may perform similarly—typically in cases with few features and when the loss transformation approximates the outcome transformation closely. However, this equivalence rarely holds in practice. In many realistic settings with high-dimensional features, the PPI loss no longer captures the full mapping, and we find that our method outperforms PPI both in our settings (Tables \ref{tab:bias-reduction} and \ref{tab:aae-ppi-census}) and in theirs, with the performance gap widening as feature dimension increases.

Moreover, we acknowledge that our ability to capture the full mapping relies on a strong first-stage assumption, which parametrizes the distribution of outcomes conditional on features and AI labels. While this assumption underpins our theoretical analysis, our empirical studies show that it does not restrict the practicality of our approach and, in fact, enables AAE to outperform PPI. Moreover, this assumption provides additional flexibility and guarantees a provable performance boost. Importantly, in our framework, $z_i$ need not be the literal “label”—as long as it carries useful information and the parametric assumption holds (In the extreme, $z_i$ could even equal $-y_i$), our method remains applicable. By contrast, the utility of PPI, as emphasized in \cite{angelopoulos2023prediction}, depends critically on the accuracy of the AI-generated label. Furthermore, Proposition \ref{prop:weak-dom} establishes the dominance of our estimator over $\betavec^{\sf P}$, a discussion absent from the PPI framework.

Finally, the formats of asymptotic and finite-sample guarantees of the two approaches are comparable (see Theorems \ref{thm:consistency} and \ref{thm:aae-bound} in our paper versus Theorem 1 in \cite{angelopoulos2023ppi++} and Corollary B.3 in \cite{angelopoulos2023prediction}). For example, in finite samples, both methods yield high-probability error bounds that ultimately depend on the informativeness of the AI-generated label. In our setting, this appears as the error term $\epsilon_1(m,\Tilde{d},\delta)$, while in PPI it appears as the width of the rectifier confidence interval $(\mathcal{R}^l_{\alpha}(\theta), \mathcal{R}^u_{\alpha}(\theta))$. Although the exact comparison depends on the problem, our empirical results consistently suggest that our approach tends to outperform PPI.
}

%% file: src/5_empirical_1.tex
\section{Empirical Analysis I: COVID-19 Vaccination}
\label{sec:empirical-1}

\revise{
In this section, we present empirical studies to validate our approach. First, we show that a range of LLM models, despite their differences in capabilities, all exhibit substantial misalignment between AI-generated and human-generated data in conjoint analysis. Second, we demonstrate that our proposed method can effectively mitigate this misalignment, leading to improved performance compared with using only human data or a naïve combination of human and AI-generated data. It is important to emphasize that our method functions as an enhancement to existing models. Even as LLMs continue to improve in their ability to simulate human responses, as long as some degree of misalignment remains between AI-generated and human labels, our approach can further enhance estimation accuracy.}

% while our method is theoretically guaranteed to be correct in the asymptotic sense, it relies on two critical assumptions: the accurate knowledge of the function $g$ and the positive definiteness of the matrix $\check{\Jvec} - \Gammavec\Lambdavec\Gammavec^{\top}$. Both assumptions cannot be directly verified, so solid empirical study is necessary to demonstrate the merit of our proposed approach.}

\subsection{Empirical Setup}
\label{subsec:empirical-setup-1}
We examine the performance of the AAE based on a high-impact real choice-based conjoint dataset for COVID-19 vaccines \citep{kreps2020factors}. The study was conducted on July 9, 2020, where 2,000 US adults were recruited to take a 15-minute survey through the Lucid platform. A quota-based sampling was employed to approximate nationally representative samples in terms of demographic characteristics. A total of 1,971 US adults responded to the survey. This survey is important because it provides evidence of factors associated with individual preferences toward COVID-19 vaccination. The results may help inform public health campaigns to address vaccine hesitancy.

The dataset consists of responses from 1,971 participants, each expressing preferences for a series of hypothetical vaccines. Every respondent was shown five comparisons between two hypothetical vaccines, described by seven attributes with multiple levels, as outlined in Table \ref{tab:attributes}. Participants were asked to choose one of the two vaccines or opt for neither. We excluded respondents who did not select any vaccines in this setting, as many public LLMs, such as ChatGPT and Gemini, do not permit opting out of vaccines due to safety requirements. Since this data was excluded from both ground truth calculations and LLM data augmentation, this sample selection should not bias our comparisons between estimators. 

Data leakage is a critical concern when selecting datasets for LLM-related empirical studies. It occurs when an LLM's training data overlaps with the testing data used to evaluate its performance, thereby compromising the validity of the evaluation. In our empirical studies, we use OpenAI's GPT models. Since OpenAI does not disclose its training data, we cannot confirm whether the conjoint dataset is part of the GPT models' training data. However, we do not consider this to be a significant issue. If data leakage were present---that is, if the GPT models had already seen the conjoint dataset---the na\"ive augmentation method would perform substantially better. Nevertheless, as we demonstrate later, the na\"ive augmentation method still produces significant biases, which our method effectively corrects. This highlights the critical importance of our approach.

Using standard MLE, we estimated the best-in-class parameters $\bm{\beta}^*$, as shown in \eqref{eqn:best-in-class}. For each experiment, we randomly selected 240 respondents from the training set, yielding a dataset of 1,200 samples. The vaccine attributes in these samples were converted to text, and various versions of GPT were used to generate labels $z_i$. Details of the label generation process are provided in \cref{subsec:data-generation}. This produced a dataset $\Dscr$ with 1,200 data points.

Based on $\Dscr$, we considered different primary and auxiliary dataset sizes with $m \in \{50, 100, 150, 200\}$ and $n = 1,000$. For each combination of $(m, n)$, we randomly sampled ${m}/{5}$ respondents from $\Dscr$, using their data as the primary set, $\Dscr^{\sf P} = \{(\x^{\sf P}_i, y^{\sf P}_i, z^{\sf P}_i)\}_{i=1}^{m}$. From the remaining dataset, we sampled ${n}/{5}$ respondents and used their vaccine features and GPT-generated labels as the auxiliary set, $\Dscr^{\sf A} = \{(\x_i, z_i)\}_{i=1}^{n}$. 
\revise{
For each GPT model, this resulted in a primary set of size $m$ and an auxiliary set of size $n$.
In the next, we will discuss how to collect the auxiliary data in the conjoint analysis using different GPT models.} 

\begin{table}
\renewcommand{\arraystretch}{1.5}
\centering
{\footnotesize
\caption{Attributes in the conjoint analysis}
\label{tab:attributes}
\begin{tabular}{p{45mm}p{112mm}}\hline\hline
\multicolumn{1}{l}{\textbf{Feature}}
 &  
 \multicolumn{1}{l}{\textbf{Levels and Description}}
 \\\hline
{Efficacy}     & 50\%, 70\%, and 90\% (protection against severe symptoms);                     \\\hline
{Protection Duration} & `1 year' and `5 years'
\\\hline
{Major Side-effects}  & `1 in 1,000,000' and `1 in 10,000' (hospitalization or death) 
\\\hline
{Minor Side-effects}  & `1 in 10' and `1 in 30' (flu-like symptoms)
\\\hline
{FDA approval process} & `The vaccine has been approved and licensed by the US Food and Drug Administration.' and `The vaccine has received an emergency use authorization from the US Food and Drug Administration.  This allows the expedited use of promising drugs that the FDA has found it reasonable to believe may be effective in combatting the virus'
\\\hline
{National Origin of Vaccine} &`China', `United Kingdom', and `United States'
\\\hline
{Endorsement} & `President Donald Trump', `US Centers for Disease Control and Prevention', `Vice President Joe Biden', and `World Health Organization'
\\\hline
\hline
\end{tabular}
}
\end{table}

\subsection{Conjoint Data Generation using LLMs}
\label{subsec:data-generation}
In this section, we present our procedure for generating conjoint data using LLMs. We propose a general framework for LLM-based conjoint data generation, where the input structure follows three key components to produce conjoint choice data:

\vspace{10pt}

\begin{mdframed} \footnotesize \underline{{\sc LLM-Based Conjoint Data Generation Framework}} 
\begin{enumerate} 
\item[(I)] \textbf{\textbf{Choice Task Instruction}}: Defining the context of the conjoint analysis, specifying the item to choose from, the task for the LLM (selecting from multiple options), and the persona the LLM should emulate. 
\item[(II)] \textbf{\textbf{Choice Task Options Description}}: Details the attributes of each option in the conjoint choice set. 
\item[(III)] \textbf{\textbf{Choice Extraction}}: Extracting the choice made by the LLM. \end{enumerate} 
\end{mdframed}

\vspace{10pt}

This framework allows customization across different prompt designs, conjoint settings, and simulated subject groups, making it adaptable to various LLMs. The structured input ensures consistency in responses. In the subsequent sections, we use this framework to generate choice data with OpenAI's GPT, specifically \texttt{GPT-3.5-Turbo-0613}, \texttt{GPT-3.5-Turbo-0125}, \texttt{GPT-4}, and the \texttt{GPT-4o}.\footnote{\revise{At the time this paper was first written, \texttt{GPT-5} was not yet available to regular OpenAI API users and was therefore excluded from our empirical analysis. Nevertheless, as discussed earlier, because our method serves as an enhancement to existing models, we do not expect the qualitative conclusions of our results to change as model quality improves—so long as some degree of misalignment between model-generated and human labels remains.}} For each model, we generate conjoint data using two prompt engineering approaches: basic prompting and CoT prompting. For \texttt{GPT-4o}, we also implemented the few-shot prompting and fine-tuning.
\revise{
Before presenting the details of the prompting and fine-tuning techniques in the next, here we remark that our approach is not intended as an alternative to fine-tuning or prompt-based methods for improving LLM outputs. Rather, they are complementary, since our method, AAE, can be applied on top of outputs from models that have already been fine-tuned or employ sophisticated prompt engineering.} 

\subsubsection{Basic Prompting.}

 The basic prompting follows the three-part framework: it starts with an instruction specifying the choice task for GPT. In particular, we ask GPT to act like a random person to simulate the choice from a general population. More detailed demographic information can be added in this part to simulate choices from a more specific population. In the second part of the prompt, we parse the choice task options to text description. We use the minimalist representation of vaccine features, which was shown to be an effective prompting technique \citep{minimalistic}. In the last part, we ask which option GPT would choose and require it to return the answer as a single letter. An example of the basic prompt for GPT is shown below.
 
\vspace{10pt}
\begin{mdframed}
{\footnotesize
\textit{Input (I)}: You should act like a random person deciding whether or not to receive a COVID-19 vaccination.\\

 \noindent \textit{Input (II)}: There are three options for you:\\
A: Efficacy: 90\%. Protective duration: 5 years. Major side effect: 1 in 10,000. Minor side effect: 1 in 30. Authorization: Approved and licensed by the US Food and Drug Administration. Origin: China. Endorsement: US Centers for Disease Control and Prevention. \\
B: Efficacy: 50\%. Protective duration: 1 year. Major side effect: 1 in 10,000. Minor side effect: 1 in 10. Authorization: Received an emergency use authorization from the US Food and Drug Administration.  This allows the expedited use of promising drugs that the FDA has found it reasonable to believe may be effective in combatting the virus. Origin: United Kingdom. Endorsement: World Health Organization. 
C: You choose neither A nor B.\\

\noindent\textit{Input (III)}:  Which option would you choose as a random person? Your response should be a single letter A, B, or C.

\noindent\textit{Output}: A
}
\end{mdframed}

\subsubsection{Chain-of-Thought (CoT) Prompting.}
Recent advancements in prompting techniques, such as CoT prompting, Tree-of-Thought (ToT), and others, have significantly enhanced reasoning in LLMs. CoT, which structures intermediate reasoning steps, has become the most widely utilized due to its simplicity, interpretability, and robust performance across diverse tasks \citep{wei2022chain}. While methods like ToT, which explores branching reasoning paths \citep{yao2024tree}, and other approaches can offer advantages for specific tasks, they often require additional computational resources or fail to consistently outperform CoT across a wide range of domains. Thus, we experiment mainly with CoT for its balance of effectiveness and broad applicability.

We apply this CoT prompting technique with the aim of improving the quality of the choice data generated. Specifically, we introduce conversations that guide GPT to compare the two vaccines feature by feature before arriving at a final choice in part \textit{(III)} of the data generation framework. An example of the CoT-based conversation is shown below.

\vspace{15pt}

\begin{mdframed}
{\footnotesize
\textit{Input}: Compare the efficacy of A and B, which one is better?\\
\textit{Output}: Option A has a better efficacy rate at 90\% compared to option B's 50\% efficacy. This means that you would be more likely to be protected against COVID-19 with option A.\\\\
\textit{Input}: Compare the protective duration of A and B, which one is better?\\
\textit{Output}: Option A has a protective duration of 5 years, which is longer than Option B's protective duration of 1 year. Therefore, Option A offers better and longer protection.\\\\
\textit{Input}: Compare the major side effects of A and B, which one is better?\\
\textit{Output}: Both Option A and Option B exhibit the same odds of experiencing a major side effect, which is 1 in 10,000. Therefore, neither option is better than the other in this particular aspect. They are equal in terms of the risk of major side effects.\\\\
\textit{Input}: Compare the minor effects of A and B, which one is better? (1 in 30 is lower than 1 in 10)\\
\textit{Output}: Actually, Option B is better in terms of minor side effects. The risk of minor side effects for Option B is 1 in 10, which means 1 out of every 10 individuals could experience them. For Option A, it's higher with 1 in 30 individuals likely to experience minor side effects. In this scenario, a lower ratio (such as 1 in 10) means a higher chance of side effects.\\\\
\textit{Input}: Compare the authorization of A and B, which one is better?\\
\textit{Output}: Option A is approved and licensed by the US Food and Drug Administration, which is a more definitive sign of its safety and effectiveness than an emergency use authorization. Option B has received an emergency use authorization, which is a quicker approval process used during public health emergencies, and still shows that the FDA believes it may be effective against the virus but doesn't represent full approval. Therefore, in terms of authorization, Option A is better as it has full FDA approval.\\\\
\textit{Input}: Compare the endorsement of A and B, which one is better?\\
\textit{Output}: Both Option A and Option B have endorsements from highly respected organizations. Option A is endorsed by the US Centers for Disease Control and Prevention, while Option B is endorsed by the World Health Organization. It's difficult to categorically say which endorsement is better as they're both credible. However, if you're based in the US, you might have more trust in the endorsement from the US CDC. Conversely, if you're based elsewhere in the world or place more trust in a globally recognized body, the WHO's endorsement might carry more weight for you.\\\\
\textit{Input}: Based on all the previous reasoning, which option would you choose?\\
\textit{Output}: Based on efficacy, protective duration, and authorization, Option A seems to be the better choice. Although it has a slightly higher chance of minor side effects, its benefits in terms of providing more robust and longer-lasting protection against COVID-19 and having full FDA approval outweigh the slightly elevated risk. However, each individual's decision should take into account a multiple set of factors, including their personal health situation and discussions with healthcare professionals.\\\\
\textit{Input}: Return the choice in the previous answer as a single letter A, B, or C.\\
\textit{Output}: A
}
\end{mdframed}
\vspace{15pt}

As the example above illustrates, CoT helps GPT make a more informed choice and it allows us to know its reasoning process more explicitly. Note that even with CoT, GPT might still make certain mistakes, such as when comparing the magnitude of minor effects in the example above. This type of mistake often happens in arithmetic tasks, and it is hypothesized in the past due to tokenization of these models. One approach to alleviate this type of situation is to provide further guidance in the reasoning steps, such as indicating ``1 in 10" is worse than ``1 in 30" regarding minor effects, as shown in the above example. In general, CoT helps GPT to have a more comprehensive evaluation of all choice features, thus becoming more robust against minor reasoning errors. 
  
\subsubsection{Few-shot Prompting.}
Few-shot prompting is a technique in natural language processing where a model is guided to perform a specific task by providing a small number of examples directly in the input prompt \citep{brown2020language}. This approach is particularly useful in zero-shot and low-resource scenarios, demonstrating strong performance across a variety of tasks such as text classification, translation, and question-answering. We implement the few-shot prompting with \texttt{GPT-4o} by demonstrating ten survey questions with their true human labels at the beginning of part \textit{(I)} of the data generation framework. Below is an example of the few-shot prompting:

\vspace{10pt}
\begin{mdframed}
{\footnotesize
\textit{Input (I)}: Demonstrations:\\
You should act like a random person deciding whether or not to receive a COVID-19 vaccination.\\
There are three options for you:\\
A: Efficacy: 70\%. Protective duration: 5 years. Major side effect: 1 in 10,000. Minor side effect: 1 in 30. Authorization: Received an emergency use authorization from the US Food and Drug Administration.  This allows the expedited use of promising drugs that the FDA has found it reasonable to believe may be effective in combatting the virus. Origin: United States. Endorsement: World Health Organization. \\
B: Efficacy: 50\%. Protective duration: 5 years. Major side effect: 1 in 1,000,000. Minor side effect: 1 in 30. Authorization: Approved and licensed by the US Food and Drug Administration. Origin: China. Endorsement: US Centers for Disease Control and Prevention.\\ 
C: You choose neither A nor B. \\\\
Which option would you choose as a random person? Your response should be a single letter A, B, or C. \\\\
{\color{red} B}\\\\
... (the other nine demonstrations).\\
\\
You should act like a random person deciding whether or not to receive a COVID-19 vaccination.}
\end{mdframed}

\subsubsection{Fine-tuning.}
%Let us have a section on fine-tuning. 

Fine-tuning is another major approach in transfer learning that enhances a model's performance on a specific task with a few examples. In this process, a pre-trained model is adapted to a specific task by further training it on new, task-specific data. Fine-tuned models have the potential to achieve higher accuracy and better performance for the specific task. In our experiments, we fine-tuned \texttt{GPT-4o} by training it further upon 50 samples of real data.\footnote{OpenAI API recommends using 50-100 well-crafted demonstrations, which typically leads to clear improvements. } Basic prompting was employed to generate outputs from the fine-tuned \texttt{GPT-4o} model. Details of the fine-tuning implementation can be found at \cite{finetune}.

\subsection{Estimators and the Evaluation Metric}
\revise{
In this section, we describe the empirical construction and evaluation of our estimators. For each version of the GPT model, we compute the primary-data-only estimator, auxiliary-data-only estimator (i.e., using only AI-generated data), naïve augmentation estimator, and AAE estimator based on $(\Dscr^{\sf P}, \Dscr^{\sf A})$, denoted respectively as
$\hat{\betavec}^{\sf P}$, $\hat{\betavec}^{\sf A}$, $\hat{\betavec}^{\sf Naive}$, and $\hat{\betavec}^{\sf AAE}$.
We also implement the PPI estimator \citep{angelopoulos2023prediction} and the PPI++ estimator \citep{angelopoulos2023ppi++}, yielding $\hat{\betavec}^{\sf PPI}$ and $\hat{\betavec}^{\sf PPI++}$, respectively. All implementation details and code are provided in XXX.

Since the other estimators are straightforward to construct, we focus here on the hyperparameter choices for our AAE estimator. In Step 1 of AAE, we use a small feed-forward neural network to model $\big(g_j(\x, z; \thetavec^*): j \in \Kscr^+\big)$, consisting of two hidden layers with ten and five neurons, respectively, and employing the sigmoid activation function for all neurons. We train this network using the Adam optimizer \citep{kingma2014adam} with a learning rate of $10^{-4}$ to minimize the cross-entropy loss.\footnote{We also approximated the $\mathbf{g}(\cdot)$ function using alternative methods, including logistic regression and random forests. The qualitative results—namely, that our estimators outperform the alternatives—remained consistent. For brevity, we report only the neural network results.} In Step 2 of AAE, we pool both primary and axillary datasets to construct the empirical loss function in \eqref{eqn:AAE-def} to gain efficiency. Our main theoretical results remain valid under this modification, though additional care is required to account for the correlation between the primary data and $\hat{\thetavec}$.}

\revise{
To assess the performance of the estimators, we calculated the MAPE aggregated across all features:
\begin{equation*}
    {\sf MAPE}^l ~=~ \frac{1}{d}\left(\sum_{j=1}^d \frac{|\hat{\betavec}^l_j - \betavec^*_j|}{|\betavec^*_j|} \right) \times 100 \%\,,~~~ \forall l \,\in\, \{{\sf P}, {\sf A}, {\sf Naive}, {\sf AAE}, \revise{{\sf PPI}, {\sf PPI^{++}}} \}\,.
\end{equation*}
For each combination of dataset sizes $(m, n)$, we conducted 50 independent experimental runs and averaged the MAPE to serve as the final performance indicator. We adjust the MAPE values by adding a small constant to the denominator to eliminate the effect of parameters with very small magnitudes.
Since MAPE may place greater weight on features with small ground truth coefficients, we also demonstrate that the results remain robust when evaluated using MSE. Details are provided in Appendix \ref{app:empirical-mse}.
In the subsequent discussion, we use ${\sf MAPE}^{\sf P}$, the MAPE based on the primary dataset, as the reference point and report the difference in MAPE between each focal estimator and ${\sf MAPE}^{\sf P}$. This difference quantifies the performance gain achieved by employing a given estimator relative to using only the primary dataset. For instance, if ${\sf MAPE}^{\sf AAE} - {\sf MAPE}^{\sf P} = -10\%$, it indicates that the AAE estimator reduces the MAPE by 10\% compared with using the primary dataset.}

\subsection{Our Approach v.s. Other Approaches of Using LLM-generated Data}

\revise{
In this section, we present a detailed discussion of the empirical results comparing our estimators to other methods of utilizing LLM-generated data. \cref{tab:bias-reduction} summarizes MAPE difference between various estimators and the MAPE of using only the primary datasets, $\hat{\betavec}^{\sf P}$, 
i.e., ${\sf MAPE}^{l} - {\sf MAPE}^{\sf P}$ for $l \,\in\, \{{\sf P}, {\sf A}, {\sf Naive}, {\sf AAE}, \revise{{\sf PPI}, {\sf PPI^{++}}}\}$.
Please note that $\hat{\bm{\beta}}^{\sf p}$ is the same across all GPT models, with or without CoT, since it employs primary data only. Therefore, the difference in MAPE represents how much gain in MAPE one can get by employing a focal estimator in comparison with simply using the primary dataset.}
Please also refer to Section \ref{subsec:empirical-setup-1} for details of the error metric.
The first column of the table indicates the versions of the GPT model. The second column indicates the prompting technique. The rest of the columns show the difference in the MAPE after augmented $n=1,000$ GPT-generated samples to a real set with $m$ samples with $\hat{\betavec}^{\sf A}$, $\hat{\betavec}^{\sf Naive}$, $\hat{\betavec}^{\sf AAE}$, \revise{$\hat{\betavec}^{\sf PPI}$, and $\hat{\betavec}^{\sf PPI++}$}, respectively. Therefore, a \textit{negative} number indicates \textit{error reduction}, while a positive number means the error becomes larger after adding GPT data. 
Below, we highlight several key observations from the results.

\begin{table}[!ht]
\renewcommand{\arraystretch}{1.3}
\centering
{\footnotesize
\caption{Bias Reduction from $\hat{\betavec}^{\sf P}$ (\%)}
\label{tab:bias-reduction}
\begin{tabular}{cccccccccccccc}\hline\hline
& & \multicolumn{6}{c}{$m=50$} & \multicolumn{6}{c}{$m=100$} \\\hline
\textbf{Model}  & \textbf{Prompt} & $\hat{\betavec}^{\sf A}$ & $\hat{\betavec}^{\sf Naive}$  &  $\hat{\betavec}^{\sf PPI}$ & $\hat{\betavec}^{\sf PPI^{++}}$ & $\hat{\betavec}^{\sf AAE}$   & $p_{\max}$ & $\hat{\betavec}^{\sf A}$ & $\hat{\betavec}^{\sf Naive}$  & $\hat{\betavec}^{\sf PPI}$ & $\hat{\betavec}^{\sf PPI^{++}}$ & $\hat{\betavec}^{\sf AAE}$  & $p_{\max}$ \\\hline
\multirow{2}{*}{{\makecell{\texttt{GPT-3.5}\\\texttt{-Turbo-0613}}}} & Basic & -5.45 & -10.80 & - & - & -13.72 & 8E-11 & 1.30 & -4.53 & - & - & -6.90 & 4E-08 \\
 & CoT & 20.18 & 16.81 & - & - & -14.79 & 1E-15 & 26.93 & 20.62 & - & - & -8.09 & 3E-12 \\
\multirow{2}{*}{\makecell{\texttt{GPT-3.5}\\\texttt{-Turbo-0125}}} & Basic & -8.91 & -9.92 & - & - & -13.29 & 2E-14 & -2.16 & -3.21 & - & - & -6.50 & 2E-09 \\
 & CoT & 15.67 & 10.95 & - & - & -14.84 & 2E-15 & 22.43 & 14.71 & 25.28 & 5.14 & -7.72 & 4E-11 \\
\multirow{2}{*}{\texttt{GPT-4}} & Basic & 14.81 & 12.39 & - & - & -15.70 & 6E-17 & 21.56 & 16.20 & - & - & -8.04 & 3E-12 \\
 & CoT & 21.77 & 18.12 & - & - & -15.80 & 2E-16 & 28.53 & 22.49 & 23.29 & 4.13 & -8.30 & 2E-09 \\
\multirow{3}{*}{\texttt{GPT-4o}} & Basic & 15.70 & 13.05 & - & - & -15.55 & 1E-16 & 22.46 & 17.34 & 23.05 & 4.93 & -8.06 & 2E-12 \\
 & CoT & 20.61 & 16.50 & - & - & -15.74 & 2E-16 & 27.37 & 20.49 & 19.30 & 4.42 & -8.06 & 3E-12 \\
 & FS & 12.46 & 9.71 & - & - & -16.16 & 9E-18 & 19.22 & 14.56 & 17.83 & 4.46 & -8.26 & 2E-12 \\
\makecell{\texttt{GPT-4o}\\   \texttt{Fine-tuned}} & Basic & 4.83 & 3.18 & - & - & -16.66 & 2E-18 & 11.59 & 7.96 & 22.90 & 4.90 & -9.58 & 2E-14\\ \hline  

& & \multicolumn{6}{c}{$m=150$} & \multicolumn{6}{c}{$m=200$} \\\hline
\textbf{Model}  & \textbf{Prompt} & $\hat{\betavec}^{\sf A}$ & $\hat{\betavec}^{\sf Naive}$  &  $\hat{\betavec}^{\sf PPI}$ & $\hat{\betavec}^{\sf PPI^{++}}$ & $\hat{\betavec}^{\sf AAE}$   & $p_{\max}$ & $\hat{\betavec}^{\sf A}$ & $\hat{\betavec}^{\sf Naive}$  & $\hat{\betavec}^{\sf PPI}$ & $\hat{\betavec}^{\sf PPI^{++}}$ & $\hat{\betavec}^{\sf AAE}$  & $p_{\max}$ \\\hline
\multirow{2}{*}{\makecell{\texttt{GPT-3.5}\\\texttt{-Turbo-0613}}} & Basic & 6.90 & -2.06 & - & - & -2.09 & 0.47 & 7.56 & -1.79 & - & - & -0.96 & 0.979 \\
 & CoT & 32.53 & 23.06 & 19.34 & 2.93 & -3.12 & 2E-05 & 33.19 & 21.10 & 16.08 & 2.19 & -1.82 & 4E-04 \\
\multirow{2}{*}{\makecell{\texttt{GPT-3.5}\\\texttt{-Turbo-0125}}} & Basic & 3.43 & 0.86 & 24.15 & 3.48 & -1.78 & 0.006 & 4.10 & 1.60 & 16.70 & 1.92 & -0.76 & 0.091 \\
 & CoT & 28.02 & 17.72 & 16.44 & 3.39 & -2.74 & 2E-04 & 28.69 & 16.16 & 10.94 & 1.95 & -1.81 & 7E-04 \\
\multirow{2}{*}{\texttt{GPT-4}} & Basic & 27.16 & 20.03 & 11.61 & 2.91 & -3.12 & 3E-05 & 27.82 & 19.20 & 10.90 & 1.94 & -2.04 & 2E-04 \\
 & CoT & 34.12 & 26.33 & 13.86 & 2.95 & -3.37 & 9E-06 & 34.79 & 24.10 & 10.21 & 1.96 & -2.27 & 3E-05 \\
\multirow{3}{*}{\texttt{GPT-4o}} & Basic & 28.05 & 20.90 & 13.93 & 3.23 & -3.07 & 4E-05 & 28.72 & 19.65 & 11.13 & 1.92 & -1.84 & 2E-04 \\
 & CoT & 32.96 & 23.65 & 13.28 & 3.29 & -3.31 & 8E-06 & 33.63 & 21.28 & 10.20 & 1.99 & -2.25 & 5E-05 \\
 & FS & 24.81 & 18.41 & 11.68 & 2.78 & -3.44 & 6E-06 & 25.48 & 17.50 & 11.31 & 2.18 & -2.10 & 9E-05 \\
\makecell{\texttt{GPT-4o}\\   \texttt{Fine-tuned}} & Basic & 17.18 & 12.46 & 13.25 & 2.87 & -4.81 & 8E-10 & 17.85 & 11.31 & 11.21 & 1.93 & -3.36 & 5E-09\\\hline\hline
\end{tabular} 
\begin{minipage}{15.5cm}
    \footnotesize
    \vspace{3pt}
    \begin{tabular}{p{1cm} p{14cm}}
         \textbf{Notes}: & \setlength{\baselineskip}{1.2\baselineskip} 
        For each estimator under each experimental condition, we perform 50 independent runs and report the average MAPE difference between the focal estimator and $\hat{\betavec}^{\sf P}$. In each condition, the estimators have access to a real dataset with $m$ samples and an auxiliary dataset containing $n = 1{,}000$ GPT-generated samples. The reported average MAPE difference thus represents the increase or decrease in MAPE resulting from incorporating these $n = 1{,}000$ AI-generated samples, relative to using only the $m$ human-labeled samples. For each experimental condition, we also report the maximum p-value from all pairwise t-tests comparing the MAPE of $\hat{\betavec}^{\sf AAE}$ with those of the other estimators under the column $p_{\max}$. The test shows that $\hat{\betavec}^{\sf AAE}$ outperforms other estimators at the 99\% significance level for all instances, except for the ones marked with a star. A “–” symbol denotes cases where the numerical value exceeds 1{,}000, which primarily arises from singularity issues encountered by PPI-based methods when the primary dataset is small, leading to parameter estimates dominated by random initialization.    
    \end{tabular}
  \end{minipage}}
\end{table}

\subsubsection{AI-only Estimation or Na\"ive Augmentation Incurs More Error.} 
\revise{
Our results indicate that relying solely on AI-generated data (yielding estimator $\hat{\betavec}^{\sf A}$) or na\"ively combining AI-generated and human data (yielding estimator $\hat{\betavec}^{\sf Naive}$) can introduce substantial estimation errors relative to simply using primary dataset $\hat{\betavec}^{\sf P}$. As shown in \cref{tab:bias-reduction}, while $\hat{\betavec}^{\sf A}$ and $\hat{\betavec}^{\sf Naive}$ based on advanced models with advanced prompting techniques may actually increase estimation errors, these simple estimators can reduce error when using more basic models and prompts, such as \texttt{GPT-Turbo-3.5-0613} and \texttt{GPT-Turbo-3.5-0125}. This finding highlights an inconsistency between an LLM’s intrinsic capability and its effectiveness in generating data that resemble human preferences, an observation consistent with prior research \citep{goli2024frontiers}.
}

\revise{Interestingly, this inconsistency is counterintuitive: the performance of the AI-only estimator and the naïve augmentation approach does not improve with more advanced GPT models or more sophisticated prompting strategies; in many cases, it actually deteriorates. For example, after implementing CoT prompting, we observe that MAPE further increases across models. We conjecture that, in such cases, the LLM becomes “overly rational,” generating choices that may encode richer information but deviate more strongly from human behavior. Although fine-tuning a base model (in this case, GPT-4o) on the primary dataset helps mitigate some of the bias introduced by incorporating auxiliary data, it does not fully resolve the issue: including auxiliary datasets still increases MAPE relative to using the primary dataset alone.}

% only based on selected features.  
% Indeed, pure AI estimation and na\"ive augmentation both rely on the assumption that AI-generated data closely resembles the distribution of human data, which often does not hold, because LLMs are trained as general-purpose token-generators, not for conjoint analysis. 
% With limit data, fine-tuning and prompt engineering methods might not solve the problem either. 
% Using AI-generated data can become a wild goose chase—experimenting with various AI models, prompting techniques, or fine-tuning methods in hopes that one dataset aligns with human data. 
% However,
% as will be discussed in detail in the next section, our approach consistently adds value on top of these approaches on the engineering side. 
% These methods and ours operate under different constraints: in principle our approach can be applied off the shelf to any data consistent, without altering the LLM itself, whereas fine-tuning and prompt engineering are problem-specific and often time-consuming.}

From \cref{tab:bias-reduction}, one can see that the na\"ive augmentation consistently outperforms pure AI estimation. Indeed, blending in some real data can help align the estimation results. In particular, pure AI estimation can be regarded as a special case of na\"ive augmentation. In the following, we examine the performance of AAE by comparing it with the na\"ive augmentation.

\subsubsection{AAE Regulates Error Consistently.} 
\revise{First, as shown in \cref{tab:bias-reduction}, $\hat{\betavec}^{\sf AAE}$ is consistently negative across all models and prompting techniques, indicating that our AAE approach significantly reduces estimation errors relative to using human data alone—regardless of the underlying model or inference quality. We then perform pairwise t-tests comparing $\hat{\betavec}^{\sf AAE}$ with all other estimators in each setting reported in \cref{tab:bias-reduction}, and summarize the maximum p-values across these comparisons. The results show that $\hat{\betavec}^{\sf AAE}$ outperforms all benchmark estimators at the 99\% significance level in every instance, except for the few cases marked with a star.}

\revise{More interestingly, unlike the simpler estimators, AAE’s performance improves consistently with more advanced versions of GPT models and more sophisticated prompting strategies. It achieves the largest error reduction when applied to the fine-tuned \texttt{GPT-4o} model using all available data. We find that CoT prompting, few-shot prompting, and fine-tuning each contribute to significant error reductions, with fine-tuning being the most effective. As more advanced GPT versions and carefully designed prompts produce more informative AI-generated labels, the effectiveness of AAE correspondingly increases. This observation is consistent with our theoretical discussion in Appendix \ref{app:comparison}, which provides formal support for this intuition. This underscores the potential of AAE to deliver superior results with future iterations of LLMs or other AI-based data generators—provided that the augmentation follows a statistically sound estimation strategy such as ours. Indeed, AAE not only regulates the estimation error in AI-generated data but also offers a principled framework for guiding future improvements in AI-assisted estimation.}

\revise{Furthermore, we observe that AAE generally outperforms the PPI-based methods across different models and prompting strategies. The PPI estimator’s MAPE differences are missing for some experimental cases because its performance is unstable due to singularity issues, particularly when the primary dataset size $m$ is small. In \cref{sec:empirical-2-aae-ppi}, we provide a more detailed comparison between AAE and PPI-based methods using both a real dataset and a synthetic dataset, to further investigate the sources of their performance differences.}
% We conduct statistical tests on the results in \cref{tab:bias-reduction}. Specifically, pairwise t-tests show that $\hat{\betavec}^{\sf AAE}$ outperforms other benchmark estimatorse at the 99\% significance level for all instances, except for the few cases marked with a star. 

\begin{figure}[!t]
    \centering
    \includegraphics[width=0.5\textwidth]{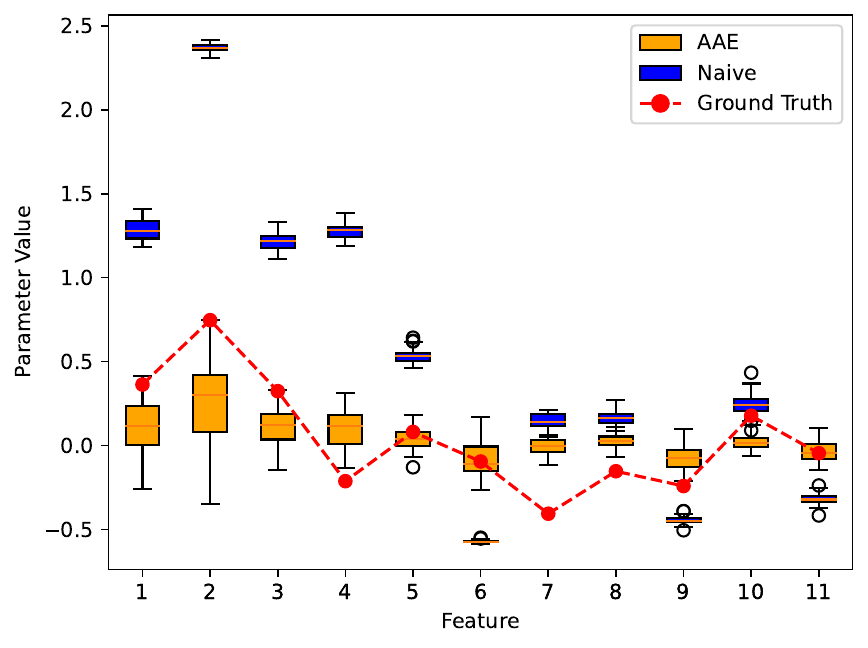}
    \caption{Illustration of  $\hat{\betavec}^{\sf AAE}$ and $\hat{\betavec}^{\sf Naive}$ by Feature}
    \label{fig:comparison}
    \vspace{3mm}
    \begin{minipage}{\textwidth}
    \footnotesize
    \begin{tabular}{p{1cm} p{15cm}}\hspace{-1mm}\textbf{Notes}: & \setlength{\baselineskip}{1.2\baselineskip} 
    The figures presents the ground-truth parameters $\betavec^*$ (in the red curve), $\hat{\betavec}^{\sf AAE}$ (orange box) and $\hat{\betavec}^{\sf Naive}$  (blue box). 
        The percentiles in the box plots are based on the 50 experimental runs with $m = 100$. 
    \end{tabular}
  \end{minipage}
\end{figure}

\revise{Finally, beyond the aggregate comparison in \cref{tab:bias-reduction}, Figure \ref{fig:comparison} illustrates the feature-level discrepancies between $\hat{\betavec}^{\sf Naive}$ and $\hat{\betavec}^{\sf AAE}$. The figure compares the estimated parameter values for eleven features across $\hat{\betavec}^{\sf Naive}$, $\hat{\betavec}^{\sf AAE}$, and the ground truth, using $m = 100$ primary data points and $n = 1{,}000$ auxiliary data points. As shown, the  na\"ive augmentation approach produces substantially larger estimation errors—particularly for the most influential vaccine features—whereas AAE yields estimates that are much more closely aligned with the ground truth.}

\begin{table}[!t]
\renewcommand{\arraystretch}{1.3}
\centering
{\footnotesize
\caption{Percentage of Saving in Data Size (\%)}
\label{tab:saving}
\begin{tabular}{cccccc}\hline\hline
Model    & Prompt & $m=50$          & $m=100$         & $m=150$         &$m=200$         \\\hline
\multirow{2}{*}{\texttt{GPT-3.5-Turbo-0613}} & Basic & 73.59   (0.69) & 46.48 (1.64) & 30.97 (1.68) & -0.86 (2.80)* \\
 & CoT & 76.05 (1.63) & 52.84 (2.61) & 37.34 (2.93) & 8.37 (4.49) \\
\multirow{2}{*}{\texttt{GPT-3.5-Turbo-0125}} & Basic & 71.64 (0.77) & 42.67 (1.63) & 27.65 (1.36) & -3.72 (2.75)* \\
 & CoT & 77.24 (1.23) & 52.44 (1.98) & 36.22 (2.19) & 10.73 (3.77) \\
\multirow{2}{*}{\texttt{GPT-4}} & Basic & 79.30 (1.20) & 53.23 (2.28) & 39.83 (2.00) & 14.01 (3.17) \\
 & CoT & 78.40 (1.68) & 53.70 (2.89) & 40.76 (2.22) & 14.96 (3.78) \\
\multirow{3}{*}{\texttt{GPT-4o}} & Basic & 77.86 (1.56) & 51.91 (2.74) & 38.10 (2.41) & 9.39 (4.35) \\
 & CoT & 78.81 (1.45) & 52.64 (2.66) & 41.02 (2.13) & 15.14 (3.82) \\
 & FS & 79.39 (1.41) & 53.79 (2.56) & 42.47 (1.97) & 13.00 (3.99) \\
\texttt{GPT-4o Fine-tuned} & Basic & 79.81 (1.63) & 58.04 (3.22) & 50.20 (2.27) & 24.86 (4.16)       \\\hline\hline
\end{tabular}}
\begin{minipage}{15.5cm}
\vspace{3pt}
    \footnotesize
    \begin{tabular}{p{1cm} p{14cm}}
         \textbf{Notes}: & \setlength{\baselineskip}{1.2\baselineskip} 
    This table presents the percentage of data saved using AAE with various GPT models, averaged over 50 experimental runs. The standard errors are shown in parentheses. One sample t-tests show that the savings are all significant at the 95\% level except for the ones marked with a star.
    \end{tabular}
  \end{minipage}
\end{table}

\subsection{Our Approach v.s. Human-only Data---Economic Value of Our Approach } 
\label{sec:empirical-1-economic-value}

So far, we have focused on evaluating our method based on estimation error against other ways of utilizing AI data. In practice, the primary goal of using AI as a data augmenter in market research is to reduce the costs of hiring real survey participants. Thus, we evaluated the percentage of data saved using AAE with various GPT models, as summarized in \cref{tab:saving}. Specifically, for a given error reduction achieved by applying AAE to a real dataset of size $n_1$, we calculated the amount of real data samples, $n_2$, required to achieve the same error reduction without AI augmentation. The percentage of data saved is then estimated as $((n_2 - n_1)/n_2) \times 100\%$.

Our results show that with fine-tuned \texttt{GPT-4o}, data savings range from 24.9\% to 79.8\%. When the primary set is small ($m=50$), AAE saves between 71.64\% and 79.8\% of data samples, regardless of the GPT model or prompt design. With a moderate primary set size ($m=100, 150$), savings range from 27.7\% to 58.0\%, again consistent across GPT models and prompts. Compared to the costs of recruiting real survey participants, the costs of generating AI-based data are negligible and will continue to decrease as generative AI technology advances. In practice, conjoint surveys must be regularly re-administered across different product categories and customer segments to account for evolving consumer preferences. This can result in a substantial number of surveys being required over time. Thus, we conclude that AAE offers significant cost savings in the long run.

%% file: src/6_empirical_2.tex
\section{Additional Empirical Analysis}
\label{sec:empirical-2}
In this section, we present additional empirical studies. In \cref{sec:empirical-2-car}, we assess the robustness of our method on an alternative dataset. In \cref{sec:empirical-2-aae-ppi}, we provide a more detailed comparison between AAE and PPI-based methods. 
% Finally, in \cref{sec:empirical-2-misspecification}, we explore that the performance of AAE under model misspecification.

\subsection{Empirical Analysis on Other Data}
\label{sec:empirical-2-car}

This section presents an alternative empirical study on a real choice-based conjoint dataset for sports cars \citep{sccdata} as a robustness check for the results in \cref{sec:empirical-1}. 
This dataset consists of responses from 200 participants, each expressing preferences for a series of hypothetical sports cars. Each respondent was shown ten sets of three sports cars, described by five attributes with multiple levels, as outlined in Table \ref{tab:attributes-2}.  Participants were asked to choose one of the three cars. 
We randomly selected 120 respondents from the training set for each experiment, resulting in a dataset of 1,200 samples. The car attributes in these samples were converted to text and used to generate GPT datasets, following a similar procedure to \cref{subsec:data-generation}, details of which are discussed in \cref{app:empirical-dgp-car}. The remainder of the experimental setup is also consistent with the approach described in \cref{subsec:empirical-setup-1}.

\begin{table}[!t]
\renewcommand{\arraystretch}{1.5}
\centering
{\footnotesize
\caption{Attributes in the Sports Car Conjoint Analysis}
\label{tab:attributes-2}
\begin{tabular}{p{45mm}p{112mm}}\hline\hline
\multicolumn{1}{l}{\textbf{Feature}}
 &  
 \multicolumn{1}{l}{\textbf{Levels and Description}}
 \\\hline
{Segment (Type of car model)}     & ``Basic", ``Fun", and ``Racer"                  
\\\hline
{Number of Seats} & 2, 4, and 5
\\\hline
{Type of transmission}  & ``Manual" and ``Automatic"
\\\hline
{Convertible Roofs}  & ``Yes" and ``No"
\\\hline
{Price} & Price in thousands \$, with three possible choices (30, 35, 40)
\\\hline \hline
\end{tabular}
}
\end{table}

\cref{tab:bias-reduction-2} summarizes the change in MAPE of $\hat{\betavec}^{\sf A}$, $\hat{\betavec}^{\sf Naive}$, and $\hat{\betavec}^{\sf AAE}$ in comparison to $\hat{\betavec}^{\sf P}$. Similarly to the findings in \cref{sec:empirical-1}, AAE significantly reduces MAPE in all scenarios, achieving in general the best performance using \texttt{GPT-4} with CoT and \texttt{GPT-4o} with few-shot prompting, while the na\"ive or AI-only methods tend to elicit errors. Interestingly, AAE's performance shows less correlation with more advanced GPT or prompting techniques compared to the vaccine setting, which is expected given that sports car preferences are highly subjective. Despite being an older version, \texttt{GPT-3.5-Turbo-0613} performs well with AAE in this context. One can notice that \texttt{GPT-3.5-Turbo-0613} incurs a small estimation error with na\"ive augmentation, suggesting it may happen to have a close alignment with real human data, which in turn enhances AAE's effectiveness. Moreover, fine-tuning \texttt{GPT-4o} does not improve the performance of na\"ive augmentation. This again underscores that fine-tuning is not a universal solution to all problems.

\begin{table}[!t]
\renewcommand{\arraystretch}{1.3}
\centering
{\footnotesize
\caption{Bias Reduction from $\hat{\betavec}^{\sf P}$ in Empirical Setting II(\%)}
\label{tab:bias-reduction-2}
\begin{tabular}{cccccccccc}\hline\hline
& & \multicolumn{4}{c}{$m=50$} & \multicolumn{4}{c}{$m=100$} \\\hline
\textbf{Model}  & \textbf{Prompt} & $\hat{\betavec}^{\sf A}$ & $\hat{\betavec}^{\sf Naive}$  & $\hat{\betavec}^{\sf AAE}$  & $p_{\max}$ & $\hat{\betavec}^{\sf A}$ & $\hat{\betavec}^{\sf Naive}$  & $\hat{\betavec}^{\sf AAE}$   & $p_{\max}$ \\\hline
\multirow{2}{*}{\texttt{GPT-3.5-Turbo-0613}} & Basic & -17.27 & -34.20 & -37.81 & 0.105* & 15.09 & -1.15 & -11.02 & 4e-04 \\
 & CoT & 148.81 & 119.98 & -34.33 & 3e-07 & 181.18 & 130.32 & -9.50 & 0.002 \\
\multirow{2}{*}{\texttt{GPT-3.5-Turbo-0125}} & Basic & 29.70 & 5.58 & -30.22 & 2e-07 & 62.06 & 32.47 & -9.80 & 0.001 \\
 & CoT & 129.80 & 112.47 & -35.35 & 9e-08 & 162.17 & 124.39 & -11.07 & 6e-04 \\
\multirow{2}{*}{\texttt{GPT-4}} & Basic & -12.43 & -16.70 & -30.63 & 2e-04 & 19.94 & 10.17 & -8.62 & 2e-04 \\
 & CoT & 173.34 & 132.71 & -38.32 & 2e-07 & 205.71 & 137.60 & -11.36 & 3e-04 \\
\multirow{3}{*}{\texttt{GPT-4o}} & Basic & 421.27 & 292.59 & -25.68 & 6e-06 & 453.64 & 263.41 & -8.92 & 0.006 \\
 & CoT & 304.98 & 229.80 & -28.65 & 4e-06 & 337.34 & 219.76 & -7.86 & 0.007 \\
 & FS & 153.33 & 89.90 & -31.61 & 2e-07 & 185.69 & 98.61 & -8.45 & 0.007 \\
\texttt{GPT-4o Fine-tuned} & Basic & 201.95 & 133.85 & -36.74 & 3e-07 & 234.32 & 132.10 & -12.85 & 6e-05 \\\hline
& & \multicolumn{4}{c}{$m=150$} & \multicolumn{4}{c}{$m=200$} \\\hline
\textbf{Model}  & \textbf{Prompt} & $\hat{\betavec}^{\sf A}$ & $\hat{\betavec}^{\sf Naive}$  & $\hat{\betavec}^{\sf AAE}$  & $p_{\max}$ & $\hat{\betavec}^{\sf A}$ & $\hat{\betavec}^{\sf Naive}$  & $\hat{\betavec}^{\sf AAE}$   & $p_{\max}$ \\\hline
\multirow{2}{*}{\texttt{GPT-3.5-Turbo-0613}} & Basic & 28.53 & 8.57 & -6.83 & 4e-04 & 30.71 & 8.26 & -7.09 & 5e-05 \\
 & CoT & 194.62 & 126.11 & -4.80 & 0.019 & 196.80 & 119.07 & -5.39 & 5e-04 \\
\multirow{2}{*}{\texttt{GPT-3.5-Turbo-0125}} & Basic & 75.50 & 37.96 & -5.30 & 0.005 & 77.68 & 34.69 & -6.59 & 2e-04 \\
 & CoT & 175.60 & 123.92 & -5.17 & 0.006 & 177.78 & 113.93 & -6.64 & 3e-04 \\
\multirow{2}{*}{\texttt{GPT-4}} & Basic & 33.37 & 21.31 & -4.60 & 0.005 & 35.55 & 19.68 & -4.72 & 2e-04 \\
 & CoT & 219.15 & 136.34 & -7.69 & 2e-04 & 221.32 & 121.56 & -6.62 & 7e-04 \\
\multirow{3}{*}{\texttt{GPT-4o}} & Basic & 467.07 & 234.85 & -4.56 & 0.030 & 469.25 & 207.47 & -7.03 & 2e-04 \\
 & CoT & 350.78 & 199.59 & -5.92 & 0.010 & 352.96 & 178.51 & -6.26 & 1e-04 \\
 & FS & 199.13 & 96.48 & -2.84 & 0.155* & 201.31 & 84.70 & -4.38 & 0.019 \\
\texttt{GPT-4o Fine-tuned} & Basic & 247.76 & 123.92 & -9.02 & 2e-04 & 249.93 & 110.3 & -6.99 & 4e-04
\\\hline\hline
\end{tabular} 
\begin{minipage}{15.5cm}
    \footnotesize
    \vspace{3pt}
    \begin{tabular}{p{1cm} p{14cm}}
         \textbf{Notes}: & \setlength{\baselineskip}{1.2\baselineskip}For each estimator under each experimental condition, we perform 50 independent runs and report the average MAPE difference between the focal estimator and $\hat{\betavec}^{\sf P}$. In each condition, the estimators have access to a real dataset with $m$ samples and an auxiliary dataset containing $n = 500$ GPT-generated samples. The reported average MAPE difference thus represents the increase or decrease in MAPE resulting from incorporating these $n = 500$ AI-generated samples, relative to using only the $m$ human-labeled samples. For each experimental condition, we also report the maximum p-value from all pairwise t-tests comparing the MAPE of $\hat{\betavec}^{\sf AAE}$ with those of the other estimators under the column $p_{\max}$. The test shows that $\hat{\betavec}^{\sf AAE}$ outperforms other estimators at the 95\% significance level for all instances, except for the ones marked with a star. 
    % This table presents the difference MAPE after augmented $n=500$ GPT-generated samples to a real set with $m$ samples with the AI-only estimator $\hat{\betavec}^{\sf A}$, na\"ive estimator $\hat{\betavec}^{\sf Naive}$ and the AAE estimator $\hat{\betavec}^{\sf AAE}$, respectively, averaged over 50 experimental runs in the second empirical setting. As in \cref{sec:empirical-1}, AAE significantly reduces error from the primary set, and outperforms the AI-only and na\"ive augmentation estimators. Pairwise t-tests show that $\hat{\betavec}^{\sf AAE}$ outperforms $\hat{\betavec}^{\sf P}$, $\hat{\betavec}^{\sf A}$, and $\hat{\betavec}^{\sf Naive}$ at the 95\% significance level for all instances, except for the ones marked with a star. The maximum p-value of the pairwise t-tests are shown under the column $p_{\max}$.
    \end{tabular}
  \end{minipage}}
\end{table}

\cref{tab:saving-2} shows the percentage of data that can be saved using AAE. Our results show data savings ranging from 5.1\% to 60.3\% across different models, prompts, and primary data sizes. When the primary set is small ($m=50$), AAE saves between 46.9\% and 60.3\% of data samples, regardless of the GPT model or prompt design. With a moderate primary set size ($m=100, 150$), savings range from 18.3\% to 40.2\%, again consistent across GPT models and prompts. These results further support the practical value of AAE in broad conjoint settings.

\begin{table}[!t]
\renewcommand{\arraystretch}{1.3}
\centering
{\footnotesize
\caption{Percentage of Saving in Data Size in Empirical Setting II(\%)}
\label{tab:saving-2}
\begin{tabular}{cccccc}\hline\hline
Model    & Prompt & $m=50$          & $m=100$         & $m=150$         &$m=200$         \\\hline
\multirow{2}{*}{\texttt{GPT-3.5-Turbo-0613}} & Basic & 59.39 (3.65) & 33.26 (5.47) & 30.81 (6.53) & 17.05 (10.32)* \\
 & CoT & 55.04 (4.04) & 29.30 (5.50) & 26.05 (6.04) & 10.47 (10.76)* \\
\multirow{2}{*}{\texttt{GPT-3.5-Turbo-0125}} & Basic & 52.96 (4.31) & 29.22 (7.26) & 26.29 (6.35) & 16.06 (10.71)* \\
 & CoT & 56.44 (3.48) & 34.18 (4.88) & 25.91 (6.99) & 14.97 (9.83)* \\
\multirow{2}{*}{\texttt{GPT-4}} & Basic & 48.99 (5.25) & 25.74 (7.29) & 24.37 (7.20) & 5.12 (12.09)* \\
 & CoT & 60.29 (3.54) & 34.24 (5.26) & 34.83 (5.69) & 15.81 (8.67) \\
\multirow{3}{*}{\texttt{GPT-4o}} & Basic & 46.90 (4.30) & 27.10 (7.07) & 23.71 (6.55) & 17.20 (9.25) \\
 & CoT & 51.39 (3.93) & 24.86 (5.88) & 29.72 (5.46) & 14.20 (9.81)* \\
 & FS & 58.18 (3.42) & 38.11 (5.10) & 40.23 (5.06) & 20.42 (8.60) \\
\texttt{GPT-4o Fine-tuned} & Basic & 50.26 (4.93) & 27.01 (5.92) & 18.34 (6.31) & 5.56 (11.90)*      \\\hline\hline
\end{tabular}}
\begin{minipage}{15.5cm}
\vspace{3pt}
    \footnotesize
    \begin{tabular}{p{1cm} p{13cm}}
         \textbf{Notes}: & \setlength{\baselineskip}{1.2\baselineskip} 
    This table presents the percentage of data saved using AAE with various GPT models, averaged over 50 experimental runs in the second empirical setting. The standard errors are shown in parentheses. One-sample t-tests show that the savings are all significant at the 95\% level except for the ones marked with a star.
    \end{tabular}
  \end{minipage}
\end{table}

\subsection{More Detailed Comparison with PPI-based Methods}
\label{sec:empirical-2-aae-ppi}
In this section, we present a more detailed comparison between AAE and the PPI-based methods. In addition to the analysis in \cref{sec:empirical-1} using the COVID-19 vaccine survey dataset, we evaluate both approaches on the U.S. private health insurance census dataset from the original PPI study \citep{angelopoulos2023prediction}. We further investigate their differences using a synthetic dataset designed to isolate specific performance characteristics.

\subsubsection{Real Data Experiments on Private Healthcare Census Dataset.}
\cref{tab:aae-ppi-census} summarizes the performance of AAE and PPI-based methods on the private health insurance census dataset used in \cite{angelopoulos2023prediction}. Since the PPI-based methods suffered from singularity issues under small primary dataset in the experiment settings in \cref{sec:empirical-1}, we tested larger primary data size $m$ ranging from $100$ to $1{,}000$ in this experiment for a more comprehensive comparison.

Similar to the COVID-19 vaccine setting in \cref{sec:empirical-1}, AAE outperforms PPI-based methods. The performance gap is more pronounced in the vaccine setting. A key difference between the two datasets lies in feature dimensionality: the vaccine dataset includes $11$ features, while the census dataset contains only two. 

As discussed earlier, the distinction between AAE and PPI-based methods is where the correction is applied—at the loss-function level (PPI) versus the mapping-function level (AAE). This implies that performance differences become more significant when the loss function captures less information embedded in the data-generating process, a scenario more likely with higher feature dimensionality. The results from the two datasets are consistent with this reasoning. To further investigate this effect, we conduct a synthetic data study examining how feature dimensionality impacts the relative performance of AAE and PPI-based methods.

\begin{table}[!ht]
\renewcommand{\arraystretch}{1.3}
\centering
{\footnotesize 
\caption{Bias Reduction from $\hat{\betavec}^{\sf P}$ in the Private Healthcare Census Setting (\%)}
\label{tab:aae-ppi-census}
\begin{tabular}{ccccc}\hline\hline
$m$  & $\hat{\betavec}^{\sf AAE}$    & $\hat{\betavec}^{\sf PPI}$    & $\hat{\betavec}^{\sf PPI++}$ & $p_{\max}$ \\\hline
100  & -65.86 & -22.29 & -15.22 & 0.04       \\
250  & -35.98 & -12.7  & -4.7   & 0.03       \\
500  & -29.93 & -7.58  & -2.41  & 0.01       \\
750  & -10.74 & -2.13  & -1.06  & 0.11*       \\
1000 & -11.87 & -4.56  & -1.14  & 0.13*   \\\hline\hline
\end{tabular}
\begin{minipage}{15.5cm}
\vspace{3pt}
    \footnotesize 
    \begin{tabular}{p{1cm} p{13cm}}
         \textbf{Notes}: & \setlength{\baselineskip}{1.2\baselineskip} For each estimator under each experimental condition, we perform 50 independent runs and report the average MAPE difference between the focal estimator and $\hat{\betavec}^{\sf P}$. In each condition, the estimators have access to a real dataset with $m$ samples and an auxiliary dataset containing $n = 2{,}000$ auxiliary samples. The reported average MAPE difference thus represents the increase or decrease in MAPE resulting from incorporating these $n = 2{,}000$ auxiliary samples, relative to using only the $m$ real samples. Data features are normalized before applying the methods. For each experimental condition, we also report the maximum p-value from all pairwise t-tests comparing the MAPE of $\hat{\betavec}^{\sf AAE}$ with those of the other estimators under the column $p_{\max}$. The test shows that $\hat{\betavec}^{\sf AAE}$ outperforms other estimators at the 90\% significance level for all instances, except for the ones marked with a star.      
    % This table reports the MAPE for AAE, the human-data-only estimator, PPI, and PPI++, obtained by augmenting $m$ real data points with $n = 2,000$ auxiliary data points using the census dataset. Results are averaged over 50 experimental runs. Data features are normalized before applying the methods. One-sample t-tests indicate that AAE outperforms the other methods at the 90\% confidence level, except for those marked with a star.
    \end{tabular}
  \end{minipage}}
\end{table}

\subsubsection{Synthetic Data Experiment}
\label{sec:empirical-2-aae-ppi-synthetic}
\begin{table}[!ht]
\renewcommand{\arraystretch}{1.3}
\centering
{\footnotesize
\caption{Bias Reduction from $\hat{\betavec}^{\sf P}$ under various dimensionality $d$(\%)}
\label{tab:aae-ppi-synthetic}
\begin{tabular}{cccccc}
\hline\hline
& $\hat{\betavec}^{\sf AAE}$ & $\hat{\betavec}^{\sf PPI}$ & $\hat{\betavec}^{\sf PPI++}$ & $p_{\max}$ \\\hline
$d=2$ & -17.12 & -14.01 & -17.40 & 0.53* \\
$d=4$ & -19.78 & -12.35 & -16.27 & 0.06 \\
$d=6$ & -61.41 & -29.39 & -34.64 & 3E-08 \\
$d=8$ & -145.71 & -5.95 & -66.36 & 2E-08 \\
$d=10$ & -171.98 & -80.86 & -115.00 & 5E-06    \\\hline\hline
\end{tabular}
\begin{minipage}{15.5cm}
\vspace{3pt}
    \footnotesize
    \begin{tabular}{p{1cm} p{13cm}}
         \textbf{Notes}: & \setlength{\baselineskip}{1.2\baselineskip} 
    % This table reports the MAPE for AAE, the human-data-only estimator, PPI, and PPI++, obtained by augmenting $m=100$ real data points with $n = 1,200$ auxiliary data points using the synthetic dataset with various feature dimension $d$. Results are averaged over 50 experimental runs. 
    For each estimator under each experimental condition, we perform 50 independent runs and report the average MAPE difference between the focal estimator and $\hat{\betavec}^{\sf P}$. In each condition, the estimators have access to a real dataset with $m=100$ samples and an auxiliary dataset containing $n = 1{,}000$ auxiliary samples. The reported average MAPE difference thus represents the increase or decrease in MAPE resulting from incorporating these $n = 1{,}200$ auxiliary samples, relative to using only the $m$ real samples. For each experimental condition, we also report the maximum p-value from all pairwise t-tests comparing the MAPE of $\hat{\betavec}^{\sf AAE}$ with those of the other estimators under the column $p_{\max}$. The test shows that $\hat{\betavec}^{\sf AAE}$ outperforms other estimators at the 95\% significance level for all instances, except for the ones marked with a star.      
    \end{tabular}
  \end{minipage}}
\end{table}

% \begin{table}[!ht]
% \renewcommand{\arraystretch}{1.3}
% \centering
% {\footnotesize 
% \caption{MAPE under various $d$ values using different methods(\%)}
% \label{tab:aae-ppi-synthetic}
% \begin{tabular}{c|ccccc}
% \hline\hline
% \multicolumn{1}{l|}{} & \multicolumn{5}{c}{MAPE(\%)}           \\
% Method               & $d=2$ & $d=4$ & $d=6$ & $d=8$ & $d=10$ \\\hline
% Human-data only      & 0.51  & 0.48  & 0.94  & 2.51  & 2.66   \\
% AAE                  & 0.34  & 0.28  & 0.32  & 0.97  & 0.86   \\
% PPI                  & 0.36  & 0.36  & 0.64  & 2.29  & 1.80   \\
% PPI++                & 0.33  & 0.32  & 0.59  & 1.78  & 1.47  \\\hline\hline
% \end{tabular}}
% \begin{minipage}{15.5cm}
% \vspace{3pt}
%     \footnotesize
%     \begin{tabular}{p{1cm} p{13cm}}
%          \textbf{Notes}: & \setlength{\baselineskip}{1.2\baselineskip} 
%     This table reports the MAPE for AAE, the human-data-only estimator, PPI, and PPI++, obtained by augmenting $m=100$ real data points with $n = 1,200$ auxiliary data points using the synthetic dataset with various feature dimension $d$. Results are averaged over 50 experimental runs. 
%     \end{tabular}
%   \end{minipage}
% \end{table}

We conduct a simulated study to compare AAE and PPI-based methods as the number of features varies. We generated synthetic datasets with varying number of feature $d$. Details of the synthetic data generation procedure and the experimental setups can be found in \cref{app:empirical-aae-ppi-synthetic}. \cref{tab:aae-ppi-synthetic} summarizes the model performances using these datasets. The results in \cref{tab:aae-ppi-synthetic} support our previous intuition: when the number of features $d$ is small (i.e., $\x$ is low-dimensional), the loss function of the outcomes retains much of the information about the mapping function, and the two methods perform similarly. However, as the dimensionality of $\x$ increases, AAE significantly outperforms PPI-based methods. These findings are consistent with our results from the previous section.

%% file: src/7_conclusion.tex
\section{General Discussion}
\label{sec:conclusion}

This paper presents a new approach for incorporating LLM-generated data into conjoint analysis, addressing the growing need for scalable and cost-effective methods in market research. While LLMs can mimic human-like responses, our study underscores the persistent errors and limitations inherent in directly using these AI-generated responses. Through our proposed data augmentation framework, which combines LLM-generated labels with real data, we demonstrate that it is possible to extract valuable insights from AI-generated data while mitigating errors that can distort market research outcomes.
Our theoretical framework, inspired by transfer learning and knowledge distillation, establishes a method to transfer the valuable but imperfect knowledge embedded in LLMs into a simpler, aligned model. This approach is validated empirically, where we show that our estimator not only reduces errors but also achieves significant data savings. Importantly, our findings highlight that while state-of-the-art LLMs, such as GPT-4, can improve the quality of AI-generated labels, their usefulness ultimately depends on how we integrate them with real data.

Looking ahead, our method opens the door for more effective use of LLMs in market research and beyond. As LLM technology advances, this framework can serve as a foundation for further innovations in data augmentation, helping researchers and practitioners balance the trade-offs between AI-generated and human-generated data. Future work may explore how this approach can be extended to other domains where real data is scarce, unlocking new applications for AI in understanding human preferences and behaviors. Future work could explore additional ways to enhance the usability of LLMs in combination with our methods. For instance, a promising research direction is examining the impact of incorporating different personas into LLM queries to generate data that more closely resembles a specific set of users. \revise{Finally, the framework extends well beyond the specific model and data-generation process considered here. For example, Appendix \ref{app:regression} discusses a continuous target in a regression setting. Additional extensions—such as accommodating general loss functions rather than just the KL-divergence or square loss—represent promising directions for future research.}

% while our paper focuses on conjoint analysis, the underlying theory applies more broadly to other knowledge distillation tasks involving LLMs, such as transferring knowledge from a teacher model to a student model. This opens up promising opportunities to generalize our ideas to other contexts and explore their statistical properties.

%% file: src/app_1_example.tex
\section{Comparison between the Covariances under a Specific Parametric Form}
\label{app:comparison}

Consider the specification of $\Prob(y = j\,|\,\x, z)$ given by 
\begin{align}
    g_j(\x, z; \thetavec^*) ~=~ \frac{e^{\check{\thetavec}^{\top}\x_{(j)}+ \eta\ind_{(z = j)}}}{1+\sum_{\ell \in \Kscr} e^{\check{\thetavec}^{\top}\x_{(\ell)}+ \eta\ind_{(z = \ell)}}}, \quad \forall j \in \Kscr.
    \label{eqn:mnl-g}
\end{align} 
Let us define the random variable
$
\rvec(\x, z) := \sum_{j\in \Kscr}\big(\sigma_{j}(\x; \betavec^*) 
 - g_{j}(\x, z; \thetavec^*)\big)\x_{(j)}
$. 
The next result follows. 

\begin{prop}[Projection Error Decomposition under MNL]
\label{prop:error-mnl}
    If \eqref{eqn:mnl-g} holds, under Assumptions \ref{assumption:regularity-consistency} and \ref{assumption:regularity-normality},  $\check{\Jvec} -  \Gammavec\Lambdavec\Gammavec^{\top}$ equals to:
  \begin{align}
       &\E\left[\rvec(\x, z)\rvec(\x, z)^{\top}\right]
        ~=~
     \underbrace{\E_{\x}\left[
      \left(\sum_{j\in \Kscr}\big(\sigma_{j}(\x; \betavec^*) 
 - \Prob(y = j\,|\,\x)\big)\x_{(j)}
     \right)\left(\sum_{j\in \Kscr}\big(\sigma_{j}(\x; \betavec^*) 
 - \Prob(y = j\,|\,\x)\big)\x_{(j)}
     \right)^{\top}
     \right]}_{\text{\upshape{The degree of model misspecification}}}\notag\\
     &~~~~~~~~~~~~~~~~~~~~+\underbrace{\E_{\x,z}\left[
      \left(\sum_{j\in \Kscr}\big(g_{j}(\x, z; \thetavec^*)
 - \Prob(y = j\,|\,\x)\big)\x_{(j)}
     \right)\left(\sum_{j\in \Kscr}\big(g_{j}(\x, z; \thetavec^*)
 - \Prob(y = j\,|\,\x)\big)\x_{(j)}
     \right)^{\top}
     \right]}_{\text{\upshape{The covariance of $\sum_{j\in \Kscr}g_{j}(\x, z; \thetavec^*))\x_{(j)}$}}}\succ 0.
     \label{eqn:decomposition-variance}
  \end{align}
\end{prop}

In this case, the difference $ \check{\Jvec} - \Gammavec\Lambdavec\Gammavec^{\top} $ decomposes neatly into two terms: one term reflects the extent to which the MNL choice model is misspecified, and the other term captures the covariance of $ \sum_{j \in \Kscr} g_{j}(\x, z; \thetavec^*) \x_{(j)} $,
because  
$
\E_{z}\left[
g_{j}(\x, z; \thetavec^*) \,\big|\,\x
\right]\,=\,\Prob(y = j\,|\,\x).
$
This decomposition offers valuable insights into the conditions under which AAE achieves variance reduction compared to $ \hat{\betavec}^{\sf P}$. 
Specifically, if model misspecification exists, so that the matrix from above is positive definite, or if the auxiliary label $ z $ is sufficiently informative, such that the associated covariance matrix is positive definite, then $\check{\Jvec} \succ \Gammavec\Lambdavec\Gammavec^{\top}$ holds, establishing AAE’s dominance. 
Furthermore, as model misspecification increases or $ z $ becomes more informative, this dominance is amplified, resulting in a larger variance reduction by AAE.

\begin{figure}[!ht]
    \centering
    \includegraphics[width=120mm]{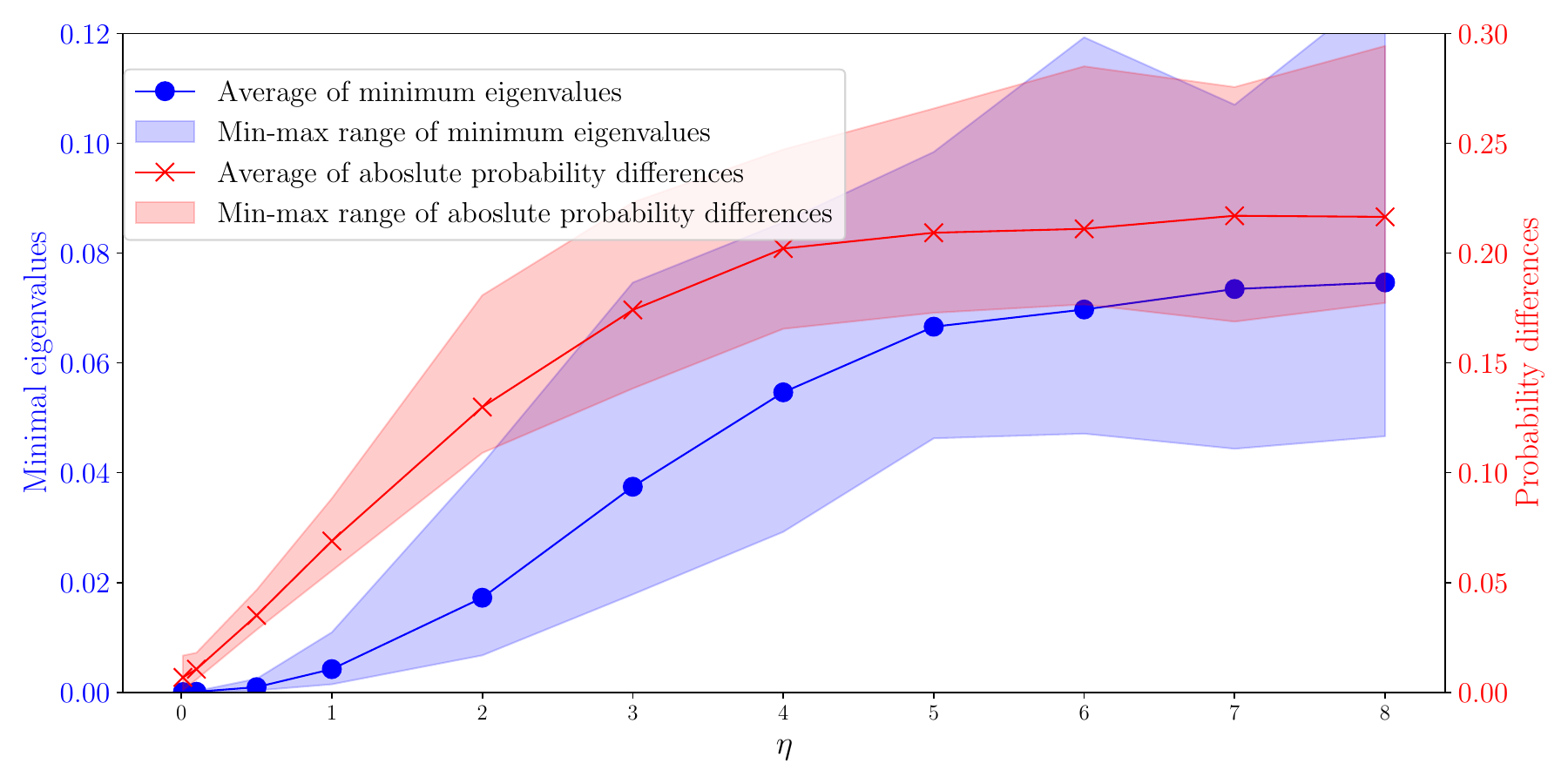}
    \caption{Minimum Eigenvalues and Absolute Probability Differences}
    \label{fig:eigen}
    \begin{minipage}{\textwidth}
    \footnotesize
    \begin{tabular}{p{1cm} p{15cm}}\hspace{-1mm}\textbf{Notes}: & \setlength{\baselineskip}{1.2\baselineskip} The plot shows averages of minimum eigenvalues (blue curve) and absolute probability differences (red curve) across 50 instances when $\eta$ changes.
         The ribbons give the maximum and minimum values for either the minimum eigenvalue or the absolute probability difference corresponding to each $\eta$ value. The vertical axes are adjusted for the minimum eigenvalue or the absolute probability difference corresponding for better visualization. 
    \end{tabular}
  \end{minipage}
\end{figure}

Using this proposition, we can compute $ \check{\Jvec} -  \Gammavec\Lambdavec\Gammavec^{\top} = (1 - \alpha)\alpha(2p - 1)^2 > 0 $ in the context of Example \ref{examp:ai-bias}, given that $ p \neq 1/2 $ and $ \alpha \notin \{0,1\} $. Note that, in this simplified setting, there is no model misspecification as in \eqref{eqn:decomposition-variance}. When $ p = 1/2 $, $z$ becomes pure noise, and when $ \alpha \in \{0,1\} $, $ z $ is constant. In both cases, $ z $ is uninformative. Outside of these cases, we find that AAE performs better than using only the primary data. The advantage becomes more evident if we send $p \rightarrow 0$ or $1$. 
In these cases, $z$ becomes more informative. 

Next, we expand on this discussion, and numerically illustrate $ \check{\Jvec} -  \Gammavec\Lambdavec\Gammavec^{\top} $ under the specification of \eqref{eqn:mnl-g}, 
and plot the results in Figure \ref{fig:eigen}. Specifically, we assume that 
$
\Prob(z = j \mid \x) = {e^{\bm{\zeta}^{\top}\x_{(j)}}}/\big(\sum_{\ell \in \Kscr} e^{\bm{\zeta}^{\top}\x_{(\ell)}}\big),\,\forall j\in \Kscr
$,
with $ d = 5 $ and $ k = 2 $.
Using the specification in \eqref{eqn:mnl-g}, we vary $ \eta $ across the set $ \{0.01, 0.1, 0.5, 1, 2, 3, 4, 5, 6, 7, 8\} $, generating 50 instances for each $ \eta $ value and calculating the minimum eigenvalue of $ \check{\Jvec} -  \Gammavec\Lambdavec\Gammavec^{\top} = \E\left[\rvec(\x, z)\rvec(\x, z)^{\top}\right] $. In each instance, we sample each component of $ \check{\thetavec} $ and $ \bm{\zeta} $ from $ U[-2, 2] $ and each component of $ \x $ from $ U[-1, 1] $, independently of one another.
In Figure \ref{fig:eigen}, we plot the average minimum eigenvalues across the 50 instances for each $ \eta $ (blue curve) alongside the average \textit{absolute probability differences} (red curve). The absolute probability difference here is defined as
\[
\E_{\x, z}\left[\frac{1}{k} \sum_{j \in \Kscr} \left| \sigma_{j}(\x; \betavec^*) - g_{j}(\x, z; \thetavec^*) \right|\right],
\]
which serves as an indicator of the distance between $ \sigma_{j}(\x; \betavec^*) $ and $ g_{j}(\x, z; \thetavec^*) $, thus intuitively reflecting the importance of $ z $.
Note that one can also use the KL-divergence as the metric, but we find its scale difficult to comprehend intuitively compared with the absolute probability difference. 

As $ \eta $ increases, the alignment between the random realizations of $ y $ and $ z $ also increases, signifying the growing value of the AI-generated label $ z $. This effect is illustrated in the red curve. 
The minimum eigenvalue also rises in a similar trend. Notably, the minimum eigenvalue remains consistently positive across all instances, even when $ \eta = 0.01 $, very close to zero. In this case, $z$ is nearly a noise; yet in the worst case across the 50 instances, the minimum eigenvalue is still approximately $ 1 \times 10^{-6} > 0 $, indicating that as long as $ z $ contains a minimal amount of information, we generally expect the second part of Proposition \ref{prop:weak-dom} to hold.

\subsection{Additional Proofs}

\noindent
\textbf{Proof of Proposition \ref{prop:error-mnl}.}
To begin, let us recall that $
\rvec ~=~ \sum_{j\in \Kscr}\big(\sigma_{j}(\x; \betavec^*) 
 - g_{j}(\x, z; \thetavec^*)\big)\x_{(j)}
$, where we suppress the dependency on $\x$ and $z$ in $\rvec$ for notation brevity. 
Recycling the notations used in the proof of Proposition \ref{prop:weak-dom}, we have 
$$\wvec~=~\sum_{j\in \Kscr}\big(\ind_{\{y = j\}} - \sigma_{j}(\x; \betavec^*)\big)\x_{(j)},
$$
and 
$$
\uvec = \nabla_{\thetavec} \log g_y(\x, z, \thetavec^*) = \sum_{j\in \Kscr}\big(\ind_{\{y = j\}} - g_{j}(\x, z; \thetavec^*)\big)\begin{pmatrix}
    \x_{(j)}\\\\\ind_{\{z = j\}}
\end{pmatrix}=
\begin{pmatrix}
    \tilde{\wvec}\\\\u_0
\end{pmatrix},
$$
where $\tilde{\wvec} := \sum_{j\in \Kscr}\big(\ind_{\{y = j\}} - g_{j}(\x, z; \thetavec^*)\big)\x_{(j)}$ and $u_0 :=\sum_{j\in \Kscr}\big(\ind_{\{y = j\}} - g_{j}(\x, z; \thetavec^*)\big)\ind_{\{z = j\}}$.
Therefore, $\rvec = \tilde{\wvec} - \wvec$. 
Also, recall that $$
\check{\Jvec} - \Gammavec\Lambdavec\Gammavec^{\top}~=~
\E\left[\wvec \wvec^{\top} \right] - \E\left[\wvec \uvec^{\top} \right]\E\left[\uvec \uvec^{\top} \right]^{-1}\E\left[\uvec \wvec^{\top}\right].
$$
We examine each term separately. 
We note 
\begin{align*}
 \wvec \uvec^{\top} = \wvec 
 \begin{pmatrix}
    \tilde{\wvec}^{\top} & u_0
 \end{pmatrix}
 = (\tilde{\wvec} - \rvec) 
 \begin{pmatrix}
    \tilde{\wvec}^{\top} & u_0
 \end{pmatrix}
 =  \begin{pmatrix}
    \tilde{\wvec}\tilde{\wvec}^{\top} & u_0\tilde{\wvec}   
 \end{pmatrix} - 
 \begin{pmatrix}
    \rvec\tilde{\wvec}^{\top} & u_0\rvec   
 \end{pmatrix},
\end{align*}
and 
$$
\uvec \uvec^{\top} ~=~
\begin{pmatrix}
    \tilde{\wvec}\tilde{\wvec}^{\top} & u_0\tilde{\wvec}\\
    u_0\tilde{\wvec}^{\top} & u_0^2
\end{pmatrix}.
$$
Thus, 
\begin{align*}
    &\E\left[\wvec \uvec^{\top} \right]\E\left[\uvec \uvec^{\top} \right]^{-1}\E\left[\uvec \wvec^{\top}\right]~=~
    \underbrace{\E\left[\begin{pmatrix}
    \tilde{\wvec}\tilde{\wvec}^{\top} & u_0\tilde{\wvec}   
 \end{pmatrix} \right]
 \E\left[
 \begin{pmatrix}
    \tilde{\wvec}\tilde{\wvec}^{\top} & u_0\tilde{\wvec}\\
    u_0\tilde{\wvec}^{\top} & u_0^2
\end{pmatrix}\right]^{-1}
\E\left[\begin{pmatrix}
    \tilde{\wvec}\tilde{\wvec}^{\top} \\ u_0\tilde{\wvec}^{\top}  
 \end{pmatrix} \right]}_{=:\Bvec_1}\\
 &-\underbrace{\E\left[\begin{pmatrix}
    \rvec\tilde{\wvec}^{\top} & u_0\rvec  
 \end{pmatrix} \right]
 \E\left[
 \begin{pmatrix}
    \tilde{\wvec}\tilde{\wvec}^{\top} & u_0\tilde{\wvec}\\
    u_0\tilde{\wvec}^{\top} & u_0^2
\end{pmatrix}\right]^{-1}
\E\left[\begin{pmatrix}
    \tilde{\wvec}\tilde{\wvec}^{\top} \\ u_0\tilde{\wvec}^{\top}  
 \end{pmatrix} \right]}_{=:\Bvec_2} - \Bvec_2^{\top} +
 \underbrace{\E\left[\begin{pmatrix}
    \rvec\tilde{\wvec}^{\top} & u_0\rvec  
 \end{pmatrix} \right]
 \E\left[
 \begin{pmatrix}
    \tilde{\wvec}\tilde{\wvec}^{\top} & u_0\tilde{\wvec}\\
    u_0\tilde{\wvec}^{\top} & u_0^2
\end{pmatrix}\right]^{-1}
\E\left[\begin{pmatrix}
    \tilde{\wvec}\rvec^{\top} \\ u_0\rvec^{\top}  
 \end{pmatrix} \right]}_{=:\Bvec_3}\\
\stackrel{\text{(a)}}{=}&
\E\left[\tilde{\wvec}\tilde{\wvec}^{\top} \right]
\end{align*}
To show (a), we argue as follows.
First of all, 
$$
\E\left[
\rvec\tilde{\wvec}^{\top}\right]~=~
\E_{ z}\left[\rvec\left(\sum_{j\in \Kscr}
\E\left[\ind_{\{y = j\}} - g_{j}(\x, z; \thetavec^*)|\x, z
\right]\x_{(j)}\right)^{\top}
\right] = \Ovec
$$
and similarly $\E\left[
u_0\rvec\right] = 0$ so $\Bvec_2 = \Bvec_3 = \Ovec$.
Also, 
$$
\Bvec_1 ~=~ \begin{pmatrix}
  \Ivec & \Ovec   
\end{pmatrix}
\E\left[\begin{pmatrix}
    \tilde{\wvec}\tilde{\wvec}^{\top} \\ u_0\tilde{\wvec}^{\top}  
 \end{pmatrix} \right]
~=~\E\left[\tilde{\wvec}\tilde{\wvec}^{\top} \right].
$$
This proves (a).

Further, 
$$
\E\left[\wvec \wvec^{\top} \right] ~=~ \E\left[(\tilde{\wvec} - \rvec)(\tilde{\wvec} - \rvec)^{\top} \right] ~\stackrel{\text{(b)}}{=}~
\E\left[\tilde{\wvec}\tilde{\wvec}^{\top} \right] + \E\left[\rvec\rvec^{\top} \right],
$$
where (b) follows because $\E\left[
\rvec\tilde{\wvec}^{\top}\right] = \Ovec$.
This proves that 
$\check{\Jvec} - \Gammavec\Lambdavec\Gammavec^{\top} = \E\left[\rvec\rvec^{\top} \right]$. 
Also, 
using the fact that 
$$
\rvec~=~\sum_{j\in \Kscr}\big(\sigma_{j}(\x; \betavec^*) 
 - \Prob(y = j\,|\,\x)\big)\x_{(j)} + \big(\Prob(y = j\,|\,\x) - g_{j}(\x, z; \thetavec^*)\big)\x_{(j)}
$$
and that $\E_{z}[g_{j}(\x, z; \thetavec^*)|\x] = \Prob(y = j\,|\,\x),$ 
we have the decomposition. 
\hfill $\blacksquare$
\vspace{3mm}

\vspace{3mm}

%% file: src/app_2_regression.tex
\section{AAE for Regression}
\label[app]{app:regression}

\revise{
In this section, we extend our approach to the regression setting. 
The setting is similar to our discussion in Section \ref{sec:setup}. 
Specifically, we still have two datasets, i.e., the primary data, 
$\left\{\big(\x^{\sf P}_{i}, y^{\sf P}_{i}, z^{\sf P}_{i}\big)\right\}_{i = 1}^m$, and auxiliary data $\left\{\big(\x_{i}, z_{i}\big)\right\}_{i = 1}^n$, with the label $y_i$ missing in the auxiliary data. 
Here we assume that these data points are i.i.d. generated according to the same distribution as the random vector $(\x, y, z)$, where $y, z \in \Real$ and $\x \in \Real^{d}$.
Similar to the discussion in Section \ref{sec:setup}, here the variable $z$ represents a predicted label generated by an AI model based on $\x$. 
With a continuous target variable $y$, we assume that $\E[y\,|\,\x, z] = g(\x, z; \thetavec^*)$, where $\thetavec^* \in \Real^q$.
We again seek a best-in-class estimator. Formally, we aim to estimate 
\begin{align}
\label{eqn:regression-best-in-class}
    \betavec^* ~\in~
    \argmin_{\betavec \in \Real^{d}} \E\bigg[\left(y - \x^{\top}\betavec\right)^2\bigg]~\stackrel{\text{(a)}}{=}~
    \E\bigg[\left(g(\x, z; \theta^*) - \x^{\top}\betavec\right)^2\bigg],
\end{align}
where one can verify (a) by expanding the square $\left(y - \x^{\top}\betavec\right)^2$ and using the fact hat $y-g(\x, z; \thetavec^*) \perp \x, z$.
Note that we do not assume that the regression model $\x^{\top}\betavec^*$ is correctly specified in the sense that it represents the conditional mean of $y$ given $\x$, which is a strong assumption. 
Instead, \eqref{eqn:regression-best-in-class} views it as the best linear predictor of $y$, or the projection of $y$ on the space of all linear functions of $\x$, and this is always possible. 
Our proposed estimator, the \textit{AI-augmented regression estimator} (AARE) works as follows: 

\vspace{10pt}

\begin{mdframed} \footnotesize \underline{{\sc AI-augmented Regresssion Estimator}} 

\noindent \textbf{Step 1.} 
Obtain an estimator $\hat{\thetavec}$ to $\thetavec^*$, using the primary data.

\noindent \textbf{Step 2.} With the auxiliary data, we construct the estimator $\hat{\betavec}^{\sf AARE}$ as $$\hat{\betavec}^{\sf AARE} ~\in~\argmin_{\betavec \in \Real^{d}}\,\left\{ \widehat{Q}(\hat{\thetavec};\betavec)\,=\, \frac{1}{n}\sum_{i=1}^n
\left(
g(\x_i, z_i; \hat{\thetavec}) - \x_i^{\top}\betavec\right)^2
\right\}.$$
\end{mdframed}

\vspace{10pt}

In the following we will analyze the properties of our AAE theoretically and compare it against a natural benchmark $\betavec^{\sf P} = \argmin_{\betavec \in \Real^d} \frac{1}{m}\sum_{i=1}^m
\left(
y_i - \x_i^{\top}\betavec\right)^2$, i.e., the esimator obtained using primary data only. 
We impose several technical assumptions.

\begin{assm}[Assumptions for AARE] 
\label{assumption:aare}
We assume the following:
\begin{itemize}
\setlength\itemsep{-0.3em}
    \item [$(i)$] It holds that 
    $\ddot{\mathbf{\Omega}} := \E[\x \x^{\top}] \succ 0$, $\hat{\thetavec} \pconverge \thetavec^*$, and $g(\x, z; \thetavec)$ is continuous in $\thetavec$ for all $\x$ and $z$. Furthermore, there is an open neighborhood of  $\theta^*$, $\Nscr$, such that $\,\E\big[\sup_{\thetavec \subset \Nscr}\lVert  \x g(\x, z; \thetavec) \rVert_2\big] \,<\, \infty$.  

    \item [$(ii)$] It holds that $\sqrt{m}\big(
    \hat{\thetavec} - \thetavec^*
    \big) \leadsto N(\bm{0}, \ddot{\mathbf{\Lambda}})$, $n/m \rightarrow \rho \in (0, \infty)$, $g(\x, z; \thetavec)$ is differentiable in $\thetavec$ for all $\x$ and $z$, and $\E\left[\sup_{\thetavec \subset \Nscr}\lVert  \x \nabla_{\theta} g(\x, z; \thetavec)^{\top} \rVert_2\right] < \infty$.

    \item [$(iii)$] Assume that $\E\Big[(y - g(\x, z; \thetavec^*))^2 \,\Big|\,\x, z \Big] = \sigma^2$ for all $\x$ and $z$, $\E\left[(g(\x, z; \thetavec^*) - y)\nabla_\theta g(\x, z; \thetavec^*)\right] = 0$, $\,\E\big[\sup_{\thetavec \subset \Nscr}\lVert  \nabla_{\thetavec} g(\x, z; \thetavec) \nabla_{\thetavec} g(\x, z; \thetavec)^{\top} \rVert_2\big] \,<\, \infty$ and $\,\E\big[\nabla_{\thetavec} g(\x, z; \thetavec^*) \nabla_{\thetavec} g(\x, z; \thetavec^*)^{\top}\big]$ is invertible. 
\end{itemize}    
\end{assm}

The first two items above are fairly standard in the statistics literature. All conditions in the last item are also standard for ensuring a conventional analysis of the asymptotic variance of $\hat{\thetavec}$ and are not restrictive, except for the additional requirement of homoscedasticity in our case.
We summarize our key theoretical development in the following theorem. 
We define $$\ddot{\Jvec} ~:=~ \E\left[ \left(g(\x, z; \thetavec^*) - \x^{\top}\betavec^*\right)^2 \x \x^{\top} \right],~
\ddot{\mathbf{\Gamma}}~:=~ \E\left[ \x\nabla g(\x, z; \thetavec^*)^{\top} \right]
,~\text{ and }~
\Sigmavec ~:=~ \E
\left[
(y - \x^{\top}\betavec^*)^2 \x\x^{\top}
\right].
$$
For notational convenience in the statement of the theorem. 

\begin{thm}[Properties of AARE]
\label{thm:aare}
~
\begin{itemize}
    \item [$(i)$] Under Assumption \ref{assumption:aare} item (i), the optimizer  
    $\betavec^*$ defined in \eqref{eqn:regression-best-in-class} is unique and the AARE satisfies 
    $\hat{\betavec}^{\sf AARE} \pconverge \betavec^*$, when $m,n\rightarrow \infty$.

    \item [$(ii)$] Under items (i) and (ii) of Assumption \ref{assumption:aare}, $
    \sqrt{m}(\hat{\betavec} - \betavec^*) \rightsquigarrow N\Big(\Ovec,\, \ddot{\Omegavec}^{-1}\big(\ddot{\Jvec} + \rho\times\ddot{\Gammavec}\ddot{\Lambdavec}\ddot{\Gammavec}^{\top}\big)\ddot{\Omegavec}^{-1}\Big).
    $

    \item [$(iii)$] Assume all items of Assumption \ref{assumption:aare}, it follows that $\sqrt{m}(\hat{\betavec}^{\sf P} - \betavec^*)\rightsquigarrow \ddot{\Omegavec}^{-1}\Sigmavec \ddot{\Omegavec}^{-1}$. Also, $\Sigmavec \succeq \ddot{\Gammavec}\ddot{\Lambdavec}\ddot{\Gammavec}^{\top}.$
\end{itemize}
\end{thm}

We note that the first two parts show that in the regression setting, AARE has the desired asymptotic properties. 
The last result is analogous to 
Proposition  \ref{prop:weak-dom}. 
AARE dominates the primary-data-only estimator provided that $g(\cdot)$ has a linear structure.} 

\subsection{Additional Proofs}

\revise{
\noindent
\textbf{Proof of Theorem \ref{thm:aare}.}
We will first prove the first item of Theorem \ref{thm:aare}.
Whenever $\sum_{i=1}^n\x_i \x_i^{\top}$ is invertible, it holds that 
$$
\hat{\betavec}^{\sf AARE} ~=~ \left(\frac{1}{n}\sum_{i=1}^n\x_i \x_i^{\top}\right)^{-1} \cdot\frac{1}{n}\sum_{i=1}^n\x_i g(\x_i, z_i; \hat{\thetavec}). 
$$
Since $\ddot{\mathbf{\Omega}} = \E[\x \x^{\top}] \succ 0$, it holds that $\frac{1}{n}\sum_{i=1}^n \x_i \x_i^{\top}\pconverge \ddot{\mathbf{\Omega}}$, so one can verify that 
\begin{align*}
    \hat{\betavec}^{\sf AARE} ~&=~ \ddot{\mathbf{\Omega}}^{-1} \cdot\frac{1}{n}\sum_{i=1}^n\x_i g(\x_i, z_i; \hat{\thetavec}) + o_{\text{P}}(1)\\
    ~&=~ \betavec^* + \ddot{\mathbf{\Omega}}^{-1}\cdot \frac{1}{n}\sum_{i=1}^n\x_i \bigg[
    \left(g(\x_i, z_i; \hat{\thetavec})-g(\x_i, z_i, {\thetavec}^*)\right)+
    \left(g(\x_i, z_i, {\thetavec}^*) - \x_i^\top \betavec^*\right)\bigg] + o_{\text{P}}(1).
\end{align*}
Note that the first-order condition of \eqref{eqn:regression-best-in-class} implies that $\E\bigg[\left(g(\x, z; \theta^*) - \x^{\top}\betavec^*\right)\x\bigg] = 0$ so $\frac{1}{n}\sum_{i=1}^n\left(g(\x_i, z_i, {\thetavec}^*) - \x_i^\top \betavec^*\right)\x_i \pconverge 0$.
Furthermore, since $\hat{\thetavec} \pconverge \thetavec^*$, it holds that there is a sequence of positive constants $\{\epsilon_n\}_{n=1}^\infty$ that converges to zero such that $\Prob\left(
\big \lVert
\hat{\thetavec} - \thetavec^*
\big \rVert > \epsilon_n
\right) \rightarrow 0$.
By Markov's inequality, for any fixed $\epsilon > 0$, 
\begin{align}
\label{eqn:chain-x-times-g}
    & \Prob\left(\left\lVert\frac{1}{n}\sum_{i=1}^n\x_i 
    \left(g(\x_i, z_i; \hat{\thetavec})-g(\x_i, z_i, {\thetavec}^*)\right)\right \rVert\geq \epsilon\right)\\
    &~~~~~~~~~~~~~~~~~~\leq~\Prob\left(\sup_{{\thetavec}: \big \lVert
{\thetavec} - \thetavec^*
\big \rVert \leq \epsilon_n}\left\lVert\frac{1}{n}\sum_{i=1}^n\x_i 
    \left(g(\x_i, z_i, {\thetavec})-g(\x_i, z_i, {\thetavec}^*)\right)\right \rVert\geq \epsilon\right) + \Prob\left(
\big \lVert
\hat{\thetavec} - \thetavec^*
\big \rVert > \epsilon_n
\right) \\
&~~~~~~~~~~~~~~~~~~\leq~ 
\frac{1}{\epsilon}\E\left[\sup_{{\thetavec}: \big \lVert
{\thetavec} - \thetavec^*
\big \rVert \leq \epsilon_n}\left\lVert\frac{1}{n}\sum_{i=1}^n\x_i 
    \left(g(\x_i, z_i, {\thetavec})-g(\x_i, z_i, {\thetavec}^*)\right)\right \rVert\right]
+ \Prob\left(
\big \lVert
\hat{\thetavec} - \thetavec^*
\big \rVert > \epsilon_n
\right) \\
&~~~~~~~~~~~~~~~~~~\stackrel{\text{(a)}}{\leq}~ 
\frac{1}{\epsilon}\E\left[\sup_{{\thetavec}: \big \lVert
{\thetavec} - \thetavec^*
\big \rVert \leq \epsilon_n}\big\lVert \x 
    \left(g(\x, z, {\thetavec})-g(\x, z, {\thetavec}^*)\right)\big \rVert\right]
+ \Prob\left(
\big \lVert
\hat{\thetavec} - \thetavec^*
\big \rVert > \epsilon_n
\right), 
\end{align}
where (a) follows by triangular inequality and Jensen's inequality. 
By the continuity of $g(\cdot)$ in $\thetavec$ as stated in the assumption, it follows that $\sup_{{\thetavec}: \big \lVert
{\thetavec} - \thetavec^*
\big \rVert \leq \epsilon_n}\big\lVert \x 
    \left(g(\x, z, {\thetavec})-g(\x, z, {\thetavec}^*)\right)\big \rVert$ converges to zero when $n$ goes to infinity. 
    Since $\,\E\big[\sup_{\thetavec \subset \Nscr}\lVert  \x g(\x, z; \thetavec) \rVert_2\big] \,<\, \infty$ and $\epsilon_n$ falls into $\Nscr$ eventually, by the dominated convergence theorem, we have that $\E\left[\sup_{{\thetavec}: \big \lVert
{\thetavec} - \thetavec^*
\big \rVert \leq \epsilon_n}\big\lVert \x 
    \left(g(\x, z, {\thetavec})-g(\x, z, {\thetavec}^*)\right)\big \rVert\right]$ converges to zero. 
    This concludes the proof for the first item. 

    For the second item, we note that 
    \begin{align*}
        & \sqrt{n}\left(\hat{\betavec}^{\sf AARE} - \betavec^* \right)~=~ \ddot{\mathbf{\Omega}}^{-1} \cdot\frac{1}{\sqrt{n}}\sum_{i=1}^n\x_i \left(g(\x_i, z_i; \tilde{\thetavec}) - \x_i^{\top} \betavec_i\right) + o_{\text{P}}(1)\\&~~~~~~~~~~~~~~=~ \ddot{\mathbf{\Omega}}^{-1} \cdot\left[\frac{1}{\sqrt{n}}\sum_{i=1}^n\x_i \left(g(\x_i, z_i; {\thetavec}^*)
        - \x_i^\top \betavec^*
        \right)+ \frac{1}{n}\sum_{i=1}^n\x_i \nabla g(\x_i, z_i; \hat{\thetavec})^{\top}\times 
        \sqrt{\frac{n}{m}} \times \sqrt{m}\big(
    \hat{\thetavec} - \thetavec^*
    \big)\right] + o_{\text{P}}(1),
    \end{align*}
    where $\tilde{\thetavec} \in [\thetavec^*, \hat{\thetavec}]$.
    Using an argument similar to \eqref{eqn:chain-x-times-g}, we can show that  
    $$\frac{1}{n}\sum_{i=1}^n\x_i \nabla g(\x_i, z_i; \hat{\thetavec})^{\top} \pconverge \E\left[\x \nabla g(\x, z;{\thetavec}^*)^{\top}  \right].$$
    Therefore, the conclusion follows. 

    For the third item, we first note that 
    \begin{align*}
        \Sigmavec &=
\E
\left[
\left[(y - \psivec(\x, z)^{\top}\thetavec^*) + (\psivec(\x, z)^{\top} \thetavec^*- \x^{\top}\betavec)\right]^2 \x\x^{\top}
\right]\\
&\stackrel{\text{(b)}}{=}
\E
\left[(y - \psivec(\x, z)^{\top}\thetavec^*)^2 \x\x^{\top}
\right] + \E
\left[(\psivec(\x, z)^{\top} \thetavec^*- \x^{\top}\betavec)^2 \x\x^{\top}
\right]\stackrel{\text{(c)}}{=}\sigma^2\E
\left[\x\x^{\top}
\right] + \E
\left[(\psivec(\x, z)^{\top} \thetavec^*- \x^{\top}\betavec)^2 \x\x^{\top}
\right],
    \end{align*}
where (b) follows because 
$\E[y - \psivec(\x, z)^{\top} \thetavec^* \,|\, \x, z] = 0$ and (c) follows from the homoscedasticity assumption.
Following standard analysis of M-estimators \citep{newey1994large}, under the current conditions, one can compute the  asymptotic variance of 
$\hat{\thetavec}$ as 
$\ddot{\bm{\Lambda}} = \bm{\Lambda}_1^{-1}\bm{\Lambda}_2  \bm{\Lambda}_1^{-1}$. 
Here
$$
\bm{\Lambda}_1 ~=~ \E\Big[
\nabla g(\x, z; \thetavec^*)\nabla g(\x, z; \thetavec^*)^{\top} + 
\left(g(\x, z; \thetavec^*) - y \right)
\nabla^2_{\thetavec} g(\x, z; \thetavec^*)
\Big]~\stackrel{\text{(d)}}{=}~
\E\left[
\nabla_{\thetavec} g(\x, z; \thetavec^*)\nabla_{\thetavec} g(\x, z; \thetavec^*)^{\top}\right], 
$$
where (d) follows because $g(\x, z; \thetavec^*) = \E[y\,|\,\x, z]$, and 
$$
\bm{\Lambda}_2 ~=~ \E
\left[
\left(y - g(\x, z; \thetavec^*)\right)^2\nabla_{\thetavec} g(\x, z; \thetavec^*)\nabla_{\thetavec} g(\x, z; \thetavec^*)^{\top} 
\right]~\stackrel{\text{e}}{=}~\sigma^2\E
\left[
\nabla_{\thetavec} g(\x, z; \thetavec^*)\nabla_{\thetavec} g(\x, z; \thetavec^*)^{\top}
\right],
$$
where (e) follows from the homoscadesticity assumption. 
Therefore, 
$\ddot{\Lambdavec} =\sigma^2 \E\left[\nabla_{\thetavec} g(\x, z; \thetavec^*)\nabla_{\thetavec} g(\x, z; \thetavec^*)^{\top}\right]^{-1}$. 
Then, it holds that 
$$
\Sigmavec ~=~
\sigma^2\E
\left[\x\x^{\top}
\right] ~\succeq~ \ddot{\Gammavec}\ddot{\Lambdavec}\ddot{\Gammavec}^{\top}$$ using the same argument in Proposition  \ref{prop:weak-dom} (i.e., setting $\uvec = \x$ and $\wvec = \nabla_{\thetavec} g(\x, z; \thetavec^*)$). 
Thus, the conclusion follows.
\hfill $\blacksquare$
\vspace{3mm}}

%% file: src/app_3_empirical.tex
\section{Supplemental Materials for the Empirical Studies}
\label[app]{app:empirical}
In this section, we present supplemental materials for the empirical studies in \cref{sec:empirical-1,sec:empirical-2}.
\subsection{Efficient Frontier Diagram of Results in \texorpdfstring{\cref{sec:empirical-1-economic-value}}{}}
\begin{figure}[!t]
    \centering
    \includegraphics[width=0.6\textwidth]{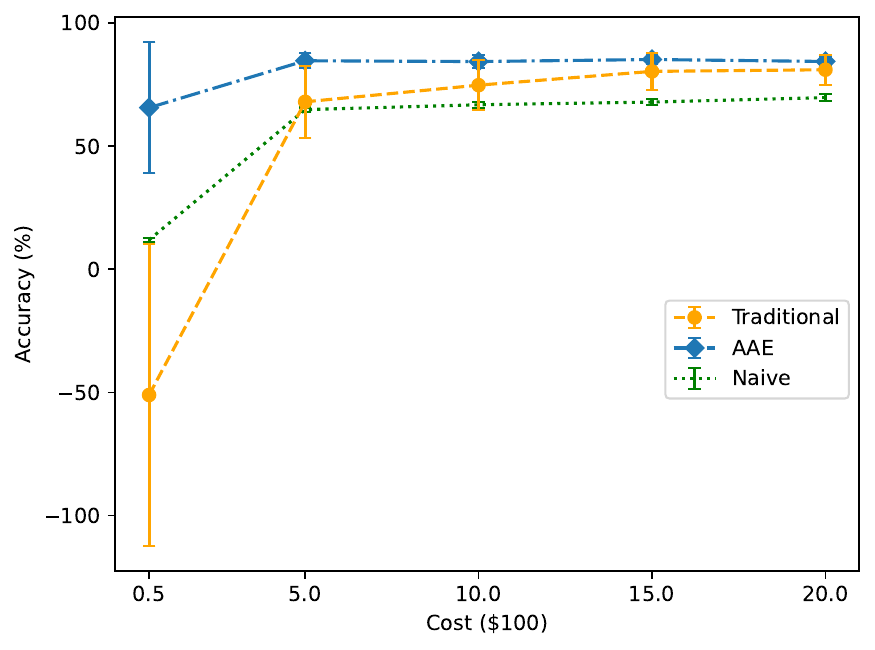}
    \caption{Conjoint estimation accuracy vs. market research costs}
    \label{fig:efficient-frontier}
    \begin{minipage}{\textwidth}
    \footnotesize
    \begin{tabular}{p{1cm} p{15cm}}
         \hspace{-1mm}\textbf{Notes}: &  \setlength{\baselineskip}{1.2\baselineskip} The plot shows the parameter estimation accuracy (1 - MAPE) relative to the total costs of hiring human subjects with the traditional market research estimator $\hat{\bm{\beta}}^{\sf P}$, the na\"ive estimator $\hat{\bm{\beta}}^{\sf P}$, and the AAE estimator $\hat{\bm{\beta}}^{\sf AAE}$. The average accuracy and error bars, which show the standard deviations,  are computed over 50 random experimental runs with the same setup in \cref{subsec:empirical-setup-1}.
\end{tabular} \end{minipage}
\end{figure}

Additional to the results in \cref{sec:empirical-1-economic-value}, we calculate the efficient frontier between estimation accuracy and recruitment costs using our methods compared to human-only data. For this analysis, we assume that recruiting each human subject costs $50$ USD, while querying the LLM incurs no cost. \cref{fig:efficient-frontier} illustrates how AAE extends the efficient frontier for conjoint market research. The yellow line represents the parameter estimation accuracy relative to the total costs of hiring human subjects, indicating the current trade-off between costs and accuracy for traditional conjoint analysis. The blue line shows the accuracy achieved with AAE, where AI-generated labels supplement the corresponding human data, significantly pushing the frontier outward. Additionally, the green line represents the accuracy of na\"ive augmentation with AI data. As observed, na\"ive utilization of AI data does not enhance efficiency. This figure further underscores the practical value of AAE, especially when the sample size is small. We note that, although the sample size may not be small at the population level, market research often focuses on responses from specific demographic groups, where sample sizes are typically smaller. This increases the value of our estimators significantly.

\subsection{Conjoint Data Generation using GPT for the Sports Car Dataset}
\label[app]{app:empirical-dgp-car}
In the following, we illustrate the conjoint data generation procedure with the sports car conjoint setting. The models and prompting follow the same setup as in \cref{subsec:data-generation}. In particular, we show how the previously established conjoint data generation framework can be applied in other settings. 

\subsubsection{Basic Prompting.}
In this part, we present an example of basic prompting following the three-step framework for the sports car data:

\vspace{10pt}
\begin{mdframed}
{\footnotesize
\textit{Input (I)}: You should act like a random person purchasing a sports car.\\
\noindent \textit{Input (II)}: There are three options for you:\\
A: Type of car model: Basic. Number of seats: 2. Type of transmission: Manual. Convertible roofs: Yes. Price: \$30k\\
B: Type of car model: Basic. Number of seats: 4. Type of transmission: Manual. Convertible roofs: Yes. Price: \$40k \\
C: Type of car model: Basic. Number of seats: 4. Type of transmission: Auto. Convertible roofs: Yes. Price: \$35k \\
\noindent\textit{Input (III)}:  Which option would you choose as a random person? Your response should be a single letter A, B, or C.\\
\noindent\textit{Output}: B
}
\end{mdframed}
\vspace{10pt}

\subsubsection{CoT Prompting.}
In CoT prompting, we allow GPT to compare the features of cars systematically before reaching a final decision in part \textit{(III)}. An example of the CoT-based conversation is provided below. As illustrated, GPT demonstrates more personal insight when selecting sports cars compared to the vaccine context, with less clearly defined preference rankings involved in comparing each feature.

\vspace{15pt}

\begin{mdframed}
{\footnotesize
\textit{Input}: Compare the number of seats of A, B, and C, which one would you prefer?\\
\textit{Output}: Well, I think I'd like option B. You know, the idea of cruising around in a sports car sounds fantastic, but practically speaking, having four seats could be a real benefit. I mean, the chances of me going to a sports event or on a road trip with friends are high. So, the extra seats could come handy then. Plus, it's not that much costly either. Only \$40k, right? I think it's worth the price.
\\
\textit{Input}: Compare the type of transmission of A, B, and C, which one would you prefer?\\
\textit{Output}: To be honest, I've always enjoyed the feeling of control I get with a manual transmission. It makes driving more engaging for me. So, I'm leaning more towards options A or B. But if I consider the number of seats as well...you know, having room for more friends... I'd have to say option B is looking like a better fit for me.\\
\textit{Input}: Compare the roof of A, B, and C, which one would you prefer?\\
\textit{Output}: All the cars A, B, and C have convertible roofs, which is fantastic. I love the idea of feeling the wind in my hair while cruising down the highway on a sunny day. So, considering the other factors, I'm still leaning toward option B. Hard to beat the full experience it offers - a manual transmission, room for friends, and a convertible roof!\\
\textit{Input}: Compare the price of A, B, and C, which one would you prefer?\\
\textit{Output}: Option A is the least expensive at \$30k, which is quite tempting. But, I think I'd still stick with option B. It might be \$10k more, but it also offers two extra seats and the joy of manual driving, which fit my preferences better. I see it as an investment in fun experiences with friends and personal enjoyment when driving. It's looking like a worthwhile expenditure.\\
\textit{Input}: Based on all the previous reasonings, which option would you choose?\\
\textit{Output}: Based on all the comparisons around the number of seats, transmission type, roof, and price, I would choose Option B. It might be a bit more expensive, but it fits my needs in terms of the extra seating, manual transmission, and the awesome convertible roof.\\
\textit{Input}: Return the choice in the previous answer as a single letter A, B, or C.\\
\textit{Output}: B
}
\end{mdframed}

\subsection{Synthetic Data Generation in \texorpdfstring{\cref{sec:empirical-2-aae-ppi-synthetic}}{}}
\label[app]{app:empirical-aae-ppi-synthetic}
We first generate a MNL ground truth model with $d \in \{2,4,6,8,10\}$ feature variables (not including the intercept) and $k=1$ option. The ground truth parameter $\bm{\beta}^*$ is randomly generated standard multivariate normal distribution. We randomly generate a dataset with $1,200$ data points based on the MNL ground truth model. For each real data sample $(\x_i, y_i)$, we generate the auxiliary label by adding a random perturbation to the real label. Specifically, $z_i = y_i$ with probability 0.9 and $z_i = 1 - y_i$ with probability 0.1. 

% \subsection{Synthetic Data Generation in \texorpdfstring{\cref{sec:empirical-2-misspecification}}{}}
% \label[app]{app:empirical-2-misspecification}
% We first generate a non-parametric ground truth model with $d \in \{10,20,30,40\}$ binary feature variables (not including the intercept) and $k=1$ option. We generated a total of $\sqrt{d}$ clusters. Each cluster is centered at $\mathbf{w}$, where $\mathbf{w}$ is randomly sampled from the uniform distribution over $\{-1,0,1\}^d$. Each cluster correspond to a choice from $\{0,1\}$. For each $\x$, we compute $\x^{\sf diff} = \x_{(1)} - \x_{(2)}$ and find its closest cluster center $\mathbf{w}$ with respect to the $l-1$ norm. The choice $y$ given $\x$ is assigned as the choice corresponding to this cluster. We generate a data set with $1,200$ data points by randomly sampling from the ground truth model. For each real data sample $(\x_i, y_i)$, we generate the auxiliary label by adding a random perturbation to the real label. Specifically, $z_i = y_i$ with probability 0.7 and $z_i = 1 - y_i$ with probability 0.3. 
% \color{black}
\subsection{Empirical Results Evaluated by the MSE Metric}
\label[app]{app:empirical-mse}

\begin{table}[!ht]
\renewcommand{\arraystretch}{1.3}
\centering
{\footnotesize
\caption{Bias Reduction with MSE in Empirical Setting I}
\label{tab:bias-reduction-mse}
\begin{tabular}{cccccccccc}\hline\hline
& & \multicolumn{4}{c}{$m=50$} & \multicolumn{4}{c}{$m=100$} \\\hline
\textbf{Model}  & \textbf{Prompt} & $\hat{\betavec}^{\sf A}$ & $\hat{\betavec}^{\sf Naive}$  & $\hat{\betavec}^{\sf AAE}$  & $p_{\max}$ & $\hat{\betavec}^{\sf A}$& $\hat{\betavec}^{\sf Naive}$  & $\hat{\betavec}^{\sf AAE}$   & $p_{\max}$ \\\hline
\multirow{2}{*}{\texttt{GPT-3.5-Turbo-0613}} & Basic & -8.15 & -15.15 & -18.18 & 3e-10 & 3.05 & -4.65 & -6.88 & 2e-06 \\
 & COT & 36.84 & 28.92 & -19.14 & 9e-11 & 48.04 & 33.88 & -7.86 & 1e-09 \\
\multirow{2}{*}{\texttt{GPT-3.5-Turbo-0125}} & Basic & -13.62 & -14.86 & -17.85 & 5e-10 & -2.42 & -4.00 & -6.81 & 1e-07 \\
 & COT & 18.62 & 10.52 & -19.25 & 1e-10 & 29.82 & 16.76 & -7.69 & 1e-08 \\
\multirow{2}{*}{\texttt{GPT-4}} & Basic & 42.76 & 37.78 & -20.24 & 2e-11 & 53.96 & 42.71 & -8.43 & 2e-10 \\
 & COT & 49.96 & 40.15 & -20.05 & 4e-11 & 61.15 & 44.53 & -8.21 & 5e-10 \\
\multirow{3}{*}{\texttt{GPT-4o}} & Basic & 40.16 & 33.81 & -19.94 & 3e-11 & 51.35 & 39.53 & -8.13 & 3e-10 \\
 & COT & 41.53 & 32.26 & -20.13 & 3e-11 & 52.73 & 37.12 & -8.03 & 8e-10 \\
 & FS & 44.72 & 37.47 & -20.37 & 8e-12 & 55.92 & 43.50 & -8.38 & 4e-10 \\
\texttt{GPT-4o Fine-tuned} & Basic & 36.81 & 31.37 & -20.81 & 4e-12 & 48.01 & 35.79 & -9.44 & 5e-12 \\\hline
& & \multicolumn{4}{c}{$m=150$} & \multicolumn{4}{c}{$m=200$} \\\hline
\textbf{Model}  & \textbf{Prompt} & $\hat{\betavec}^{\sf A}$& $\hat{\betavec}^{\sf Naive}$  & $\hat{\betavec}^{\sf AAE}$   & $p_{\max}$ & $\hat{\betavec}^{\sf A}$& $\hat{\betavec}^{\sf Naive}$  & $\hat{\betavec}^{\sf AAE}$   & $p_{\max}$ \\\hline
\multirow{2}{*}{\texttt{GPT-3.5-Turbo-0613}} & Basic & 9.17 & -1.52 & -1.22 & 0.816* & 10.13 & -1.07 & -0.18 & 0.999* \\
 & COT & 54.16 & 32.80 & -2.23 & 8e-04 & 55.12 & 28.44 & -0.92 & 0.015 \\
\multirow{2}{*}{\texttt{GPT-3.5-Turbo-0125}} & Basic & 3.70 & 0.53 & -1.23 & 0.035 & 4.66 & 1.78 & -0.12 & 0.401* \\
 & COT & 35.95 & 18.67 & -1.96 & 0.003 & 36.91 & 17.67 & -0.92 & 0.022 \\
\multirow{2}{*}{\texttt{GPT-4}} & Basic & 60.08 & 45.44 & -2.96 & 4e-05 & 61.05 & 41.60 & -1.91 & 7e-06 \\
 & COT & 67.28 & 45.76 & -2.66 & 1e-04 & 68.24 & 40.20 & -1.50 & 3e-04 \\
\multirow{3}{*}{\texttt{GPT-4o}} & Basic & 57.48 & 41.23 & -2.61 & 2e-04 & 58.44 & 37.61 & -1.31 & 3e-04 \\
 & COT & 58.86 & 37.42 & -2.59 & 2e-04 & 59.82 & 32.62 & -1.42 & 8e-04 \\
 & FS & 62.04 & 45.43 & -3.12 & 1e-05 & 63.00 & 41.61 & -1.71 & 2e-05 \\
\texttt{GPT-4o Fine-tuned} & Basic & 54.13 & 38.74 & -4.20 & 2e-08 & 55.09 & 34.00 & -2.73 & 5e-10
\\\hline\hline
\end{tabular} 
\begin{minipage}{15.5cm}
    \footnotesize
    \vspace{3pt}
    \begin{tabular}{p{1cm} p{14cm}}
         \textbf{Notes}: & \setlength{\baselineskip}{1.2\baselineskip} 
    This table presents the change in the MSE after augmented $n=1,000$ GPT-generated samples to a real set with $m$ samples with the AI-only estimator $\hat{\betavec}^{\sf A}$, na\"ive estimator $\hat{\betavec}^{\sf Naive}$ and the AAE estimator $\hat{\betavec}^{\sf AAE}$, respectively, averaged over 50 experimental runs. As in \cref{sec:empirical-1}, AAE significantly reduces error from the primary set, and outperforms the AI-only and na\"ive augmentation estimators. Pairwise t-tests show that $\hat{\betavec}^{\sf AAE}$ outperforms $\hat{\betavec}^{\sf P}$, $\hat{\betavec}^{\sf A}$, and $\hat{\betavec}^{\sf Naive}$ at the 95\% significance level for all instances, except for the ones marked with a star. The maximum p-value of the pairwise t-tests are shown under the column $p_{\max}$.
    \end{tabular}
  \end{minipage}
}
\end{table}
In this section, we present a robustness check for the results in \cref{sec:empirical-1} with an alternative evaluation metric, i.e., the Mean Squared Error (MSE). The empirical setups are the same as \cref{sec:empirical-1}. 
In particular, to assess the performance of the estimators, we calculated the MSE aggregated across all features as:
\begin{equation*}
    {\sf MSE}^l ~=~ \frac{1}{d}\sum_{j=1}^d (\hat{\beta}^l_j - \beta^*_j)^2 \,,~~~ \forall l \,\in\, \{{\sf P}, {\sf A}, {\sf Naive}, {\sf AAE}\}\,.
\end{equation*}
For each combination of $(m, n)$, we conducted 50 independent experimental runs and averaged the MSE as the final performance indicator. In the subsequent discussion, as in the main text, we use ${\sf MSE}^{\sf P}$ as the benchmark, and report the change in MSE to compare different estimators. For example, if ${\sf MSE}^{\sf AAE} - {\sf MSE}^{\sf P}$ is a negative number, the experiment then suggests that the AAE outperforms the primary-data-only estimator. 

\cref{tab:bias-reduction-mse} summarizes the change in MSE of $\hat{\betavec}^{\sf A}$, $\hat{\betavec}^{\sf Naive}$, and $\hat{\betavec}^{\sf AAE}$, using $\hat{\betavec}^{\sf P}$ as the benchmark. Similarly to the findings in \cref{sec:empirical-1}, AAE significantly reduces MSE in all scenarios, achieving the best performance using fine-tuned \texttt{GPT-4o}, while the AI-only or na\"ive method tends to elicit more errors.
In general, AAE's performance improves with higher versions of GPT and more advanced prompting methods. Na\"ive or AI-only methods do not exhibit a consistent correlation with higher versions of GPT and could induce more errors with more sophisticated prompting. Moreover, fine-tuning \texttt{GPT-4o} does not improve the performance of na\"ive or AI-only methods.

We evaluated the percentage of data saved using AAE with various GPT models based on the MSE reduction. Specifically, similar to the main text, for a given MSE reduction achieved by applying AAE to a real dataset of size $n_1$, we calculated the amount of real data samples, $n_2$, required to achieve the same error reduction without AI augmentation. The percentage of data saved is then estimated as $((n_2 - n_1)/n_2) \times 100\%$. \cref{tab:saving-mse} summarizes the percentage of data that can be saved using AAE. Results show our method saves up to 77.7\% of data across different models, prompts, and primary data sizes. When the primary set is small ($m=50$), AAE saves between 67.7\% and 77.7\% of data samples, regardless of the GPT model or prompt design. With a moderate primary set size ($m=100, 150$), savings range from 12.8\% to 54.5\%, again consistent across GPT models and prompts. These results verifies the robustness of our findings in \cref{sec:empirical-1}.

\begin{table}[!h]
\renewcommand{\arraystretch}{1.3}
\centering
{\footnotesize
\caption{Percentage of Saving in Data Size Evaluated by MSE in Empirical Setting I (\%)}
\label{tab:saving-mse}
\begin{tabular}{cccccc}\hline\hline
Model    & Prompt & $m=50$          & $m=100$         & $m=150$         &$m=200$         \\\hline
\multirow{2}{*}{GPT-3.5-Turbo-0613} & Basic & 69.13 (0.47) & 37.53 (1.32) & 12.79 (1.68) & -17.93 (2.47)* \\
 & COT & 73.58 (1.54) & 45.95 (2.64) & 25.21 (3.24) & -5.31 (4.47)* \\
\multirow{2}{*}{GPT-3.5-Turbo-0125} & Basic & 67.72 (0.64) & 36.97 (1.26) & 12.91 (1.59) & -18.98 (2.57)* \\
 & COT & 74.25 (1.12) & 44.75 (1.94) & 22.71 (2.49) & -3.78 (3.38)* \\
\multirow{2}{*}{GPT-4} & Basic & 77.33 (1.60) & 49.29 (2.85) & 33.31 (3.29) & 10.44 (3.96) \\
 & COT & 76.36 (1.84) & 47.92 (3.27) & 29.47 (3.27) & 3.60 (4.32)* \\
\multirow{3}{*}{GPT-4o} & Basic & 76.21 (1.76) & 47.07 (3.12) & 28.60 (3.51) & 0.58 (5.12)* \\
 & COT & 76.71 (1.53) & 46.94 (2.74) & 29.92 (2.78) & 3.15 (4.13)* \\
 & FS & 77.15 (1.69) & 48.78 (3.26) & 34.86 (3.24) & 6.80 (5.02) \\
GPT-4o Fine-tuned & Basic & 77.70 (1.87) & 54.49 (3.48) & 43.56 (3.07) & 18.71 (4.87)     \\\hline\hline
\end{tabular}}
\begin{minipage}{15.5cm}
\vspace{3pt}
    \footnotesize
    \begin{tabular}{p{1cm} p{14cm}}
         \textbf{Notes}: & \setlength{\baselineskip}{1.2\baselineskip} 
    This table presents the percentage of data saved using AAE with various GPT models, averaged over 50 experimental runs in the first empirical setting. The standard errors are shown in parentheses. One sample t-tests show that the savings are all significant at the 90\% level except for the ones marked with a star.
    \end{tabular}
  \end{minipage}
\end{table}

% \begin{table}[ht]
% \renewcommand{\arraystretch}{1.3}
% \centering
% {\footnotesize
% \caption{Percentage of saving in data size in empirical setting II(\%)}
% \label{tab:saving-2}
% \begin{tabular}{cccccc}\hline\hline
% ~~~~~~~~~~~~~~~~~~~~~~Model~~~~~~~~~~~~~~~~~~~~~~                               & ~~~Prompt~~~ & ~~~$m=50$~~~          & ~~~$m=100$~~~         & ~~~$m=150$~~~         &~~~$m=200$~~~         \\\hline
% \multirow{2}{*}{GPT-3.5-Turbo-0613} & Basic & 60.08 & 38.94 & 35.00 & 18.60 \\
%  & CoT & 55.27 & 36.57 & 28.71 & 15.00 \\\hline
% \multirow{2}{*}{GPT-3.5-Turbo-0125} & Basic & 49.01 & 37.05 & 30.92 & 17.58 \\
%  & CoT & 56.78 & 39.00 & 30.35 & 17.67 \\\hline
% \multirow{2}{*}{GPT-4} & Basic & 49.41 & 35.11 & 27.76 & 13.49 \\
%  & CoT & 61.88 & 39.45 & 36.45 & 17.64 \\\hline
% \multirow{3}{*}{GPT-4o} & Basic & 44.09 & 35.61 & 27.58 & 18.49 \\
%  & CoT & 47.40 & 33.79 & 33.38 & 16.89 \\
%  & FS & 58.68 & 41.57 & 38.58 & 18.40 \\\hline
% GPT-4o Fine-tuned & Basic & 50.67 & 34.81 & 21.82 & 12.71\\\hline\hline
% \end{tabular}
% }
% \end{table}

%% file: src/app_4_theory.tex
\section{Supporting Arguments for Section \ref{sec:consistency-normality}}

\subsection{Proofs of Main Results}

\noindent
\revise{
\textbf{Proof of Theorem \ref{thm:consistency}.} We first prove the first part of the Theorem. 
As the first step, we argue that the optimizer in \eqref{eqn:best-in-class} exists and is uniquely defined, whose proof is presented later. 
This result only requires $(i)$ in Assumption \ref{assumption:regularity-consistency}.

\begin{lem}[Unique Optimizer]
\label{lem:unique-optimizer}
    There exists a unique optimizer in \eqref{eqn:best-in-class}. 
\end{lem}

Given the lemma, we can link the global loss function to the empirical loss function that we use as follows.
By Lemma \ref{lem:unique-optimizer}, $\betavec^*$ is the unique solution to $\max_{\betavec\in \Real^{k}}Q(\betavec)$, where 
\begin{align}
\label{eqn:reformulation}
     Q(\betavec) ~:=~&\E_{\x}\left[
    \sum_{j\in \Kscr^+} \Prob\big(y = j| \x\big)\log\sigma_{j}(\x, \betavec)
    \right]~\stackrel{\text{(a)}}{=}~\E_{\x}\left[
    \sum_{j\in \Kscr^+} \E_{z}\big[\Prob\big(y = j| \x, z\big)\big|\x\big]\log\sigma_{j}(\x, \betavec)
    \right]\nonumber\\
    ~=~&\E_{\x}\hspace{-1mm}\left[
     \E_{z}\left[\sum_{j\in \Kscr^+}\Prob\big(y = j| \x, z\big)\log\sigma_{j}(\x, \betavec)\,\Bigg|\,\x\right]
    \right]~=~\E_{\x, z}\hspace{-1mm}\left[
     \sum_{j\in \Kscr^+}\Prob\big(y = j| \x, z\big)\log\sigma_{j}(\x, \betavec)
    \right]\nonumber\\
    ~\stackrel{\text{(b)}}{=}~&    
    \E_{\x}\left[
    \sum_{j\in \Kscr^+}g_j(\x_i, z_i; {\thetavec}^*)\log\sigma_{j}(\x, \betavec)
    \right],
\end{align}
where (a) follows from an application of the law of total expectation and (b) follows from the definition of $\big(g_j(\x, z; \thetavec^*):j\in \Kscr^+\big)$.
Then, using the representation of $Q(\betavec)$ in 
\eqref{eqn:reformulation}, the next lemma shows that our empirical loss function converges uniformly to $Q(\betavec)$. 
Let $\epsilon$ be an arbitrary positive constant and $\Bscr(\betavec^*, \epsilon)$ be a closed ball around $\betavec^*$ with a radius $\epsilon$.

\begin{lem}[Uniform Convergence]
\label{lem:uniform-convergence}
    $
    \sup_{\betavec \in \Bscr(\betavec^*, \epsilon)}
    \left|
    \widehat{Q}(\hat{\thetavec};\betavec) - {Q}(\betavec)
    \right| \pconverge 0
    $ as $m,n \rightarrow \infty$.
\end{lem}

From now on, let us fix any $\epsilon_1 > \epsilon_2 > 0$. 
Since $\Bscr(\betavec^*, \epsilon_1)$ is compact and ${Q}(\betavec)$ is continuous, as implied by the dominated convergence theorem, there exists $\delta > 0$ such that 
\begin{align}
\label{eqn:outer-circle}
    {Q}(\betavec^*) ~=~ \sup_{\betavec\in\Bscr(\betavec^*, \epsilon_1)\setminus \Bscr(\betavec^*, \epsilon_2)}{Q}(\betavec) + \delta.
\end{align}
Indeed, if $\sup_{\betavec\in\Bscr(\betavec^*, \epsilon_1)\setminus \Bscr(\betavec^*, \epsilon_2)}{Q}(\betavec)$ is equal to the ${Q}(\betavec^*)$, due to the compactness of 
$\Bscr(\betavec^*, \epsilon_1)$, 
we can find a sequence in $\Bscr(\betavec^*, \epsilon_1)\setminus \Bscr(\betavec^*, \epsilon_2)$ that converges to a point $\tilde{\betavec}$ such that $\lVert \tilde{\betavec} - \betavec^*\rVert_2 \geq \epsilon_2$ and ${Q}(\betavec^*) = {Q}(\tilde{\betavec})$.
This contradicts Lemma \ref{lem:unique-optimizer}.
Let us define $\event$ as the event on which $\sup_{\betavec \in \Bscr(\betavec^*, \epsilon_1)}
\left|
\widehat{Q}(\hat{\thetavec};\betavec) - {Q}(\betavec)
\right| \leq \frac{\delta}{3}$. Therefore, by Lemma \ref{lem:uniform-convergence}, $\Prob(\event^c)\rightarrow 0$. 
On $\event$, we have 
\begin{align}
    &\max_{\betavec \in \Bscr(\betavec^*, \epsilon_2)} \widehat{Q}(\hat{\thetavec};\betavec)~\geq~\widehat{Q}(\hat{\thetavec};\betavec^*)~\stackrel{\text{(c)}}{\geq}~{Q}(\betavec^*) -\frac{\delta}{3} \stackrel{\text{(d)}}{=} \sup_{\betavec\in\Bscr(\betavec^*, \epsilon_1)\setminus \Bscr(\betavec^*, \epsilon_2)}{Q}(\betavec)  + \frac{2\delta}{3}    \label{eqn:delta/3}\\ &~~~~~~~~~~~~~~~~~~~~~~~~~~~~~~~~~~~~~~~~~~~~~~~~~~~~~~~~~~~~~~~~~~~~~~~~~~~~~~~~~~~~\stackrel{\text{(e)}}{\geq}~\sup_{\betavec\in\Bscr(\betavec^*, \epsilon_1)\setminus \Bscr(\betavec^*, \epsilon_2)}\widehat{Q}(\hat{\thetavec};\betavec) + \frac{\delta}{3},\nonumber
\end{align}
where (c) and (e) follow from the definition of $\event$ and (d)
follows from \eqref{eqn:outer-circle}.
Let us assume $\betavec' \in \argmax _{\betavec\in \Bscr(\betavec^*, \epsilon_2)}
\widehat{Q}(\hat{\thetavec};\betavec)$.
For any $\betavec \in \Bscr(\betavec^*, \epsilon_1)^{\text{c}}$, there exists $\lambda \in (0,1)$ with $\lambda\betavec' +(1 - \lambda)\betavec \in \Bscr(\betavec^*, \epsilon_1)\setminus \Bscr(\betavec^*, \epsilon_2)$.
Then, by \eqref{eqn:delta/3},
$$
\widehat{Q}(\hat{\thetavec};\betavec')~>~\frac{\delta}{3} + \widehat{Q}\Big(\hat{\thetavec};\lambda\betavec' +(1 - \lambda)\betavec\Big) ~\geq~ \frac{\delta}{3} + \lambda\widehat{Q}(\hat{\thetavec};\betavec') + (1 - \lambda)\widehat{Q}(\hat{\thetavec};\betavec)
$$
due to the concavity of $\widehat{Q}(\hat{\thetavec};\cdot)$.
Thus, $\widehat{Q}(\hat{\thetavec};\betavec') ~>~ \frac{\delta}{3( 1 - \lambda)} + \widehat{Q}(\hat{\thetavec};\betavec)> \frac{\delta}{3} + \widehat{Q}(\hat{\thetavec};\betavec)$. In sum, 
\begin{align}
\label{eqn:beta-prime-delta/3}
    \widehat{Q}(\hat{\thetavec};\betavec') ~\geq~ \sup_{\betavec \in \Bscr(\betavec^*, \epsilon_2)^{\text{c}}} \widehat{Q}(\hat{\thetavec};\betavec) + \frac{\delta}{3}.
\end{align}
Therefore, on $\event$, it must be that $\hat{\betavec}^{\sf AAE} \in \Bscr(\betavec^*, \epsilon_2)$, since $\hat{\betavec}^{\sf AAE} ~\in~\argmax_{\betavec \in \Real^{d}}\widehat{Q}(\hat{\thetavec};\betavec)$. Since $\Prob(\event^c)\rightarrow 0$, the conclusion follows.\footnote{\revise{In case $\hat{\betavec}^{\sf AAE}$ is chosen such that $
\widehat{Q}(\hat{\thetavec};\hat{\betavec}^{\sf AAE}) = \sup_{\betavec \in \Real^{d}} \widehat{Q}(\hat{\thetavec};\betavec) - o_{\text{P}}(1)
$, we notice that \eqref{eqn:beta-prime-delta/3} implies that 
$$
\Prob\big(\hat{\betavec}^{\sf AAE} \notin \Bscr(\betavec^*, \epsilon_2)\big) \leq \Prob\left(\widehat{Q}(\hat{\thetavec};\hat{\betavec}^{\sf AAE}) \leq \sup_{\betavec \in \Real^{d}} \widehat{Q}(\hat{\thetavec};\betavec) - \delta/3\right) + 
\Prob(\event^c)\rightarrow 0
,$$
so the consistency still follows.} 
} 

For the second part of the theorem, we argue as follows. 
By the Taylor expansion and the first order condition, which holds with probability approaching one since $\hat{\betavec}^{\sf AAE} \stackrel{\text{P}}{\rightarrow}{\betavec}^*$, we have 
\begin{align}
\label{eqn:taylor-1}
    \Ovec ~=&~ \frac{1}{n}\sum_{i=1}^n\sum_{j\in \Kscr^+}g_j(\x_i, z_i; \hat{\thetavec})\nabla_{\betavec}\log \sigma_j(\x_i; \hat{\betavec}^{\sf AAE})\\
    &~~~~~~~~~~~~~~~~~~~~~~~=~\frac{1}{n}\sum_{i=1}^n\sum_{j\in \Kscr^+}g_j(\x_i, z_i; \hat{\thetavec})\Big(\nabla_{\betavec}\log \sigma_j(\x_i; {\betavec}^*) + \nabla^2_{\betavec}\log \sigma_j(\x_i; \tilde{\betavec})(\hat{\betavec}^{\sf AAE} - \betavec^*)\Big),\nonumber
\end{align}
where $\tilde{\betavec} \in [\betavec^*, \hat{\betavec}^{\sf AAE}]$.
Because by Theorem \ref{thm:consistency}, $\hat{\betavec}^{\sf AAE} \pconverge \betavec^*$, 
$\tilde{\betavec} \pconverge \betavec^*$.
Similarly, we have the following Taylor expansion, 
where $\tilde{\thetavec} \in [{\thetavec}^*, \hat{\thetavec}]$ and  $\tilde{\thetavec}\pconverge \thetavec^*$: 
\begin{align}
\label{eqn:Taylor-2}
    & \sum_{i=1}^n\sum_{j\in \Kscr^+}g_j(\x_i, z_i; \hat{\thetavec})\nabla_{\betavec}\log \sigma_j(\x_i; {\betavec}^*) ~=~\\ 
     & ~~~~~~~~~~~~~~\sum_{i=1}^n\sum_{j\in \Kscr^+}g_j(\x_i, z_i; {\thetavec}^*)\nabla_{\betavec}\log \sigma_j(\x_i; {\betavec}^*) + \sum_{i=1}^n\sum_{j\in \Kscr^+}\nabla_{\betavec}\log \sigma_j(\x_i; {\betavec}^*) \nabla_{\thetavec}g_j(\x_i, z_i; \tilde{\thetavec})^{\top}
      (\hat{\thetavec} - \thetavec^*).\nonumber
\end{align}
We next present an intermediate result regarding the derivatives.

\begin{lem}[Convergence to $\Omegavec$ and $\Gammavec$] 
\label{lem:Omega}
It holds that 
\begin{align*}
        \frac{1}{n} \sum_{i=1}^n \sum_{j\in \Kscr^+} g_j(\x_i, z_i; \hat{\thetavec})\nabla^2_{\betavec}\log \sigma_j(\x_i; \tilde{\betavec}) \,\pconverge\, \Omegavec ~~\text{and}~~\frac{1}{{n}}\sum_{i=1}^n\sum_{j\in \Kscr^+}
    \nabla_{\betavec}\log \sigma_j(\x_i; {\betavec}^*) \nabla_{\thetavec}g_j(\x_i, z_i; \tilde{\thetavec})^{\top} \,\pconverge\, \Gammavec.
\end{align*}
\end{lem}
Therefore, \eqref{eqn:taylor-1} implies that 
\begin{align*}
    ~\sqrt{n}(\hat{\betavec}^{\sf AAE} - \betavec^*)
    ~=&~\Omegavec^{-1}\times \frac{1}{\sqrt{n}}\sum_{i=1}^n\sum_{j\in \Kscr^+}g_j(\x_i, z_i; \hat{\thetavec})\nabla_{\betavec}\log \sigma_j(\x_i; {\betavec}^*) + o_{\text{\upshape{P}}}(1)\\
    ~\stackrel{\text{(a)}}{=}&~\Omegavec^{-1}\times \Bigg[\frac{1}{\sqrt{n}}\sum_{i=1}^n\sum_{j\in \Kscr^+}
    g_j(\x_i, z_i; {\thetavec}^*)\nabla_{\betavec}\log \sigma_j(\x_i; {\betavec}^*)\\
    &+ \sqrt{\frac{n}{m}}\times
    \left(\frac{1}{{n}}\sum_{i=1}^n\sum_{j\in \Kscr^+}
    \nabla_{\betavec}\log \sigma_j(\x_i; {\betavec}^*) \nabla_{\thetavec}g_j(\x_i, z_i; \tilde{\thetavec})^{\top}\right)\times
      \sqrt{m}(\hat{\thetavec} - \thetavec^*)\Bigg]
     + o_{\text{\upshape{P}}}(1),
\end{align*}
where (a) follows from \eqref{eqn:Taylor-2}.
Using the second display of Lemma \ref{lem:Omega} and verifying 
$$
\sum_{j\in \Kscr^+}
    g_j(\x, z; {\thetavec}^*)\nabla_{\betavec}\log \sigma_j(\x; {\betavec}^*) ~=~\sum_{j\in \Kscr}\big(g_j(\x, z; {\thetavec}^*) - \sigma_{j}(\x; \betavec^*)\big)\x_{(j)}~~\forall\, \x \in \Xscr, \forall\, z \in \Kscr,
$$
which follows from a straightforward calculation as remarked in Lemma \ref{lem:derivative},
we have 
$$\sqrt{n}(\hat{\betavec}^{\sf AAE} - \betavec^*)~=~ \Omegavec^{-1}\times \left(\frac{1}{n}\sum_{i=1}^n  \sum_{j\in \Kscr}\big(g_j(\x_i, z; {\thetavec}^*) - \sigma_{j}(\x_i; \betavec^*)\big)\x_{(j)} + \sqrt{\frac{n}{m}} \Gammavec\times \sqrt{m}(\hat{\thetavec} - \thetavec^*)\right) + o_{\text{\upshape{P}}}(1).
$$
The weak convergence results from the Lindeberg–L\'evy central limit theorem.
\hfill $\blacksquare$}
\vspace{3mm}

\revise{\noindent
\textbf{Proof of Proposition  \ref{prop:weak-dom}.}
The proof begins with a reformulation of $\Gammavec$ as follows 
\begin{align*}
    \Gammavec ~&=~ \E_{\x,z}\left[\sum_{j \in \Kscr^+} \left(\ind_{\{j\neq 0\}}\x_{(j)} - \sum_{j' \in \Kscr}\sigma_{j'}(\x, \betavec^*)\x_{(j')}\right)\nabla_{\thetavec} g_j(\x, z, \thetavec^*)^{\top} \right]\\
    &~~~~~~~~~~~~~~~~~~=~
    \E_{\x,z}\left[\sum_{j \in \Kscr^+}g_j(\x, z, \thetavec^*)  \left(\ind_{\{j\neq 0\}}\x_{(j)} - \sum_{j' \in \Kscr}\sigma_{j'}(\x, \betavec^*)\x_{(j')}\right)\frac{1}{g_j(\x, z, \thetavec^*) }\nabla_{\thetavec} g_j(\x, z, \thetavec^*)^{\top} \right]\\
    &~~~~~~~~~~~~~~~~~~\stackrel{\text{(a)}}{=}~
    \E_{\x,z}\left[\sum_{j \in \Kscr^+} g_j(\x, z, \thetavec^*) \left(\ind_{\{j\neq 0\}}\x_{(j)} - \sum_{j' \in \Kscr}\sigma_{j'}(\x, \betavec^*)\x_{(j')}\right)\nabla_{\thetavec} \log g_j(\x, z, \thetavec^*)^{\top} \right]\\
    ~&~~~~~~~~~~~~~~~~~~\stackrel{\text{(b)}}{=}~
    \E_{\x,y,z}\left[\bigg(\sum_{j\in \Kscr}\big(\ind_{\{y = j\}} - \sigma_{j}(\x; \betavec^*)\big)\x_{(j)}\bigg)\nabla_{\thetavec} \log g_y(\x, z, \thetavec^*)^{\top} \right],
\end{align*}
where (a) follows from the definition of $\nabla_{\thetavec} \log g_j(\x, z, \thetavec^*)^{\top}$ and (b) follows because
\begin{align*}
    &\sum_{j \in \Kscr^+} g_j(\x, z, \thetavec^*) \left(\ind_{\{j\neq 0\}}\x_{(j)} - \sum_{j' \in \Kscr}\sigma_{j'}(\x, \betavec^*)\x_{(j')}\right)\nabla_{\thetavec} \log g_j(\x, z, \thetavec^*)^{\top}\\
    &~~~~~~~~~~~~~~~~~~~~~~~~~~~~~~~~~~~~~~~~~~~~~~~~~=~
\E_{y}
\left[
\bigg(\sum_{j\in \Kscr}\big(\ind_{\{y = j\}} - \sigma_{j}(\x; \betavec^*)\big)\x_{(j)}\bigg)\nabla_{\thetavec} \log g_y(\x, z, \thetavec^*)^{\top}~\bigg|~\x, z
\right].
\end{align*}
Therefore, if we define 
$$
\uvec ~=~ \nabla_{\thetavec} \log g_y(\x, z, \thetavec^*)^{\top}~\text{ and }~\wvec~=~\sum_{j\in \Kscr}\big(\ind_{\{y = j\}} - \sigma_{j}(\x; \betavec^*)\big)\x_{(j)},
$$
it follows that the projection of $\wvec$ on $\uvec$ is $\E\left[\wvec \uvec^{\top} \right]\E\left[\uvec \uvec^{\top} \right]^{-1}\uvec$ and the covariance of the residual of the projection, i.e., $\wvec - \E\left[\wvec \uvec^{\top} \right]\E\left[\uvec \uvec^{\top} \right]^{-1}\uvec$,
is 
\begin{align*}
&
\E\left[
\left(\wvec - \E\left[\wvec \uvec^{\top} \right]\E\left[\uvec \uvec^{\top} \right]^{-1}\uvec\right)\left(\wvec - \E\left[\wvec \uvec^{\top} \right]\E\left[\uvec \uvec^{\top} \right]^{-1}\uvec\right)^{\top}
\right]\\
&~~~~~~~~~~~~~~~~~~~~~~~~~~~~~~~~~~~~~~~~~~~~~~~~~~~~~~~~~~~\stackrel{\text{(c)}}{=}~\E\left[\wvec \wvec^{\top} \right] - \E\left[\wvec \uvec^{\top} \right]\E\left[\uvec \uvec^{\top} \right]^{-1}\E\left[\uvec \wvec^{\top} \right]~\stackrel{\text{(d)}}{=}~\check{\Jvec} - \Gammavec\Lambdavec\Gammavec^{\top},
\end{align*}
where (c) follows by algebra and (d) follows by the definitions of $\check{\Jvec}$, $\Gammavec$ and $\Lambdavec$.
Therefore, $\check{\Jvec} \succeq \Gammavec\Lambdavec\Gammavec^{\top}$ and the first claim follows.
As indicated in the main text, the rest of the results follow from sending $n$ to infinity. 
\hfill $\blacksquare$
\vspace{3mm}}

\revise{
\noindent
\textbf{Proof of Theorem  \ref{thm:aae-bound}.}
Let us assume that $-\widehat{Q}(\hat{\thetavec};\betavec)$ is strongly convex with parameter $\hat{\eta}$. 
    Then, we derive a basic inequality
    $$
    -\widehat{Q}(\hat{\thetavec};\betavec^*) - \nabla\widehat{Q}(\hat{\thetavec};\betavec^*)^{\top}\left(\hat{\betavec}^{\sf AAE} - {\betavec}^* \right) + \frac{\hat{\eta}}{2}\big\lVert \hat{\betavec}^{\sf AAE} - {\betavec}^* \big\rVert^2_2~\leq~-\widehat{Q}(\hat{\thetavec};\hat{\betavec}^{\sf AAE})~\leq~-\widehat{Q}(\hat{\thetavec};\hat{\betavec}^*),
    $$
    which implies 
    \begin{align}
    \label{eqn:basic-consequence}
        \big\lVert \hat{\betavec}^{\sf AAE} - {\betavec}^* \big\rVert_2 ~\leq~ \frac{2}{\hat{\eta}}\big\lVert 
        \nabla\widehat{Q}(\hat{\thetavec};\betavec^*)
        \big\rVert_2.  
    \end{align}
   We note that 
   \begin{align*}
          &\nabla\widehat{Q}(\hat{\thetavec};\betavec^*)~=~ \frac{1}{n}\sum_{i=1}^n\sum_{j \in \Kscr}\left(g_j(\x_i, z_i; \hat{\thetavec}) - \sigma_j(\x_i; \betavec^*)\right)\x_{i(j)}\\
          ~&~~~~~~~=~ \frac{1}{n}\sum_{i=1}^n\sum_{j \in \Kscr}\left(g_j(\x_i, z_i; \hat{\thetavec}) - g_j(\x_i, z_i; {\thetavec}^*)\right)\x_{i(j)} + \frac{1}{n}\sum_{i=1}^n\sum_{j \in \Kscr}\left(g_j(\x_i, z_i; {\thetavec}^*) - \sigma_j(\x; \betavec^*)\right)\x_{i(j)}.
   \end{align*}
   By assumption, for any $\ell = 1,\dots, d$, $\E_{z}[g_j(\x, z; {\thetavec}^*)x_{(j)\ell}|\x] = \sigma_j(\x; \betavec^*)x_{(j)\ell}$ and 
    the sub-Gaussian norm of $\frac{1}{n}\sum_{i=1}^n\sum_{j \in \Kscr}\left(g_j(\x_i, z_i; {\thetavec}^*) - \sigma_j(\x_i; \betavec^*)\right)\x_{i(j)}$ satisfies 
    \begin{align*}
        & \left \lVert \frac{1}{n}\sum_{i=1}^n\sum_{j \in \Kscr}\left(g_j(\x_i, z_i; {\thetavec}^*) - \sigma_j(\x_i; \betavec^*)\right)\x_{i(j)}\right\rVert_{\psi_2}\leq \sum_{j \in \Kscr}\left \lVert \frac{1}{n}\sum_{i=1}^n\left(g_j(\x_i, z_i; {\thetavec}^*) - \sigma_j(\x_i; \betavec^*)\right)\x_{i(j)}\right\rVert_{\psi_2}\\
        \stackrel{\text{(a)}}{\leq} & \frac{1}{\sqrt{n}}\sum_{j \in \Kscr}\left \lVert \left(g_j(\x, z; {\thetavec}^*) - \sigma_j(\x; \betavec^*))\right)\x_{(j)}\right\rVert_{\psi_2} \leq \frac{2}{\sqrt{n}}\sum_{j \in \Kscr}\left \lVert\x_{(j)}\right\rVert_{\psi_2} = \frac{2}{\sqrt{n}}\sum_{j \in \Kscr} \sigma_{\x_{(j)}},
    \end{align*}
    where (a) follows because the samples are i.i.d.
    Let $N({1/2}, \ell_2)$ be a $1/2$-cover of the unit $\ell_2$-norm ball in $\Real^d$ so $|N({1/2}, \ell_2)| \leq 5^d$. 
    For any $\mathbf{b}$ in the  unit $\ell_2$-norm ball, we assume that it is covered by $T(\bvec) \in N({1/2}, \ell_2)$. 
    For any $\mathbf{a} \in \Real^d$, we note that 
    $$\lVert \mathbf{a} \rVert_2 = \sup_{\mathbf{b} \in \Real^d \lVert \mathbf{b} \rVert_2 \leq 1}\avec^{\top}\bvec \leq \sup_{\mathbf{b} \in \Real^d \lVert \mathbf{b} \rVert_2 \leq 1}\avec^{\top}(\bvec - T(\bvec)) + \max_{\tilde{\bvec} \in N_{1/2}} \avec^{\top}\tilde{\bvec}\leq \frac{1}{2}\lVert \mathbf{a} \rVert_2 + \max_{\tilde{\bvec} \in N({1/2}, \ell_2)} \avec^{\top}\tilde{\bvec}$$ 
    so $\lVert \mathbf{a} \rVert_2  \leq 2\max_{\tilde{\bvec} \in N({1/2}, \ell_2)} \avec^{\top}\tilde{\bvec}$. 
    Then by Lemma 2.2.2 in \cite{vaart2023empirical}, it holds that 
    \begin{align*}
         & \left \lVert \left \lVert \frac{1}{n}\sum_{i=1}^n\sum_{j \in \Kscr}\left(g_j(\x_i, z_i; {\thetavec}^*) - \sigma_j(\x_i; \betavec^*)\right)\x_{i(j)}\right\rVert_2 \right\rVert_{\psi_2}\\
        & ~~~~~~\preceq   \sqrt{d\log(5)}\max_{\tilde{\bvec} \in N_{1/2}} 
         \left\lVert
        \left(\frac{1}{n}\sum_{i=1}^n\sum_{j \in \Kscr}\left(g_j(\x_i, z_i; {\thetavec}^*) - \sigma_j(\x_i; \betavec^*)\right)\x_{i(j)}\right)\tilde{\bvec}
         \right \rVert_{\psi_2} \preceq \sqrt{\frac{d\log(5)}{n}}\sum_{j \in \Kscr}\sigma_{\x_{(j)}}.
    \end{align*}
    Thus, using Markov inequality, we observe that for some absolute constant $c_1$
    \begin{align}
    \label{eqn:hoeffding-1}
    \Prob \left(
     \left \lVert \frac{1}{n}\sum_{i=1}^n\sum_{j \in \Kscr}\left(g_j(\x_i, z_i; {\thetavec}^*) - \sigma_j(\x_i; \betavec^*)\right)\x_{i(j)}\right\rVert_2 \geq t_1 \right) \leq 2e^{-\frac{c_1n t_1^2}{d\left(\sum_{j \in \Kscr}\sigma_{\x_{(j)}}\right)^2}}
    \end{align}
    Also, for any random variable $w$, we let $\lVert w\rVert_{L_2(\Prob)}:=(\E[w^2])^{\frac{1}{2}}$ and obtain 
    \begin{align*}
        & \E_{\x, z}\left[\left\lVert
        \frac{1}{n}\sum_{i=1}^n\sum_{j \in \Kscr}\left(g_j(\x_i, z_i; {\thetavec}^*) - g_j(\x_i, z_i; \hat{\thetavec})\right)\x_{i(j)}\right \rVert_2
        \right]\leq
        \E_{\x, z}\left[\left\lVert
        \sum_{j \in \Kscr}\left(g_j(\x, z; {\thetavec}^*) - g_j(\x, z; \hat{\thetavec})\right)\x_{(j)}\right \rVert_2
        \right]\\
        \leq & \E_{\x, z}\left[\sum_{j \in \Kscr}
        \left|g_j(\x, z; {\thetavec}^*) - g_j(\x_i, z_i; \hat{\thetavec})\right| 
        \left\lVert
        \x_{(j)}\right \rVert_2
        \right]\leq
        \E_{\x, z}\left[\max_{j \in \Kscr}\left\lVert
        \x_{(j)}\right \rVert_2\cdot \sum_{j \in \Kscr}
        \left|g_j(\x, z; {\thetavec}^*) - g_j(\x_i, z_i; \hat{\thetavec})\right| 
        \right]\\
        \leq &\left\lVert\max_{j \in \Kscr}\left\lVert
        \x_{(j)}\right \rVert_2
        \right\rVert_{L_2(\Prob)}
        \cdot 
        \left\lVert\sum_{j \in \Kscr}
        \left|g_j(\x, z; {\thetavec}^*) - g_j(\x_i, z_i; \hat{\thetavec})\right|\right\rVert_{L_2(\Prob)}\\
        \stackrel{\text{(b)}}{\leq} &\frac{1}{\sqrt{2}}\left\lVert\max_{j \in \Kscr}\left\lVert
        \x_{(j)}\right \rVert_2
        \right\rVert_{L_2(\Prob)}\cdot\left( \E_{\x, z}\left[{\sf KL}\big( \mathbf{g}(\x, z, \thetavec^*)\,\big\vert\,\mathbf{g}(\x, z, \hat{\thetavec})\big) \right]\right)^{1/2} \leq \sqrt{\frac{\kappa\epsilon_1\big(m,\Tilde{d},\delta\big)}{2}},
    \end{align*}
    where (b) follows from the Pinsker's inequality.
    Next, using an argument similar to \eqref   {eqn:hoeffding-1}, we have 
    \begin{align}
    \label{eqn:hoeffding-2}
        \Prob \left(
    \left\lVert
        \frac{1}{n}\sum_{i=1}^n\sum_{j \in \Kscr}\left(g_j(\x_i, z_i; {\thetavec}^*) - g_j(\x_i, z_i; \hat{\thetavec})\right)\x_{i(j)}\right \rVert_2 \geq \sqrt{\frac{\kappa\epsilon_1\big(m,\Tilde{d},\delta\big)}{2}} + t_1 \right) \leq 2e^{-\frac{c_1n t_1^2}{d\left(\sum_{j \in \Kscr}\sigma_{\x_{(j)}}\right)^2}}
    \end{align}
    Thus, combining \eqref{eqn:hoeffding-1} and \eqref{eqn:hoeffding-2}, we have 
    \begin{align}
    \label{eqn:derivative-infity-norm}
       \Prob \left(\left\lVert\nabla\widehat{Q}(\hat{\thetavec};\betavec^*) \right \rVert_{2} \geq \frac{\kappa}{\sqrt{2}}\sqrt{\epsilon_1\big(m,\Tilde{d},\delta\big)} +2t_1 \right) \leq 4e^{-\frac{c_1n t_1^2}{2d\left(\sum_{j \in \Kscr}\sigma_{\x_{(j)}}\right)^2}}.
    \end{align}

    To characterize $\hat{\eta}$, we note that $$-\nabla^2\widehat{Q}(\hat{\thetavec};\betavec) = \frac{1}{n}\sum_{i=1}^n \sum_{j \in \Kscr}\sigma_j(\x; \betavec)\left(
    \x_{i(j)}\x^{\top}_{i(j)} - \sum_{j' \in \Kscr}\sigma_{j'}(\x; \betavec)\x_{i(j)}\x^{\top}_{i(j')} 
    \right),$$ so for any $\qvec \in \Real^{d}$, 
    \begin{align*}
        &\qvec^{\top}\left(-\nabla^2\widehat{Q}(\hat{\thetavec};\betavec)\right)\qvec~=~ \frac{1}{n}\sum_{i=1}^n \sum_{j \in \Kscr}\sigma_j(\x; \betavec)\left(
    \left(\qvec^\top \x_{i(j)}\right)^2 - \sum_{j' \in \Kscr}\sigma_{j'}(\x; \betavec)\left(\qvec_{(j)}^\top \x_{i(j)}\right)\left(\qvec^\top \x_{i(j')}\right)
    \right)\\
    \geq~ &\frac{1}{n}\sum_{i=1}^n \sum_{j \in \Kscr}\sigma_0(\x_i; \betavec)\min_{j\in \Kscr}\sigma_j(\x_i; \betavec) \cdot\qvec^{\top}\left(\sum_{j \in \Kscr}\x_{i(j)}\x^{\top}_{i(j)}\right)\qvec ~\geq~ \frac{1}{e^R(1 + ke^R)^2}\cdot\qvec^{\top}\widehat{\Sigmavec}\qvec,
    \end{align*}
    where we write $\widehat{\Sigmavec}:= (1/n)\sum_{i=1}^n\sum_{j \in \Kscr}\x_{i(j)}\x^{\top}_{i(j)}$.
    Let us also  write $\Sigmavec_{(j)} := \E\big[\sum_{j\in \Kscr}\x_{(j)}\x_{(j)}^{\top}\big]$ and its empirical counterpart as $\widehat{\Sigmavec}_{(j)}$. 
    Adapting the arguments in Theorem 5.39 of \cite{vershynin2010introduction}, we obtain that $\Prob\left(\left\lVert \widehat{\Sigmavec}_{(j)} - {\Sigmavec}_{(j)}\right\rVert_2 \leq t_3\right) \geq 1 - e^{-c_3 t_2^2}$ for all $t_3 \in (0,1)$ and $t_2 > 1$ as long as $n \geq c_2dt_2t_3$, where $c_1$ and $c_2$ are constants that depends on $\max_{j \in \Kscr}\sigma_{\x_{(j)}}$. 
    Let us set $t_3 = \lambda_{\min}(\Sigmavec)/2K \wedge 1$. 
    Then as long as $n \geq
\frac{dt_2}{\lambda_{\min}(\Sigmavec)/2K \wedge 1}$, with an application of union bound and triangular inequality, we have that with probability at least $1 - ke^{-c_3t^2_2}$,  
$\left\lVert \widehat{\Sigmavec}- {\Sigmavec}\right\rVert_2 \leq \lambda_{\min}(\Sigmavec)/2$ so $\lambda_{\min}(\widehat{\Sigmavec}) \geq \lambda_{\min}(\Sigmavec)/2$.
Together with \eqref{eqn:derivative-infity-norm} and \eqref{eqn:basic-consequence}, 
it holds that 
$$
\Prob
\left(
\big\lVert \hat{\betavec}^{\sf AAE} - {\betavec}^* \big\rVert_2 ~\leq~ \frac{
\left(2\sqrt{2\kappa\epsilon_1\big(m,\Tilde{d},\delta\big)} + 8t_1  
       \right)e^R(1 + ke^R)^2
}{\lambda_{\min}(\Sigmavec)}
\right)
\geq 1 - 4e^{-\frac{c_1n t_1^2}{2d\left(\sum_{j \in \Kscr}\sigma_{\x_{(j)}}\right)^2}} - ke^{-c_3t^2_2} - \delta.
$$
We set $t_1 = \sqrt{\epsilon_1\big(m,\Tilde{d}, ,\delta\big)}$ and $t_2 = (1/c_2)\log\left(k/\delta\right)\vee 1$, we observe that as long as 
$$n ~\geq~ 
\max\left\{
\frac{d
\left(
\frac{1}{c_2}\log\left(\frac{k}{\delta}\right) \vee 1\right)
}{\frac{\lambda_{\min}(\Sigmavec)}{2K}\wedge 1},~ \frac{2d\left(\sum_{j \in \Kscr}\sigma_{\x_{(j)}}\right)^2}{c_1\epsilon_1\big(m,\Tilde{d},\delta\big)}\log\left(\frac{4}{\delta}\right)
    \right\},
$$
the conclusion holds.
\hfill $\blacksquare$}
\vspace{3mm}

\revise{
\noindent
\textbf{Proof of Proposition  \ref{prop:kl}.}
The proof relies on a localization argument. 
    On the notation side, for simplicity we follow that of empirical process literature. 
    To begin with, we define the function class
    $$
    \mathscr{F}~:=~ \left\{
    \fvec \in \Real^k\,:\,f_j(\tilde{\x}_{(j)}) = \tilde{\x}_{(j)}^{\top}\thetavec~\forall j \in \Kscr ~\text{for some}~ \thetavec \in \Real^d ~\text{with}~ \lVert \thetavec \rVert_1 \leq \tilde{R}
    \right\}.
    $$
    In particular, $\fvec^*(\tilde{\x}) = (\tilde{\x}_{(1)}^{\top}\thetavec^*, \dots, \tilde{\x}_{(k)}^{\top}\thetavec^*)$ and we write $\Fscr^* = \Fscr - \fvec^*$. 
    It is easy to check that $\Fscr^*$ is star-shaped, i.e., if $\hvec \in \Fscr^*$, $\alpha\hvec \in \Fscr^*$ for all $\alpha \in [0,1]$, since the constraint $\lVert \thetavec \rVert_1 \leq \tilde{R}$ defines a convex set. 
    We define 
    $$
    \sigma_j(\fvec(\tilde{\x})) := \frac{e^{f_j(\tilde{\x})}}{1 + \sum_{j'\in \Kscr}e^{f_{j'(\tilde{\x})}}}~ \forall\, j\in \Kscr,~\text{ and }~
    \Lcal(\fvec(\tilde{\x}), y) := -\Indicator\{y \neq 0\}f_y(\tilde{\x}) + \log\left(1 + \sum_{j\in \Kscr}e^{f_{j}(\tilde{\x})}\right),
    $$
    and sometimes abuse notation to write them as $\sigma_j(\fvec)$ and $\Lcal(\fvec, y)$ for a given $({\tilde{\x}}, y)$ which is clear from the context.
    Also, it is convenient to
    use $\Prob_m (\cdot)$ and $\Prob(\cdot)$ to the denote the empirical and population expectations of functions. We 
    further  define the norm, $\lVert \hvec\rVert_2(\tilde{\x}) ~=~ \left(\sum_{j \in \Kscr} h_j(\tilde{\x})^2\right)^{1/2}$, which is a function of $\tilde{
    \x}$ and sometimes write it as $\lVert \hvec\rVert_2$ when $\tilde{\x}$ is fixed and clear from the context. 
    Also, 
    $$\lVert \hvec\rVert_m~=~ \big(\Prob_m\lVert \hvec\rVert_2^2\big)^{1/2},~\text{ and }~\lVert \hvec\rVert_{L_2(\Prob)}~=~ \big(\Prob\lVert \hvec\rVert_2^2\big)^{1/2},$$
    for any function $\hvec$ in $\Fscr$ or $\Fscr^*$.
    We note that $\Lcal(\cdot, y)$ is lipschitz in the first argument in the next.} 

    \revise{    \begin{lem}[Lipschitz Continuity of $\Lcal$]\label{lem:lipschitz}
    For any $\fvec, \fvec' \in \Fscr$, $\tilde{\x}$ and $y$,
    it holds that 
        $$|\Lcal(\fvec, y) - \Lcal(\fvec', y)| \,\leq\, 2\lVert \fvec - \fvec'  \rVert_2 ~\text{ and }~ |\Lcal(\fvec, y) - \Lcal(\fvec', y)|  ~\leq~ 4\tilde{R}\sqrt{k}.$$
    \end{lem}}

    \revise{
    With the preparation so far, in the next, we focus on a
    finite sample bound of the quantity
    $Z_m(r):=\sup_{\fvec\in \Fscr:\, \lVert \fvec - \fvec^*\rVert_{L_2(\Prob)}\leq r}\big|
    \Prob_m(\Lcal_{\fvec} - \Lcal_{\fvec^*} ) - \Prob(\Lcal_{\fvec} - \Lcal_{\fvec^*} ) 
    \big|$,
    where for clarity we re-iterate $$
    \Prob_m(\Lcal_{\fvec} - \Lcal_{\fvec^*} ) = \frac{1}{m}\sum_{i=1}^m \Lcal(\fvec(\tilde{\x}_i), y_i) - \Lcal(\fvec^*(\tilde{\x}_i), y_i)~\text{ and }~
    \Prob(\Lcal_{\fvec} - \Lcal_{\fvec^*} ) = \E_{\tilde{\x}, z}\left[
    \Lcal(\fvec(\tilde{\x}), y) - \Lcal(\fvec^*(\tilde{\x}), y)
    \right]. 
    $$
    Particularly, we focus on $r \geq \delta_m$ where $\delta_m$ is a specific solution that we will define later to the equation 
    $\overline{\sf Rad}(\delta, \Fscr^*)\leq 
    \frac{\delta^2}{\tilde{R}\sqrt{k}}$. 
    Here we define the localized Rademacher complexity 
    $$
    \overline{\sf Rad}(\delta, \Fscr^*) ~:=~ \E_{\tilde{\x},\epsvec}\left[
    \sup_{\hvec \in \Fscr^*: \lVert \hvec\rVert_{L_2(\Prob)}\leq \delta}\left|
    \frac{1}{m}\sum_{i=1}^m\sum_{j \in \Kscr}\epsilon_{ij}h_j(\tilde{\x}_{i,(j)})
    \right|
    \right], 
    $$
    where $\epsilon_{ij}$ are i.i.d. Rademacher random variables that take values in $\{-1, +1\}$, each with probability $1/2$.
    Note that this generalizes the classical single-dimensional localized Rademacher complexity to a multi-dimensional setting. 
    We remark on its key properties.}
    \revise{
    \begin{lem}[Localized Rademacher Complexity]
    \label{lem:local-rade}
        It holds that 
        $\overline{\sf Rad}(\delta, \Fscr^*)$ is non-decreasing and 
        $\overline{\sf Rad}(\delta, \Fscr^*)/\delta$ is non-increasing in $\delta$. Define $\delta_m^* := \inf\left\{\delta \geq 0:\, \overline{\sf Rad}(\delta, \Fscr^*)\leq  \frac{\delta^2}{\tilde{R}\sqrt{k}}\right\}$.
        Then $\delta_m^* > 0$ is finite and $\overline{\sf Rad}(\delta_m^*, \Fscr^*) = \frac{(\delta^*_m)^2}{\tilde{R}\sqrt{k}}$. 
        For any $\delta \geq \delta_m^*$, $\overline{\sf Rad}(\delta, \Fscr^*)\leq  \frac{\delta^2}{\tilde{R}\sqrt{k}}$.
    \end{lem}}

    \revise{
    Let us assume that $r \geq \delta_m \geq \delta^*_m$. 
    Then, defining $\epsilon_i$'s as i.i.d. Rademacher variables, it holds that 
    \begin{align}
    \label{eqn:Zm(r)}
        \E_{\x}\left[ Z_m(r) \right]\stackrel{\text{(a)}}{\leq} 2\E\left[\sup_{\substack{
        \fvec \in \Fscr:\\ \lVert \fvec - \fvec^*\rVert_{L_2(\Prob)} \leq r}} 
        \frac{1}{m}\left|\sum_{i=1}^m \epsilon_i \Big(\Lcal(\fvec(\tilde{\x}_i), y_i) - \Lcal(\fvec^*(\tilde{\x}_i), y_i)
        \Big) \right|
        \right]
        \stackrel{\text{(b)}}{\leq}4\overline{\sf Rad}(r, \Fscr^*)
        \stackrel{\text{(c)}}{\leq}\frac{4r\delta_m}{\tilde{R}\sqrt{k}} \leq \frac{4r^2}{\tilde{R}\sqrt{k}},
    \end{align}
    where (a) follows from a classical symmetrization argument \cite[e.g., see Lemma 2.3.1 in][]{vaart2023empirical}, (b) follows from follows from Lemma \ref{lem:lipschitz} and a multi-variate contraction inequality for Rademacher complexity, \cite[e.g., see][]{maurer2016vector} and (c) follows because $\frac{\overline{\sf Rad}(r, \Fscr^*)}{r^2} \leq \frac{\overline{\sf Rad}(\delta_m, \Fscr^*)}{\delta_m^2} \leq \delta_m.$

    If $\fvec \in \Fscr$
    and $\lVert \fvec - \fvec^*\rVert_{L_2(\Prob)}\leq r$,
    $
    {\sf Var}\left(\Lcal_{\fvec} - \Lcal_{\fvec^*}\right) ~\leq~ \Prob\left[(\Lcal_{\fvec} - \Lcal_{\fvec^*})^2\right]~\leq~ 4 \lVert \fvec - \fvec^*\rVert^2_{L_2(\Prob)}~\leq~ 4r^2$ due to Lemma \ref{lem:lipschitz}.
    Then, 
    \begin{align*}
        &\E_{\x, \y}\left[\sup_{\substack{\fvec \in \Fscr: \\\lVert \fvec - \fvec^*\rVert_{L_2(\Prob)} \leq r}} \Prob_m\left[\left(\Lcal_{\fvec} - \Lcal_{\fvec^*} - \Prob(\Lcal_{\fvec} - \Lcal_{\fvec^*} ) 
        \right)^2\right]
        \right]\\~\leq~ & \E_{\x, \y}\left[\sup_{\substack{\fvec \in \Fscr: \\\lVert \fvec - \fvec^*\rVert_{L_2(\Prob)} \leq r}}\Prob_m\left[\left(\Lcal_{\fvec} - \Lcal_{\fvec^*} - \Prob(\Lcal_{\fvec} - \Lcal_{\fvec^*} ) 
        \right)^2 - {\sf Var}(\Lcal_{\fvec} - \Lcal_{\fvec^*})\right]
        \right] + \sup_{\substack{\fvec \in \Fscr: \\\lVert \fvec - \fvec^*\rVert_{L_2(\Prob)} \leq r}}{\sf Var}(\Lcal_{\fvec} - \Lcal_{\fvec^*})\\
        ~\leq~ & 2\E_{\x, \y,\epsvec}\left[\sup_{\substack{\fvec \in \Fscr: \\\lVert \fvec - \fvec^*\rVert_{L_2(\Prob)} \leq r}}
        \frac{1}{m}\left|\sum_{i=1}^m \epsilon_i \Big(\Lcal(\fvec(\tilde{\x}_i), y_i) - \Lcal(\fvec^*(\tilde{\x}_i), y_i) - \Prob(\Lcal_{\fvec} - \Lcal_{\fvec^*} ) 
        \Big)^2 \right|
        \right] + 4r^2\\
        ~\stackrel{\text{(d)}}{\leq}~ & 32\tilde{R}\sqrt{k}\E_{\x, \y,\epsvec}\left[\sup_{\substack{\fvec \in \Fscr: \\\lVert \fvec - \fvec^*\rVert_{L_2(\Prob)} \leq r}}
        \frac{1}{m}\left|\sum_{i=1}^m \epsilon_i \Big(\Lcal(\fvec(\tilde{\x}_i), y_i) - \Lcal(\fvec^*(\tilde{\x}_i), y_i) - \Prob(\Lcal_{\fvec} - \Lcal_{\fvec^*} ) 
        \Big) \right|
        \right] + 4r^2\\
        ~\stackrel{\text{(e)}}{\leq}~ & 64\tilde{R}\sqrt{k}\E_{\x,\y}\left[ Z_m(r) \right]+ 4r^2~\stackrel{\text{(f)}}{\leq}
        260r^2,
    \end{align*}
    where (d) follows by the standard Rademacher contraction inequality because by Lemma \ref{lem:lipschitz}, it holds that $|\Lcal(\fvec(\tilde{\x}_i), y_i) - \Lcal(\fvec^*(\tilde{\x}_i), y_i) - \Prob(\Lcal_{\fvec} - \Lcal_{\fvec^*} )| \leq 8\tilde{R}\sqrt{k}$,
    (e) uses the reverse symmetrization argument as in Proposition 4.11 in \cite{wainwright1945high},
    and (f) follows from \eqref{eqn:Zm(r)}.
    Therefore, using the Talagrand inequality \citep{wainwright1945high}, we conclude that for any $u > 0$ and $r \geq \delta_m \geq \delta^*_m$, 
    \begin{align}
    \label{eqn:concentration-z}
        \Prob\left(Z_m(r) \geq \frac{4\delta_m r}{\tilde{R}\sqrt{k}} + u\right) ~\leq~ c_1e^{-\frac{c_2 m u^2}{r^2 + \tilde{R}\sqrt{k}u}}.
    \end{align}

     Also, by assumption, it holds that $\lVert \fvec - \fvec^*\rVert_{L_2(\Prob)} \leq 2\tilde{R}\sqrt{k}$. 
    Let us consider a sequence of radius $r_\eta = 2^{\eta} \delta_m$ for $\eta = 0, \dots, M-1$, where $M = \left\lceil \log_2\left(\frac{2\tilde{R}\sqrt{k}}{\delta_m}\right) \right \rceil$, and let $r_M = 2\tilde{R}\sqrt{k}$. 
    For each $\eta \geq 1$, we let  $r = r_\eta$ and $u = \frac{\delta_m r_\eta}{\tilde{R}\sqrt{k}}$ in \eqref{eqn:concentration-z} to obtain $\Prob\left(Z_m(r_\eta) \geq \frac{5\delta_m r_{\eta}}{\tilde{R}\sqrt{k}}\right) \leq c_1 e^{-\frac{c_3 m \delta^2_m r^2_\eta/R^2 k}{r^2_\eta + \delta_m r_\eta}} \stackrel{\text{(g)}}{\leq} c_1 e^{-\frac{c_4 m \delta^2_m}{R^2k}}$, where (g) follows because $r_\eta \geq \delta_m$. 
    On $\cap_{\eta = 1}^M\left\{Z_m(r_\eta) \leq \frac{5\delta_m r_{\eta}}{\tilde{R}\sqrt{k}}\right\}$, it holds that for any $\fvec \in \Fscr^*$ such that $\lVert \fvec - \fvec^*\rVert_{L_2(\Prob)} > \delta_m$, 
    $$
    \big|
    \Prob_m(\Lcal_{\fvec} - \Lcal_{\fvec^*} ) - \Prob(\Lcal_{\fvec} - \Lcal_{\fvec^*} ) 
    \big|~\leq~ \frac{5\delta_m r_{\hat{\eta}}}{\tilde{R}\sqrt{k}} ~\stackrel{\text{(h)}}{\leq}~ \frac{10\delta_m\lVert \fvec - \fvec^*\rVert_{L_2(\Prob)}  }{\tilde{R}\sqrt{k}}, 
    $$
    where we let $\hat{\eta}$ be the smallest $\eta$ such that $r_{\eta} \geq \lVert \fvec - \fvec^*\rVert_{L_2(\Prob)} $. 
    Note that this implies that $r_{\hat{\eta}}/2 \leq \lVert \fvec - \fvec^*\rVert_{L_2(\Prob)}$, which enables (h). 
    Take a union bound over all $\eta = 1, \dots, M$, we observe that the event $\cap_{\eta = 1}^M\left\{Z_m(r_\eta) \leq \frac{5\delta_m r_{\eta}}{\tilde{R}\sqrt{k}}\right\}$ happens with probability 
    $1 - c_1e^{\log M -\frac{c_4 m \delta^2_m}{R^2k}} \geq 
    1 - c_1e^{-\frac{c_5 m \delta^2_m}{R^2k}}$, if $\log M \leq \frac{c_4 m \delta^2_m}{2R^2k}$ or equivalently 
    $m \delta^2_m \geq c_6 R^2k\log\left(\log\left(\frac{\tilde{R}\sqrt{k}}{\delta_m}\right)\right)$ for some absolute constant $c_6$.
    We summarize this key result next.
    
    \begin{lem} [Uniform Bound]
    \label{lem:uniform}
    If $\delta_m \geq \delta_m^*$ is such that $m \delta^2_m \geq c_6 R^2k\log\left(\log\left(\frac{\tilde{R}\sqrt{k}}{\delta_m}\right)\right)$, 
    $$
        \Prob\left(
          \big|
    \Prob_m(\Lcal_{\fvec} - \Lcal_{\fvec^*} ) - \Prob(\Lcal_{\fvec} - \Lcal_{\fvec^*} ) 
    \big|~\leq~  \frac{10\delta_m\lVert \fvec - \fvec^*\rVert_{L_2(\Prob)}  }{\tilde{R}\sqrt{k}}~~\forall~\lVert \fvec - \fvec^*\rVert_{L_2(\Prob)} > \delta_m 
        \right) \geq 1 - c_1e^{-\frac{c_5 m \delta^2_m}{R^2k}}.$$
    \end{lem}

    We also need the following result on the strong convexity of $\Prob(\Lcal_{\fvec} - \Lcal_{\fvec^*} )$.

    \begin{lem}[Strong Convexity]
    \label{lem:strong-convexity}
        It holds that $$
        \Prob(\Lcal_{\fvec} - \Lcal_{\fvec^*} ) ~\geq~ \frac{1}{2}e^{-\tilde{R}}(1 + ke^{\tilde{R}})^{-2} \lVert \fvec - \fvec^*\rVert_{L_2(\Prob)}
        $$
    \end{lem}

    In \eqref{eqn:concentration-z}, setting $r = \delta_m$ and $u = \frac{\delta^2_m}{\tilde{R}\sqrt{k}}$, we obtain $  \Prob\left(Z_m(\delta_m) \geq \frac{5\delta_m^2}{\tilde{R}\sqrt{k}}\right)\leq c_1e^{-\frac{c_7 m \delta_m^2}{R^2k}}.$
    Combining this with the previous two lemmas, and noting that $\Prob_m(\Lcal_{\fvec} - \Lcal_{\fvec^*} )\leq 0$, we have 
    \begin{align}
    \label{eqn:final-chain}
    \frac{1}{2}e^{-\tilde{R}}(1 + ke^{\tilde{R}})^{-2} \lVert \hat{\fvec} - \fvec^*\rVert_{L_2(\Prob)}^2~\leq~ \Prob(\Lcal_{\hat{\fvec}} - \Lcal_{\fvec^*} ) ~\leq~ \frac{10\delta_m\lVert \hat{\fvec} - \fvec^*\rVert_{L_2(\Prob)}  }{\tilde{R}\sqrt{k}} + \frac{5\delta_m^2}{\tilde{R}\sqrt{k}},
    \end{align}
    with probability at least $1 -
2c_1e^{-\frac{c_8 m \delta_m^2}{R^2k}}$.    After algebra (i.e., an application of Young's inequality), this implies
    $$
    \lVert \hat{\fvec} - \fvec^*\rVert_{L_2(\Prob)}^2~\leq~\frac{2\delta^2_m}{R^2{k}}\left(
    200e^{2\tilde{R}}(1 + ke^{\tilde{R}})^{4} 
    + 10e^{\tilde{R}}(1 + ke^{\tilde{R}})^{2}\tilde{R}\sqrt{k}
    \right),
    $$
    which by \eqref{eqn:final-chain} again, implies that 
    \begin{align}
        \label{eqn:pre-conclusion}
           \Prob(\Lcal_{\hat{\fvec}} - \Lcal_{\fvec^*} ) ~\leq~ \frac{\delta^2_m}{R^2 k}\left[
    10\sqrt{2}\left(
    200e^{2\tilde{R}}(1 + ke^{\tilde{R}})^{4} 
    + 10e^{\tilde{R}}(1 + ke^{\tilde{R}})^{2}\tilde{R}\sqrt{k}
    \right)^{\frac{1}{2}} + \tilde{R}\sqrt{k}
    \right]. 
    \end{align}
    Finally, the next result follows. The constant in this result is an abosolute one.
    \begin{lem}[Bounding the Localized complexity]
    \label{lem:local-complexity}
        It holds that 
        $\delta^*_m \preceq R\sqrt{\frac{kd}{m}}$. 
    \end{lem}
    In view of the conditions of Lemma \ref{lem:uniform}, it is therefore sufficient to take $\delta_m = c_9 R\sqrt{\frac{k\tilde{d} \log(m/\delta)}{m}}$ for some absolute constant $c_9$.
    Then, our desired conclusion  follows from \eqref{eqn:pre-conclusion} because $\Prob(\Lcal_{\hat{\fvec}} - \Lcal_{\fvec^*} ) = \E_{\x, z}\Big[
    {\sf KL}\big( \mathbf{g}(\x, z, \thetavec^*)\,\big\vert\,\mathbf{g}(\x, z, \hat{\thetavec})\big)
    \Big]$.
\hfill $\blacksquare$}
\vspace{3mm}

\subsection{Additional Supporting Lemmas and Proofs}

\begin{lem}[Derivative Computation]
\label{lem:derivative}
Suppose that $\alphavec \in \Real^{k+1}_+$ satisfies $\sum_{j\in\Kscr^+}\alpha_j = 1$. 
Then
$$
\nabla_{\betavec}\left(
\sum_{j\in\Kscr^+}\alpha_j \log \sigma_{j}(\x; \betavec)
\right) ~=~\sum_{j\in \Kscr}(\alpha_j - \sigma_{j}(\x; \betavec))\x_{(j)}
~~\text{ 
and }~~
\nabla_{\betavec}^2\left(
\sum_{j\in\Kscr^+}\alpha_j \log \sigma_{j}(\x; \betavec)
\right) ~=~-\Avec(\x; \betavec).
$$
\end{lem}

\noindent
\textbf{Proof.}
    The proof of this lemma follows from a straightforward computation and we skip the details. \hfill $\blacksquare$
    \vspace{3mm}

\begin{lem}[Bounds on the Quadratic Form]
\label{lem:eigen}
    Let us fix any $\x\in \Xscr$, $\betavec \in \Real^{k}$,
    and $\uvec \in \Real^{k}$.
    $$\sigma_{0}(\x; \betavec)
    \min_{j\in \Kscr}\sigma_{j}(\x; \betavec)\sum_{j\in \Kscr}(\x_{(j)} \uvec)^2~\leq~ \uvec^{\top}\Avec(\x; \betavec) \uvec ~\leq~ \sigma_{0}(\x; \betavec)\max_{j\in \Kscr}\sigma_{j}(\x; \betavec)\sum_{j\in \Kscr}(\x_{(j)} \uvec)^2.
    $$
\end{lem}

\noindent
\textbf{Proof.}
    Define matrix $\bm{\Sigma}(\x; \betavec) \in \Real^{d\times d}$ such that $\bm{\Sigma}_{jj'}(\x; \betavec)=\ind_{\{j = j'\}}\sigma_{j}(\x; \betavec^*) - \sigma_{j}(\x; \betavec)\sigma_{j'}(\x, \betavec)$ for each $1\leq j,j' \leq d$. 
    Then, $\uvec^{\top}\Avec(\x; \betavec) \uvec = \sum_{j\in \Kscr}\sum_{j'\in \Kscr}\bm{\Sigma}_{jj'}(\x; \betavec)(\x_{(j)}^{\top} \uvec)(\x_{(j)} ^{\top}\uvec)$.
    Therefore, it suffices to  show that $$\sigma_{0}(\x; \betavec) \min_{j\in \Kscr}\sigma_{j}(\x; \betavec) \Ivec ~\preceq~ \bm{\Sigma}(\x; \betavec) ~\preceq~ \sigma_{0}(\x; \betavec)\max_{j\in \Kscr}\sigma_{j}(\x; \betavec) \Ivec.$$
    Indeed, by the Gershgorin circle theorem, it holds that any eigenvalue $\lambda$ of $\bm{\Sigma}$ satisfies 
    \begin{align*}
        &\sigma_{0}(\x; \betavec)\min_{j\in \Kscr}\sigma_{j}(\x; \betavec)~=~\min_{j\in \Kscr} \left\{\sigma_{j}(\x; \betavec)- \sum_{j' \in \Kscr}\sigma_{j}(\x; \betavec)\sigma_{j'}(\x; \betavec) \right\}~\leq~ \lambda\\
        &~~~~~~~~~~~~~~~~~~~~~~~~~~~~~~~~~~~~~~~~~~~~~~~~~~~~~~~~~~~~\leq~ \max_{j\in \Kscr} \left\{\sigma_{j}(\x; \betavec)- \sum_{j' \in \Kscr}\sigma_{j}\sigma_{j'}(\x; \betavec) \right\} = \sigma_{0}(\x; \betavec)\max_{j\in \Kscr}\sigma_{j}(\x; \betavec).
    \end{align*}
    Therefore, we conclude the proof.\hfill $\blacksquare$
    
    \vspace{3mm}

    \revise{\textbf{Proof of Theorem \ref{thm:aare}.}
We will first prove the first item of Theorem \ref{thm:aare}.
Whenever $\sum_{i=1}^n\x_i \x_i^{\top}$ is invertible, it holds that 
$$
\hat{\betavec}^{\sf AARE} ~=~ \left(\frac{1}{n}\sum_{i=1}^n\x_i \x_i^{\top}\right)^{-1} \cdot\frac{1}{n}\sum_{i=1}^n\x_i g(\x_i, z_i; \hat{\thetavec}). 
$$
Since $\ddot{\mathbf{\Omega}} = \E[\x \x^{\top}] \succ 0$, it holds that $\frac{1}{n}\sum_{i=1}^n \x_i \x_i^{\top}\pconverge \ddot{\mathbf{\Omega}}$, so one can verify that 
\begin{align*}
    \hat{\betavec}^{\sf AARE} ~&=~ \ddot{\mathbf{\Omega}}^{-1} \cdot\frac{1}{n}\sum_{i=1}^n\x_i g(\x_i, z_i; \hat{\thetavec}) + o_{\text{P}}(1)\\
    ~&=~ \betavec^* + \ddot{\mathbf{\Omega}}^{-1}\cdot \frac{1}{n}\sum_{i=1}^n\x_i \bigg[
    \left(g(\x_i, z_i; \hat{\thetavec})-g(\x_i, z_i, {\thetavec}^*)\right)+
    \left(g(\x_i, z_i, {\thetavec}^*) - \x_i^\top \betavec^*\right)\bigg] + o_{\text{P}}(1).
\end{align*}
Note that the first-order condition of \eqref{eqn:regression-best-in-class} implies that $\E\bigg[\left(g(\x, z; \theta^*) - \x^{\top}\betavec^*\right)\x\bigg] = 0$ so $\frac{1}{n}\sum_{i=1}^n\left(g(\x_i, z_i, {\thetavec}^*) - \x_i^\top \betavec^*\right)\x_i \pconverge 0$.
Furthermore, since $\hat{\thetavec} \pconverge \thetavec^*$, it holds that there is a sequence of positive constants $\{\epsilon_n\}_{n=1}^\infty$ that converges to zero such that $\Prob\left(
\big \lVert
\hat{\thetavec} - \thetavec^*
\big \rVert > \epsilon_n
\right) \rightarrow 0$.
By Markov's inequality, for any fixed $\epsilon > 0$, 
\begin{align}
\label{eqn:chain-x-times-g}
    & \Prob\left(\left\lVert\frac{1}{n}\sum_{i=1}^n\x_i 
    \left(g(\x_i, z_i; \hat{\thetavec})-g(\x_i, z_i, {\thetavec}^*)\right)\right \rVert\geq \epsilon\right)\\
    &~~~~~~~~~~~~~~~~~~\leq~\Prob\left(\sup_{{\thetavec}: \big \lVert
{\thetavec} - \thetavec^*
\big \rVert \leq \epsilon_n}\left\lVert\frac{1}{n}\sum_{i=1}^n\x_i 
    \left(g(\x_i, z_i, {\thetavec})-g(\x_i, z_i, {\thetavec}^*)\right)\right \rVert\geq \epsilon\right) + \Prob\left(
\big \lVert
\hat{\thetavec} - \thetavec^*
\big \rVert > \epsilon_n
\right) \\
&~~~~~~~~~~~~~~~~~~\leq~ 
\frac{1}{\epsilon}\E\left[\sup_{{\thetavec}: \big \lVert
{\thetavec} - \thetavec^*
\big \rVert \leq \epsilon_n}\left\lVert\frac{1}{n}\sum_{i=1}^n\x_i 
    \left(g(\x_i, z_i, {\thetavec})-g(\x_i, z_i, {\thetavec}^*)\right)\right \rVert\right]
+ \Prob\left(
\big \lVert
\hat{\thetavec} - \thetavec^*
\big \rVert > \epsilon_n
\right) \\
&~~~~~~~~~~~~~~~~~~\stackrel{\text{(a)}}{\leq}~ 
\frac{1}{\epsilon}\E\left[\sup_{{\thetavec}: \big \lVert
{\thetavec} - \thetavec^*
\big \rVert \leq \epsilon_n}\big\lVert \x 
    \left(g(\x, z, {\thetavec})-g(\x, z, {\thetavec}^*)\right)\big \rVert\right]
+ \Prob\left(
\big \lVert
\hat{\thetavec} - \thetavec^*
\big \rVert > \epsilon_n
\right), 
\end{align}
where (a) follows by triangular inequality and Jensen's inequality. 
By the continuity of $g(\cdot)$ in $\thetavec$ as stated in the assumption, it follows that $\sup_{{\thetavec}: \big \lVert
{\thetavec} - \thetavec^*
\big \rVert \leq \epsilon_n}\big\lVert \x 
    \left(g(\x, z, {\thetavec})-g(\x, z, {\thetavec}^*)\right)\big \rVert$ converges to zero when $n$ goes to infinity. 
    Since $\,\E\big[\sup_{\thetavec \subset \Nscr}\lVert  \x g(\x, z; \thetavec) \rVert_2\big] \,<\, \infty$ and $\epsilon_n$ falls into $\Nscr$ eventually, by the dominated convergence theorem, we have that $\E\left[\sup_{{\thetavec}: \big \lVert
{\thetavec} - \thetavec^*
\big \rVert \leq \epsilon_n}\big\lVert \x 
    \left(g(\x, z, {\thetavec})-g(\x, z, {\thetavec}^*)\right)\big \rVert\right]$ converges to zero. 
    This concludes the proof for the first item. 

    For the second item, we note that 
    \begin{align*}
        & \sqrt{n}\left(\hat{\betavec}^{\sf AARE} - \betavec^* \right)~=~ \ddot{\mathbf{\Omega}}^{-1} \cdot\frac{1}{\sqrt{n}}\sum_{i=1}^n\x_i \left(g(\x_i, z_i; \tilde{\thetavec}) - \x_i^{\top} \betavec_i\right) + o_{\text{P}}(1)\\&~~~~~~~~~~~~~~=~ \ddot{\mathbf{\Omega}}^{-1} \cdot\left[\frac{1}{\sqrt{n}}\sum_{i=1}^n\x_i \left(g(\x_i, z_i; {\thetavec}^*)
        - \x_i^\top \betavec^*
        \right)+ \frac{1}{n}\sum_{i=1}^n\x_i \nabla g(\x_i, z_i; \hat{\thetavec})^{\top}\times 
        \sqrt{\frac{n}{m}} \times \sqrt{m}\big(
    \hat{\thetavec} - \thetavec^*
    \big)\right] + o_{\text{P}}(1),
    \end{align*}
    where $\tilde{\thetavec} \in [\thetavec^*, \hat{\thetavec}]$.
    Using an argument similar to \eqref{eqn:chain-x-times-g}, we can show that  
    $$\frac{1}{n}\sum_{i=1}^n\x_i \nabla g(\x_i, z_i; \hat{\thetavec})^{\top} \pconverge \E\left[\x \nabla g(\x, z;{\thetavec}^*)^{\top}  \right].$$
    Therefore, the conclusion follows. 

    For the third item, we first note that 
    \begin{align*}
        \Sigmavec &=
\E
\left[
\left[(y - \psivec(\x, z)^{\top}\thetavec^*) + (\psivec(\x, z)^{\top} \thetavec^*- \x^{\top}\betavec)\right]^2 \x\x^{\top}
\right]\\
&\stackrel{\text{(b)}}{=}
\E
\left[(y - \psivec(\x, z)^{\top}\thetavec^*)^2 \x\x^{\top}
\right] + \E
\left[(\psivec(\x, z)^{\top} \thetavec^*- \x^{\top}\betavec)^2 \x\x^{\top}
\right]\stackrel{\text{(c)}}{=}\sigma^2\E
\left[\x\x^{\top}
\right] + \E
\left[(\psivec(\x, z)^{\top} \thetavec^*- \x^{\top}\betavec)^2 \x\x^{\top}
\right],
    \end{align*}
where (b) follows because 
$\E[y - \psivec(\x, z)^{\top} \thetavec^* \,|\, \x, z] = 0$ and (c) follows from the homoscedasticity assumption.
Following standard analysis of M-estimators \citep{newey1994large}, under the current conditions, one can compute the  asymptotic variance of 
$\hat{\thetavec}$ as 
$\ddot{\bm{\Lambda}} = \bm{\Lambda}_1^{-1}\bm{\Lambda}_2  \bm{\Lambda}_1^{-1}$. 
Here
$$
\bm{\Lambda}_1 ~=~ \E\Big[
\nabla g(\x, z; \thetavec^*)\nabla g(\x, z; \thetavec^*)^{\top} + 
\left(g(\x, z; \thetavec^*) - y \right)
\nabla^2_{\thetavec} g(\x, z; \thetavec^*)
\Big]~\stackrel{\text{(d)}}{=}~
\E\left[
\nabla_{\thetavec} g(\x, z; \thetavec^*)\nabla_{\thetavec} g(\x, z; \thetavec^*)^{\top}\right], 
$$
where (d) follows because $g(\x, z; \thetavec^*) = \E[y\,|\,\x, z]$, and 
$$
\bm{\Lambda}_2 ~=~ \E
\left[
\left(y - g(\x, z; \thetavec^*)\right)^2\nabla_{\thetavec} g(\x, z; \thetavec^*)\nabla_{\thetavec} g(\x, z; \thetavec^*)^{\top} 
\right]~\stackrel{\text{e}}{=}~\sigma^2\E
\left[
\nabla_{\thetavec} g(\x, z; \thetavec^*)\nabla_{\thetavec} g(\x, z; \thetavec^*)^{\top}
\right],
$$
where (e) follows from the homoscadesticity assumption. 
Therefore, 
$\ddot{\Lambdavec} =\sigma^2 \E\left[\nabla_{\thetavec} g(\x, z; \thetavec^*)\nabla_{\thetavec} g(\x, z; \thetavec^*)^{\top}\right]^{-1}$. 
Then, it holds that 
$$
\Sigmavec ~=~
\sigma^2\E
\left[\x\x^{\top}
\right] ~\succeq~ \ddot{\Gammavec}\ddot{\Lambdavec}\ddot{\Gammavec}^{\top}$$ using the same argument in Proposition  \ref{prop:weak-dom} (i.e., setting $\uvec = \x$ and $\wvec = \nabla_{\thetavec} g(\x, z; \thetavec^*)$). 
Thus, the conclusion follows.
\hfill $\blacksquare$
\vspace{3mm}}

\noindent
\textbf{Proof of Lemma \ref{lem:unique-optimizer}.}
We first show that an optimizer exists. 
Assume that there exists $\{\betavec^{(\ell)}\}_{\ell=1}^{\infty}$ such that 
$$\lim_{\ell \rightarrow \infty}\E_{\x}\Big[
    {\sf KL}\big( \Prob(y\,|\, \x)\,\big\vert\,\sigma_{y}(\x, \betavec^{(\ell)}) \big)
    \Big] ~=~ \inf_{\betavec \in \Real^{kd}}\E_{\x}\Big[
    {\sf KL}\big( \Prob(y\,|\, \x)\,\big\vert\,\sigma_{y}(\x, \betavec) \big)
    \Big], $$
or equivalently 
\begin{align}
\label{eqn:optimality-sequence}
    \lim_{\ell \rightarrow \infty}\E_{\x}\left[
    \sum_{j\in \Kscr^+} \Prob\big(y = j| \x\big)\log\sigma_{j}(\x, \betavec^{(\ell)})
    \right] ~=~\sup_{\betavec \in \Real^{k}}Q(\betavec)~=~ \sup_{\betavec \in \Real^{k}}\E_{\x}\left[
    \sum_{j\in \Kscr^+} \Prob\big(y = j| \x\big)\log\sigma_{j}(\x, \betavec^{(\ell)})
    \right],
\end{align}
If there is a subsequence of $\{\betavec^{(\ell)}\}_{\ell=1}^{\infty}$ that is bounded, by the continuity of $\E_{\x}\Big[
{\sf KL}\big( \Prob(y\,|\, \x)\,\big\vert\,\sigma_{y}(\x, \betavec) \big)\Big]$, there must exist a limit of a subsequence of $\{\betavec^{(\ell)}\}_{\ell=1}^{\infty}$ such that the infimum is achieved. 
This proves the existence. 

Otherwise, let us assume that $\{\betavec^{(\ell)}\}_{\ell=1}^{\infty}$ is unbounded, so $\lim_{\ell \rightarrow\infty}\lVert \betavec^{(\ell)}\rVert_2 = \infty$. 
We will show that this is impossible, by way of contradiction. 
Since $\{\betavec^{(\ell)}/\lVert \betavec^{(\ell)} \rVert_2\}_{\ell=1}^{\infty}$ is a bounded sequence, by taking a subsequence if necessary, we can assume without loss of generality that $\lim_{\ell \rightarrow\infty}\betavec^{(\ell)}/\lVert \betavec^{(\ell)}\rVert_2 = \tilde{\betavec}$ with $\lVert \tilde{\betavec}\rVert_2 = 1$.
Thus, 
\begin{align}
\label{eqn:bound-infty}
\lim_{\ell \rightarrow\infty}\E_{\x}\left[\max_{j \in \Kscr}\big|(\betavec^{(\ell)})^{\top}\x_{(j)}\big|\right] &~\geq~ \frac{1}{k}\lim_{\ell \rightarrow\infty}\E_{\x}\left[\sum_{j\in \Kscr}\big|(\betavec^{(\ell)})^{\top}\x_{(j)}\big|\right]~=~ \frac{1}{k}\lim_{\ell \rightarrow\infty}\lVert \betavec^{(\ell)} \rVert_2\cdot\E_{\x}\left[\sum_{j\in \Kscr}\left|\frac{(\betavec^{(\ell)})^{\top}}{\lVert \betavec^{(\ell)} \rVert_2}\x_{(j)}\right|\right]
    \nonumber\\&~\stackrel{\text{(a)}}{=}~\frac{1}{k}\E_{\x}\left[\sum_{j\in \Kscr}\left|\tilde{\betavec}^{\top}\x_{(j)}\right|\right] \cdot\lim_{\ell \rightarrow\infty}\lVert \betavec^{(\ell)} \rVert_2~=~\frac{1}{k}\E_{\x}\left[\sum_{j\in \Kscr}\left|\tilde{\betavec}^{\top}\x_{(j)}\right|\right] \cdot+\infty~\stackrel{\text{(b)}}{=}~+\infty,
\end{align}
where (a) follows from the bounded convergence theorem because by assumption $\x$ is bounded. 
To see (b), suppose that $\E_{\x}\left[\sum_{j\in \Kscr}\left|\tilde{\betavec}^{\top}\x_{(j)}\right|\right] = 0$. Thus, $\tilde{\betavec}^{\top}\x_{(j)} = 0$ for all $j \in \Kscr$ almost everywhere. 
Thus, if we let $\evec$ denote a vector of ones of dimension $k$, 
we have $(\betavec\otimes\evec)\E_{\x}[\x\x^{\top}](\betavec\otimes\evec)^{\top} = 0$, which contradicts the assumption that $\E_{\x}[\x\x^{\top}]$ is positive definite. 

We note that by the Taylor expansion, for any $u\in \Real$ and $v \in [0, k]$, 
\begin{align}
\label{eqn:taylor}
    \log \frac{e^u}{v + e^u} ~=~ u - \log\left(v + e^u\right)~=~ u - 
\left[
\log(v+1) + \frac{u}{v+1} +  \frac{ve^{{w}}u^2}{2(e^w +v
)^2}\right]
~\stackrel{\text{(c)}}{\leq}~ \frac{ku}{k+1},
\end{align}
where $w\in [0, u]$ and (c) follows because $\frac{ve^{{w}}u^2}{2(e^w +v
)^2} \geq 0$, $\log(v+1) \geq 0$ and $v\leq k$.   

Next, fix $\ell = 1, \dots, \infty$ and $\x \in \Xscr$. 
We assume that $\Tilde{j} \in \Kscr$ is such that $\Tilde{j} = \argmax_{j \in \Kscr}\big|(\betavec^{(\ell)})^{\top}\x_{(j)}\big|$.
If $(\betavec^{(\ell)})^{\top}\x_{(\Tilde{j})} > 0$,
it follows that 
\begin{align*}
    &\sum_{j\in \Kscr^+} \Prob\big(y = j| \x\big)\log\sigma_{j}(\x, \betavec^{(\ell)}) ~\leq~ \Prob\big(y = 0\,|\, \x\big)\log\sigma_{0}(\x, \betavec^{(\ell)})\\
    &~~~~~~~~~~~~~~~~~~=~\Prob\big(y = 0\,|\, \x\big)\log \left(
    \frac{e^{-(\betavec^{(\ell)})^{\top}\x_{(\Tilde{j})}}}{e^{-(\betavec^{(\ell)})^{\top}\x_{(\Tilde{j})}} + \sum_{j\in \Kscr}e^{-(\betavec^{(\ell)})^{\top}\x_{(\Tilde{j})}+(\betavec^{(\ell)})^{\top}\x_{(j)}}}
    \right) ~\stackrel{\text{(d)}}{\leq}~ \frac{-k\big|(\betavec^{(\ell)})^{\top}\x_{(\Tilde{j})}\big|}{k+1},
\end{align*}
where (d) follows from \eqref{eqn:taylor} because $-(\betavec^{(\ell)})^{\top}\x_{(\Tilde{j})}+(\betavec^{(\ell)})^{\top}\x_{(j)} \leq 0$ and therefore
$\sum_{j\in \Kscr}e^{-(\betavec^{(\ell)})^{\top}\x_{(\Tilde{j})}+(\betavec^{(\ell)})^{\top}\x_{(j)}} \in [0, k]$.
Suppose $(\betavec^{(\ell)})^{\top}\x_{(j)} \leq 0$ for all $j \in \Kscr$. 
Then, we have
\begin{align*}
    &\sum_{j\in \Kscr^+} \Prob\big(y = j\,|\, \x\big)\log\sigma_{j}(\x, \betavec^{(\ell)}) ~\leq~ \Prob\big(y = \Tilde{j}\,|\, \x\big)\log\sigma_{\Tilde{j}}(\x, \betavec^{(\ell)})\\
    &~~~~~~~~~~~~~~~~~~~~~~~~~~~~~~~~~~~~~~~~~~=~\Prob\big(y = \Tilde{j}\,|\, \x\big)\log \left(
    \frac{e^{(\betavec^{(\ell)})^{\top}\x_{(\Tilde{j})}}}{1 + \sum_{j\in \Kscr}e^{(\betavec^{(\ell)})^{\top}\x_{({j})}}}
    \right) ~\stackrel{\text{(e)}}{\leq}~ \frac{-k\big|(\betavec^{(\ell)})^{\top}\x_{(\Tilde{j})}\big|}{k+1},
\end{align*}
where (e) follows from \eqref{eqn:taylor}.
Suppose $(\betavec^{(\ell)})^{\top}\x_{(\Tilde{j})} \leq 0$ but $(\betavec^{(\ell)})^{\top}\x_{({j}'')} > 0$ for some $j'' \in \Kscr$ and $(\betavec^{(\ell)})^{\top}\x_{({j}'')} \geq (\betavec^{(\ell)})^{\top}\x_{({j})}$ for all $j \in \Kscr$.
In this case, 
\begin{align*}
    &~~~\sum_{j\in \Kscr^+} \Prob\big(y = j\,|\, \x\big)\log\sigma_{j}(\x, \betavec^{(\ell)}) ~\leq~ \Prob\big(y = \tilde{j}\,|\, \x\big)\log\sigma_{\Tilde{j}}(\x, \betavec^{(\ell)})\\
    &~=~\Prob\big(y = \Tilde{j}\,|\, \x\big)\log \left(
    \frac{e^{(\betavec^{(\ell)})^{\top}\x_{(\tilde{j})} - (\betavec^{(\ell)})^{\top}\x_{({j}'')}}}{e^{-(\betavec^{(\ell)})^{\top}\x_{({j}'')})^{\top}\x} + \sum_{j\in \Kscr}e^{-(\betavec^{(\ell)})^{\top}\x_{({j}'')}+(\betavec^{(\ell)})^{\top}\x_{({j})}}}
    \right) ~\stackrel{\text{(f)}}{\leq}~ 
    \frac{k\big((\betavec^{(\ell)})^{\top}\x_{(\tilde{j})} - (\betavec^{(\ell)})^{\top}\x_{({j}'')}x\big)}{k+1}\\&~\leq~
    \frac{-k\big|(\betavec^{(\ell)}_{\Tilde{j}})^{\top}\x\big|}{k+1},
\end{align*}
where (f) follows from \eqref{eqn:taylor} again. 
To summarize, taking the expectation, using \eqref{eqn:bound-infty}, we have 
$$
\lim_{\ell \rightarrow \infty}\E_{\x}\left[\sum_{j\in \Kscr^+} \Prob\big(y = j| \x\big)\log\sigma_{j}(\x, \betavec^{(\ell)})\right]~\leq~-\lim_{\ell \rightarrow \infty} 
\E_{\x}\left[\frac{k\max_{j \in \Kscr}\big|(\betavec^{(\ell)})^{\top}\x_{({j})}\big|}{k+1}\right] = -\infty.
$$
This contradicts \eqref{eqn:optimality-sequence}. Therefore, $\{\betavec^{(\ell)}\}_{\ell=1}^{\infty}$ cannot be unbounded. This finishes the proof of existence. 

Next, using Lemma \ref{lem:derivative}, we notice the Hessian of $\E_{\x}\left[
    \sum_{j\in \Kscr^+} \Prob\big(y = j| \x\big)\log\sigma_{j}(\x; \betavec)
    \right]$ is $-\widetilde{\Omegavec}(\betavec)$, where
    $\widetilde{\Omegavec}(\betavec) ~:=~ \E_{\x}\left[\Avec(\x;\betavec) \right] \in \Real^{k \times k}$, which is analogous to $\Omegavec$ defined in  \eqref{eqn:Omega-Gamma}.
Thus, take any $\qvec \in \Real^{k}$ such that $\lVert \qvec\rVert_2 = 1$.
Thus, for any $\betavec \in \Real^{k}$
\begin{align*}
   &\qvec^{\top} \left(-\nabla^2_{\betavec}\E_{\x}\left[
    \sum_{j\in \Kscr^+} \Prob\big(y = j| \x\big)\log\sigma_{j}(\x; \betavec)
    \right]\right)\qvec\,=\,\E_{\x}\left[\qvec^{\top}
    \Avec(\x; \betavec)\qvec^{\top}
    \right]\stackrel{\text{(g)}}{\geq}
\E_{\x}\left[\sigma_{0}(\x; \betavec) \min_{j\in \Kscr}\sigma_{j}(\x; \betavec)
    \sum_{j\in \Kscr}(\x^{\top}\qvec)^2\right]\\&~~~~~~~~~~~~~~~~~~~~~~~~~~~~~~~~~~~~~~~~~~~~~~~~\stackrel{\text{(h)}}{\succsim}\sum_{j\in \Kscr}\E_{\x}\left[(\x^{\top}\qvec)^2
    \right] \stackrel{\text{(i)}}{\geq}\, \lambda_{\min}\left(\E_{\x}[\x \x^{\top}]\right)k\lVert\qvec\rVert^2 = k\lambda_{\min}\left(\E_{\x}[\x \x^{\top}]\right),
\end{align*}
where (g) follows from Lemma \ref{lem:eigen}, (h) follows because $\x$ is bounded, and in (i), $\lambda_{\min}\left(\E_{\x}[\x \x^{\top}]\right)$ is the smallest eigenvalue of $\E_{\x}[\x \x^{\top}]$.
Therefore, the function $\E_{\x}\left[
    \sum_{j\in \Kscr^+} \Prob\big(y = j| \x\big)\log\sigma_{j}(\x; \betavec)
    \right]$ is strictly concave in $\betavec$ and has a unique maximum. 
    This concludes the proof. 
\hfill $\blacksquare$

\vspace{3mm}

\noindent
\textbf{Proof of Lemma \ref{lem:uniform-convergence}.}
Our proof starts with the decomposition: 
\begin{align}
\label{eqn:decomposition}
     \sup_{\betavec \in \Bscr(\betavec^*, \epsilon)}
    \left|
    \widehat{Q}(\hat{\thetavec};\betavec) - {Q}(\betavec)
    \right| ~\leq~  \sup_{\betavec \in \Bscr(\betavec^*, \epsilon)}
    \left|
    \widehat{Q}(\hat{\thetavec};\betavec) -  \widehat{Q}({\theta}^*;\betavec)
    \right| + \sup_{\betavec \in \Bscr(\betavec^*, \epsilon)}
    \left|
    \widehat{Q}({\theta}^*;\betavec) - {Q}(\betavec)
    \right|.
\end{align}
Then, 
\begin{align*}
     \sup_{\betavec \in \Bscr(\betavec^*, \epsilon)}
    \left|
    \widehat{Q}(\hat{\thetavec};\betavec) -  \widehat{Q}({\theta}^*;\betavec)
    \right| &= \sup_{\betavec \in \Bscr(\betavec^*, \epsilon)}\left|
    \frac{1}{n}\sum_{i=1}^n\sum_{j\in \Kscr^+}g_j(\x_i, z_i; \hat{\thetavec})\log \sigma_j(\x; \betavec)
    - \frac{1}{n}\sum_{i=1}^n\sum_{j\in \Kscr^+}g_j(\x_i, z_i; {\thetavec}^*)\log \sigma_j(\x; \betavec)
    \right|\\
    &\leq 
    \left(\sup_{\x \in \Xscr,\, \betavec \in \Bscr(\betavec^*, \epsilon),\, j\in \Kscr^+}
    \left|
    \log\sigma_j(\x; \betavec)
    \right|\right)\times
    \sum_{j\in \Kscr^+}\frac{1}{n}\sum_{i=1}^n\left|
    g_j(\x_i, z_i; \hat{\thetavec}) - g_j(\x_i, z_i; {\thetavec}^*)
    \right|\\
    &\stackrel{(a)}{=} 
    C_1
    \sum_{j\in \Kscr^+}\frac{1}{n}\sum_{i=1}^n\left|
    g_j(\x_i, z_i; \hat{\thetavec}) - g_j(\x_i, z_i; {\thetavec}^*)
    \right|,
\end{align*}
where in $(a)$ we simply define constant $C_1 = \sup_{\x \in \Xscr,\, \betavec \in \Bscr(\betavec^*, \epsilon),\, j\in \Kscr^+}
\left|\log\sigma_j(\x; \betavec)\right| < \infty$ because $\Xscr$ is bounded. 
Next, fix any $\epsilon' > 0$ and any $j \in \Kscr^+$. 

Consider any sequence $\{ \tilde{\thetavec}_p\}_{p=1}^{\infty}$ with $\tilde{\thetavec}_p\rightarrow \thetavec^*$. By the continuity of $g_j(\x, z; {\thetavec})$ in ${\thetavec}$, and the bounded convergence theorem, it must be that 
$$
\E_{\x, z}
\big[
\left|
g_j(\x, z; {\thetavec}_p) - g_j(\x, z; {\thetavec}^*)
\right|
\big]~\rightarrow~ 
\E_{\x, z}
\big[
\left|
g_j(\x, z; {\thetavec}^*) - g_j(\x, z; {\thetavec}^*)
\right|
\big] = 0. 
$$
In other words, the function $\E_{\x, z}
\big[
\left|
g_j(\x, z; {\thetavec}) - g_j(\x, z; {\thetavec}^*)
\right|
\big]$ is continuous in $\thetavec$. 
By the continuous mapping theorem, since $\hat{\thetavec} \pconverge \thetavec^*$, and the  Markov's inequality, we have 
$$
\Prob\left(
\frac{1}{n}\sum_{i=1}^n\left|
    g_j(\x_i, z_i; \hat{\thetavec}) - g_j(\x_i, z_i; {\thetavec}^*)
    \right| \geq \epsilon' \,\bigg|\, \hat{\thetavec}
\right)
~\leq~ 
\frac{1}{\epsilon'}
\E_{\x, z}
\left[
\left|
g_j(\x, z; \hat{\thetavec}) - g_j(\x, z; {\thetavec}^*)
\right|\,\bigg|\, \hat{\thetavec}
\right] \pconverge 0.
$$

Then we notice that $\Prob\left(
\frac{1}{n}\sum_{i=1}^n\left|
    g_j(\x_i, z_i; \hat{\thetavec}) - g_j(\x_i, z_i; {\thetavec}^*)
    \right| \geq \epsilon' \,\bigg|\, \hat{\thetavec}
\right) \leq 1$ so it is uniformly integrable. 
Therefore,
\begin{align*}
    & \Prob\left(
\frac{1}{n}\sum_{i=1}^n\left|
    g_j(\x_i, z_i; \hat{\thetavec}) - g_j(\x_i, z_i; {\thetavec}^*)
    \right| \geq \epsilon' 
\right) ~=~\E\left[\Prob\left(
\frac{1}{n}\sum_{i=1}^n\left|
    g_j(\x_i, z_i; \hat{\thetavec}) - g_j(\x_i, z_i; {\thetavec}^*)
    \right| \geq \epsilon' \,\bigg|\, \hat{\thetavec}
\right)\right]
\end{align*}
and the right-hand side converges to zero. 
Therefore, 
\begin{align}
\label{eqn:1st-term-convergence}
    \sup_{\betavec \in \Bscr(\betavec^*, \epsilon)}
    \left|
    \widehat{Q}(\hat{\thetavec};\betavec) -  \widehat{Q}({\theta}^*;\betavec)
    \right|~\pconverge~ 0.
\end{align}
We note that $ \widehat{Q}({\theta}^*;\betavec)$ converges to ${Q}(\betavec)$ for each $\betavec \in \Real^d$ by the law of large numbers. 
Then since $\widehat{Q}({\theta}^*;\betavec)$ is concave in $\betavec$, by Theorem 2.7 in \cite{newey1994large}, it holds that on the compact set 
$\Bscr(\betavec^*, \epsilon)$, 
\begin{align}
\label{eqn:2nd-term-convergence}
    \sup_{\betavec \in \Bscr(\betavec^*, \epsilon)}
    \left|
    \widehat{Q}({\theta}^*;\betavec) - {Q}(\betavec)
    \right| \pconverge 0.
\end{align}
Combining \eqref{eqn:decomposition}, \eqref{eqn:1st-term-convergence} and \eqref{eqn:2nd-term-convergence}, the conclusion follows. 
\hfill $\blacksquare$
\vspace{3mm}

\noindent
\textbf{Proof of Lemma \ref{lem:Omega}.}
For any $\x \in \Xscr$ and $\betavec \in \Real^{d}$, we note that 
$$
\sum_{j\in \Kscr^+} g_j(\x, z;{\thetavec})
\nabla^2_{\betavec}\log \sigma_j(\x; {\betavec})~=~
\nabla^2_{\betavec}\left(\sum_{j\in \Kscr^+} 
g_j(\x, z;{\thetavec})
\right)\log \sigma_j(\x; {\betavec})
~\stackrel{\text{(a)}}{=}~ -\Avec(\x; \betavec),$$
where (a) follows from Lemma \ref{lem:derivative} because $\sum_{j\in \Kscr^+} g_j(\x, z;{\thetavec}) = 1$. 
Note that $\Avec(\x_i;{\betavec})$ is continuous in ${\betavec}$.
Also, by Lemma \ref{lem:eigen}, for any $\uvec \in \Real^{d}$  with $\Vert \uvec \rVert_2 = 1$, we have
$$
\uvec^{\top}\Avec(\x; \betavec)\uvec ~\leq~ \sigma_{0}(\x; \betavec)\max_{j\in \Kscr}\sigma_{j}(\x; \betavec)\sum_{j\in \Kscr}(\x_{(j)} \uvec)^2~\leq~ \sum_{j\in \Kscr}(\x_{(j)} \uvec)^2~\leq ~\lVert \x \rVert_2^2 \Vert \uvec \rVert_2^2 \,=\, \lVert \x \rVert_2^2.
$$
Therefore, $\E_{\x}\left[\sup_{\betavec\in \Real^{d}} \lVert - \Avec(\x; \betavec) \rVert_2\right] \leq \E_{\x}\left[\lVert \x \rVert_2^2\right] = \text{trace}(\E_{\x}[\x \x^{\top}]) < \infty$.
By Theorem 4.3 of \cite{newey1994large} and the fact that $\tilde{\betavec} \pconverge \betavec^*$, the convergence result in the first display of the lemma follows. 

Fix any $\x \in \Xscr$ and $\thetavec \in \Real^q$. 
A direct computation implies that for any $j \in \Kscr^+$
$$
\nabla_{\betavec}\log \sigma_j(\x; {\betavec}^*) \nabla_{\thetavec}g_j(\x, z; {\thetavec})^{\top} ~=~
\left(\ind_{\{j\neq 0\}}\x_j - \sum_{j' \in \Kscr}\sigma_{j'}(\x, \betavec^*)\x_{j'}\right)\nabla_{\thetavec}g_j(\x, z; {\thetavec})^{\top}. 
$$
Thus, due to the assumption that $\x$ is bounded, 
$$
\Big\lVert \nabla_{\betavec}\log \sigma_j(\x; {\betavec}^*) \nabla_{\thetavec}g_j(\x, z; {\thetavec})^{\top} \Big\rVert_2 ~\precsim~  \Big\lVert\nabla_{\thetavec}g_j(\x, z; {\thetavec})^{\top} \Big\rVert_2.
$$
Therefore, 
\begin{align*}
    \E_{\x,z} \left[\sup_{\thetavec \in \Nscr}
\Big\lVert \nabla_{\betavec}\log \sigma_j(\x; {\betavec}^*) \nabla_{\thetavec}g_j(\x, z; {\thetavec})^{\top} \Big\rVert_2 \right]~\precsim~  
\E_{\x,z}\left[\sup_{\thetavec \subset \Nscr}\lVert \nabla_{\thetavec} g_j(\x, z, \thetavec) \rVert_2\right] \,\stackrel{\text{(b)}}{<}\, \infty,
\end{align*}
where (b) follows from item $(iii)$ of Assumption \ref{assumption:regularity-normality}.
The second result in the lemma thus again follows from Theorem 4.3 of \cite{newey1994large}.
\hfill $\blacksquare$
\vspace{3mm}

\revise{
\noindent \textbf{Proof of Lemma \ref{lem:lipschitz}.}
        Note that  
    $\partial \Lcal/\partial f_j = \sigma_j(\fvec) - \Indicator( y = j)$ and therefore for some $\fvec \in [\fvec, \fvec']$,
    \begin{align*}
        &|\Lcal(\fvec, y) - \Lcal(\fvec', y)|
    ~\stackrel{\text{(a)}}{\leq}~ 
    \left(
    \sum_{j \in \Kscr}(\sigma_j(\fvec) - \Indicator( y = j))^2
    \right)^{1/2}\cdot \lVert \fvec(\tilde{\x}) - \fvec'(\tilde{\x})  \rVert_2\\
    &~~~~~~~~~~~~~~~~~~~~~~~~~~~~~~~~~~~~~~~~~~~~~~~~~~~~~~~~~~~~~~~\stackrel{\text{(b)}}{\leq}~ 
    \left[\Indicator( y \neq 0) + \left(
    \sum_{j \in \Kscr}\sigma_j(\fvec)^2
    \right)^{1/2}\right]\lVert \fvec - \fvec'  \rVert_2~\leq~ 2\lVert \fvec - \fvec'  \rVert_2,
    \end{align*}
    where (a) follows from Cauchy–Schwarz inequality and (b) follows from a triangular inequality. 
    Note that this immediately implies that 
    $$
    |\Lcal(\fvec, y) - \Lcal(\fvec', y)| ~\leq~2 \lVert \fvec - \fvec'  \rVert_2 \leq 2\left(\lVert \fvec \rVert_2 + \lVert  \fvec'  \rVert_2\right) ~\leq~ 4\tilde{R}\sqrt{k}.
    $$
    This concludes the proof.
    \hfill $\blacksquare$

     \noindent \textbf{Proof of Lemma \ref{lem:local-rade}.}
        We sketch the proof here. Since the set $\left\{\hvec \in \Fscr^*: \lVert \hvec\rVert_{L_2(\Prob)}\leq \delta \right\}$ is larger when $\delta$ increases,  $\overline{\sf Rad}(\delta, \Fscr^*)$ is non-decreasing. 
        $\overline{\sf Rad}(\delta, \Fscr^*)/\delta$ is non-increasing because $\Fscr^*$ is star-shaped. 
        Writing the critical inequality as $\frac{\overline{\sf Rad}(\delta, \Fscr^*)}{\delta}\leq  \frac{\delta}{\tilde{R}\sqrt{k}}$, we observe that the right-hand side is strictly increasing. Therefore, there infimum is well-defined and for any $\delta \geq \delta_m^*$, $\overline{\sf Rad}(\delta, \Fscr^*)\leq  \frac{\delta^2}{\tilde{R}\sqrt{k}}$. 
        We next show that $\overline{\sf Rad}(\delta_m^*, \Fscr^*) = \frac{(\delta^*_m)^2}{\tilde{R}\sqrt{k}}$. 
        First of all, we show that $\overline{\sf Rad}(\delta, \Fscr^*)$ is continuous if $\delta \neq 0$ 
        Fix $\delta$. 
        Without loss of generality, we assume that for each realization of $\epsvec$ and $\tilde{\x}$, the supremum in the defintion of $\overline{\sf Rad}(\delta, \Fscr^*)$ is achieved and denote it by $\bar{\hvec}$ (otherwise, we may introduce additive error which can be made arbitrarily small and it does not influence the conclusion).
        Let $\alpha \in (0,1)$. 
        Then,  
        \begin{align*}
                  0 \,\leq\, & \overline{\sf Rad}(\delta, \Fscr^*) - \overline{\sf Rad}(\alpha\delta, \Fscr^*)\,\stackrel{\text{(a)}}{\leq}\, \E_{\tilde{\x},\epsvec}\left[
    \left|
    \frac{1}{m}\sum_{i=1}^m\sum_{j \in \Kscr}\epsilon_{ij}\bar{h}_j(\tilde{\x}_{i,(j)})
    \right| - 
    \left|
    \frac{1}{m}\sum_{i=1}^m\sum_{j \in \Kscr}\epsilon_{ij}\frac{\alpha\delta}{\alpha\delta\vee\lVert \bar{\hvec}\rVert_{L_2(\Prob)}}\bar{h}_j(\tilde{\x}_{i,(j)})
    \right| 
    \right]\\
     \,\stackrel{\text{(b)}}{\leq}\, & (1 - \alpha)  \E_{\tilde{\x},\epsvec}\left[
    \left|
    \frac{1}{m}\sum_{i=1}^m\sum_{j \in \Kscr}\epsilon_{ij}\bar{h}_j(\tilde{\x}_{i,(j)})
    \right| 
    \right] = (1 - \alpha)  \overline{\sf Rad}(\delta, \Fscr^*).
        \end{align*}
    Here (a) follows because $\alpha\delta\lVert \bar{\hvec}\rVert_{L_2(\Prob)}/(\alpha\delta\vee\lVert \bar{\hvec}\rVert_{L_2(\Prob)})\leq \alpha\delta$ and (b) follows since $\alpha \delta \vee \lVert \bar{\hvec}\rVert_{L_2(\Prob)} \leq \delta$. 
    Letting $\alpha \rightarrow 1$, this shows that $\overline{\sf Rad}(\delta, \Fscr^*)$  is left-continuous. 
    Also, 
    \begin{align*}
      0 \,\leq\, & \frac{\overline{\sf Rad}(\delta, \Fscr^*)}{\delta} - \frac{\overline{\sf Rad}\left({\delta}/{\alpha}, \Fscr^*\right)}{{\delta}/{\alpha}} \,\leq\, \E_{\tilde{\x},\epsvec}\left[
    \left|
    \frac{1}{\delta m}\sum_{i=1}^m\sum_{j \in \Kscr}\epsilon_{ij}\bar{h}_j(\tilde{\x}_{i,(j)})
    \right| - 
    \left|
    \frac{\alpha}{\delta m}\sum_{i=1}^m\sum_{j \in \Kscr}\epsilon_{ij}\bar{h}_j(\tilde{\x}_{i,(j)})
    \right| 
    \right]\\
    \,=\,&\frac{1-\alpha}{\delta}\E_{\tilde{\x},\epsvec}\left[
    \left|
    \frac{1}{m}\sum_{i=1}^m\sum_{j \in \Kscr}\epsilon_{ij}\bar{h}_j(\tilde{\x}_{i,(j)})
    \right|
    \right]\,=\, (1-\alpha)\cdot\frac{\overline{\sf Rad}(\delta, \Fscr^*)}{\delta}. 
    \end{align*}
    Let $\alpha \rightarrow 1$, we observe that $\frac{\overline{\sf Rad}(\delta, \Fscr^*)}{\delta}$ is right-continuous and hence 
    $\overline{\sf Rad}(\delta, \Fscr^*)$ is right-continuous. 
    The continuity of $\overline{\sf Rad}(\delta, \Fscr^*)$ implies that the inequality must bind at $\delta^*_m$.
    \hfill $\blacksquare$

    \noindent \textbf{Proof of Lemma \ref{lem:strong-convexity}.}
    We note that the derivative of $\Prob \Lcal_{\fvec}$ vanishes at $\fvec^*$ by optimality. Also it is easy to check the Hession of $\E_{y}\left[\Lcal(\fvec, y) | \tilde{\x}\right]$ satisfies that 
    $\Big[\nabla^2 \E_{y}\left[\Lcal(\fvec, y) | \tilde{\x}\right] \Big]_{j, \ell} = \Indicator(j = \ell)\sigma_j(\fvec) - \sigma_j(\fvec)\sigma_\ell(\fvec)$. 
    Therefore, 
    $
    \lambda_{\min}\left(\nabla^2 \E_{y}\left[\Lcal(\fvec, y) | \tilde{\x}\right]\right) \geq \sigma_0(\fvec)\min_{j \in \Kscr} \sigma_j(\fvec) \geq e^{\tilde{R}}(1 + ke^{\tilde{R}})^2.
    $
    By the Taylor expansion, the result holds. 
    \hfill $\blacksquare$

    \noindent \textbf{Proof of Lemma \ref{lem:local-complexity}.}
    Given sample data $\{\tilde{\x}_i\}_{i=1}^{m}$, we let the localized empirical Rademacher complexity be 
        $$
    \widehat{\sf Rad}(\delta, \Fscr^*) ~:=~ \E_{\epsvec}\left[
    \sup_{\hvec \in \Fscr^*: \lVert \hvec\rVert_{m}\leq \delta}\left|
    \frac{1}{m}\sum_{i=1}^m\sum_{j \in \Kscr}\epsilon_{ij}h_j(\tilde{\x}_{i,(j)})
    \right|
    \right], 
    $$
    and we define the smallest solution to $\widehat{\sf Rad}(\delta, \Fscr^*) \leq \frac{\delta^2}{\tilde{R}\sqrt{k}}$ as $\hat{\delta}^*_m$. Since $\widehat{\sf Rad}(\delta, \Fscr^*)/\delta$ is clearly non-increasing, this solution is well-defined. 
    Next, we have for any $\delta > 0$
    \begin{align}
    \label{eqn:empirical-rade}
        & \widehat{\sf Rad}(\delta, \Fscr^*) ~=~ \sqrt{\frac{k}{m}}\E_{\epsvec}\left[
    \sup_{\substack{\hvec \in \Fscr^*:\\ 
    \frac{1}{mk}\sum_{i=1}^m\sum_{j \in \Kscr}h^2_j(\tilde{\x}_{i(j)})
    \leq \delta^2/k
    }}\left|
    \sqrt{\frac{1}{mk}}\sum_{i=1}^m\sum_{j \in \Kscr}\epsilon_{ij}h_j(\tilde{\x}_{i,(j)})
    \right|
    \right]\\
    &\stackrel{(a)}{\preceq }
    \sqrt{\frac{k}{m}} \int_0^{2\frac{\delta}{\sqrt{k}}}\sqrt{\log(D(t, d^{\dag}))}dt \stackrel{(b)}{\leq} 
    \sqrt{\frac{k}{m}} \int_0^{2\frac{\delta}{\sqrt{k}}}\sqrt{\log\left(N\left(\frac{t}{2}, d^{\dag}\right)\right)}dt \stackrel{(c)}{\leq} 
    \sqrt{\frac{k}{m}} \int_0^{2\frac{\delta}{\sqrt{k}}}\sqrt{\tilde{d}\log\left(1 + \frac{8\delta}{t\sqrt{k}}\right)}dt \\
    &\stackrel{(d)}{=} 8\delta\sqrt{\frac{\tilde{d}}{m}}\int_0^{\frac{1}{4}}\sqrt{\log\left(1 + \frac{1}{\tilde{t}}\right)}d\tilde{t} \preceq \delta\sqrt{\frac{\tilde{d}}{m}}.
    \end{align}
    Here we have define the defined the distance function $$d^{\dag}(\hvec, \hvec') := \left(
    \frac{1}{mk}\sum_{i=1}^m\sum_{j \in \Kscr}\left(h_j(\tilde{\x}_{i(j)}) - h'_j(\tilde{\x}_{i(j)}) \right)
    \right)^{\frac{1}{2}}$$
    for $\hvec, \hvec' \in \Fscr^*$. We define $D(\cdot, \cdot)$ and $N(\cdot, \cdot)$ be the corresponding packing and covering number of the ball $\big\{\hvec \in \Fscr^*: \frac{1}{mk}\sum_{i=1}^m\sum_{j \in \Kscr}h^2_j(\tilde{\x}_{i(j)})^2
    \leq \delta^2/k\big\}$ under this distance function. 
    Here (a) follows from Corollary 2.2.8 in \cite{vaart2023empirical}, (b) follows from the fact that $N\left(\frac{t}{2}, d^{\dag}\right) \geq D\left({t}, d^{\dag}\right)$, (c) is a well-known result on the covering numbers of linear functions \cite[e.g., see Example 13.8][]{wainwright1945high}, and (d) follows from a change of variable argument. 

    To connect this to $\delta^*_m$, we bound $\widetilde{Z}_m(r): = \sup_{\hvec \in \Fscr^*: \lVert h \rVert_{L_2(\Prob)}\leq r}\left|\lVert \hvec \rVert^2_m - \lVert \hvec \rVert^2_{L_2(\Prob)}   \right|$.
    The arguments are largely parallel to that leading to \eqref{eqn:concentration-z}.
    Assume that $r \geq \delta^*_m$. 
    Using a similar argument to \eqref{eqn:Zm(r)}, we have 
    \begin{align}
    \label{eqn:tilde_Zm(r)}
            \E\left[ \widetilde{Z}_m(r) \right] \stackrel{\text{(a)}}{\leq} 8\tilde{R}\sqrt{k}\cdot\overline{\sf Rad}(r, \Fscr^*) \leq 8 r\delta^*_m \leq 8r^2
    \end{align}
    where in (a) we have used the lipschitz condition that for all $\hvec, \hvec' \in \Fscr^*$
    $$
    \left|\sum_{j\in \Kscr}\left(h^2_j(\x_{(j)}) - (h'_j)^2(\x_{(j)})\right)\right|~\leq~ \left\lVert 
    \hvec+\hvec'
    \right\rVert_{2}\cdot \left\lVert 
    \hvec-\hvec'
    \right\rVert_{2} ~\leq 4\tilde{R}\sqrt{k}\left\lVert 
  \hvec-\hvec'
    \right\rVert_{2}, 
    $$
    since it is easy to check that $\left\lVert 
    \hvec
    \right\rVert_{2}, \left\lVert 
    \hvec
    \right\rVert_{2} \leq 2\tilde{R}\sqrt{k}$.
    Furthermore, 
    $
    {\sf Var}\left(\lVert\hvec\rVert^2_2\right) \leq \Prob \lVert\hvec\rVert^4_2\leq 
    4kR^2\lVert\hvec\rVert^2_{L_2(\Prob)}\leq 4kR^2r^2, 
    $ because $\lVert \hvec \rVert^2_2 \leq 4kR^2$.
    So, 
     Then, 
    \begin{align*}
        &\E_{\x, \y}\left[\sup_{\substack{\hvec \in \Fscr^*: \\\lVert \hvec\rVert_{L_2(\Prob)} \leq r}} \Prob_m\left[
        \left(\lVert \hvec \rVert^2_2 - \lVert \hvec \rVert^2_{L_2(\Prob)}\right)^2
        \right]
        \right]\\\leq~& \E_{\x, \y}\left[\sup_{\substack{\hvec \in \Fscr^*: \\\lVert \hvec\rVert_{L_2(\Prob)} \leq r}} \Prob_m\left[
        \left(\lVert \hvec \rVert^2_2 - \lVert \hvec \rVert^2_{L_2(\Prob)}\right)^2
        \right] - {\sf Var}\left(\lVert\hvec\rVert^2_2\right)
        \right] + \sup_{\substack{\hvec \in \Fscr^*: \\\lVert \hvec\rVert_{L_2(\Prob)} \leq r}}{\sf Var}\left(\lVert\hvec\rVert^2_2\right)\\
        \leq~& 2\E_{\x, \y,\epsvec}\left[\sup_{\substack{\hvec \in \Fscr^*: \\\lVert \hvec\rVert_{L_2(\Prob)} \leq r}}
        \frac{1}{m}\left|\sum_{i=1}^m \epsilon_i \left(\lVert \hvec \rVert^2_2(\tilde{\x}_i) - \lVert \hvec \rVert^2_{L_2(\Prob)}\right)^2 \right|
        \right] + 4Rk^2r^2\\
        {\text{(b)}}{\leq}~&  256kR^2\sqrt{k}\E_{\x, \y,\epsvec}\left[\sup_{\substack{\hvec \in \Fscr^*: \\\lVert \hvec\rVert_{L_2(\Prob)} \leq r}}
        \frac{1}{m}\left|\sum_{i=1}^m \epsilon_i \Big(\lVert \hvec \rVert^2_2(\tilde{\x}_i) - \lVert \hvec \rVert^2_{L_2(\Prob)} 
        \Big) \right|
        \right] + 4kR^2r^2\\
        \leq~&
       512kR^2 \E_{\x}\left[ \widetilde{Z}_m(r) \right]
    +4kR^2r^2~\stackrel{\text{(c)}}{\leq}4100kR^2r^2,
    \end{align*}
    where (b) uses the bound $\lVert \hvec \rVert^2_2 \leq 4kR^2$ again and (c) follows from \eqref{eqn:tilde_Zm(r)}.

    To summarize, using the Talagrand inequality, we have for any 
    $u > 0$ and constants $c_8$ and $c_9$,
    \begin{align*}
        \Prob \left(\widetilde{Z}_m(r) \geq 8r\delta^*_m + u\right)~\leq~ c_8 e^{-\frac{c_9m u^2}{kR^2r^2 + kR^2u}}.
    \end{align*}
    to proceed, let us consider $r \geq 32\delta^*_m$ and set $u = r^2/4$, it holds that $\Prob \left(\widetilde{Z}_m(r) \leq \frac{r^2}{2}\right) \geq 1-  c_8e^{-\frac{c_{10}mr^2}{kR^2}}$.
    Also, when this good event happens, 
    \begin{align*}
    & \lVert \hvec \rVert^2_m - \lVert \hvec \rVert^2_{L_2(\Prob)}\\
    &~~~~~~~~~~~\leq~\Indicator\left(\lVert \hvec\rVert^2_{L_2(\Prob)}\leq r\right)\frac{r^2}{2} + \Indicator\left(\lVert \hvec\rVert^2_{L_2(\Prob)} > r\right)\frac{\lVert \hvec\rVert^2_{L_2(\Prob)}}{r^2}
    \left|
    \left\lVert \frac{r}{\lVert \hvec \rVert_{L_2(\Prob)}}\cdot\hvec \right\rVert^2_m - \left\lVert \frac{r}{\lVert \hvec \rVert_{L_2(\Prob)}}\cdot\hvec \right\rVert^2_{L_2(\Prob)}
     \right|\\
     &~~~~~~~~~~~\leq~
     \Indicator\left(\lVert \hvec\rVert^2_{L_2(\Prob)}\leq r\right)\frac{r^2}{2} + \Indicator\left(\lVert \hvec\rVert^2_{L_2(\Prob)} > r\right)\frac{\lVert \hvec\rVert^2_{L_2(\Prob)}}{r^2}\cdot \frac{r^2}{2} ~\leq~\frac{1}{2}\left(
    \lVert \hvec\rVert^2_{L_2(\Prob)} + r^2
     \right),
    \end{align*}
    or equivalently, $\lVert \hvec \rVert^2_m \leq \frac{3}{2}\lVert\hvec \rVert^2_{L_2(\Prob)} +\frac{1}{2}r^2$. 
    Then, we observe that 
    $$
    \omega(\delta)~:=~
    \E_{\epsvec}\left[
    \sup_{\substack{\hvec \in \Fscr^*:\\ \lVert \hvec\rVert_{L_2(\Prob)}\leq \delta}}\left|
    \frac{1}{m}\sum_{i=1}^m\sum_{j \in \Kscr}\epsilon_{ij}h_j(\tilde{\x}_{i,(j)})
    \right|
    \right] ~\leq~  \widehat{\sf Rad}\left(\sqrt{\frac{3\delta^2}{2} + \frac{r^2}{2}}, \Fscr^*\right) ~\stackrel{\text{(d)}}{\leq}~
    \sqrt{\left(\frac{3\delta^2}{2} + \frac{r^2}{2}\right)\cdot\frac{\tilde{d}}{m}},
    $$
    where (d) follows from \eqref{eqn:empirical-rade}, holds with probability at least $1-  c_8e^{-\frac{c_{10}mr^2}{kR^2}}$ for every $r \geq 32\delta^*_m$.
    This implies 
    $$
    \Prob(\omega(\delta^*_m) \geq t) ~\leq~ c_8 e^{-\frac{c_{10}m\left(
    2mt^2/\tilde{d} - 3(\delta^*_m)^2
    \right)}{kR^2}},~~~\forall t ~\geq~ \delta^*_m\sqrt{\frac{1027\tilde{d}}{2m}}. 
    $$
    Then, 
    \begin{align*}
        & \overline{\sf Rad}(\delta^*_m, \Fscr^*) ~=~ \int_{0}^{\infty} \Prob(\omega(\delta^*_m) \geq t)dt ~\leq~ \delta^*_m\sqrt{\frac{1027\tilde{d}}{2m}} + c_8\int_{\delta^*_m\sqrt{\frac{1027\tilde{d}}{2m}}}^{\infty}e^{-\frac{c_{10}m\left(
    2mt^2/\tilde{d} - 3(\delta^*_m)^2
    \right)}{kR^2}} dt\\
    & = \delta^*_m\sqrt{\frac{1027\tilde{d}}{2m}} + c_8e^{\frac{3c_{10}m(\delta^*_m)^2}{kR^2}}\cdot \sqrt{2\pi
    \cdot 
    \frac{k\tilde{d}R^2}{4c_{10} m^2}
    }\cdot\Prob
    \left(
    Z \geq \delta^*_m\sqrt{\frac{1027\tilde{d}}{2m}}\,
    \right)\\
    & \stackrel{\text{(d)}}{\leq} \delta^*_m\sqrt{\frac{1027\tilde{d}}{2m}} + c_8e^{\frac{3c_{10}m(\delta^*_m)^2}{kR^2}}\cdot \sqrt{
    \frac{\pi k\tilde{d}R^2}{2c_{10} m^2}
    }\cdot 2e^{-\frac{1027c_{10}(\delta^*_m)^2}{kR^2}}~\leq~
    \delta^*_m\sqrt{\frac{1027\tilde{d}}{2m}} + c_8 \sqrt{
    \frac{\pi k\tilde{d}R^2}{2c_{10} m^2}
    }\cdot 2e^{-\frac{1024c_{10}(\delta^*_m)^2}{kR^2}}\\
    &\stackrel{\text{(e)}}{\leq} c_{11}\left(
    \delta^*_m\sqrt{\frac{\tilde{d}}{m}} +\frac{R\sqrt{k\tilde{d}}}{m}
    \right)
    \end{align*}
    where we introduce an auxiliary random variable $Z \sim N\left(0, \frac{k\tilde{d}R^2}{4c_{10} m^2}\right)$ and (d) follows from the sub-Gaussian tail inequality and (e) holds because $e^{-\frac{1024c_{10}(\delta^*_m)^2}{kR^2}} \leq 1$.
    Using Lemma \ref{lem:local-rade}, we have 
    $$(\delta^*_m)^2 \leq c_{11}\tilde{R}\sqrt{k}\left(
    \delta^*_m\sqrt{\frac{d}{m}} +\frac{R\sqrt{k\tilde{d}}}{m}
    \right)\leq \frac{(\delta^*)^2}{2} + \frac{2c_{11}^2R^2 k \tilde{d}}{m} + \frac{R^2k\sqrt{\tilde{d}}}{m},$$
    which implies the desired conclusion. 
    \hfill $\blacksquare$}
\vspace{3mm}